%% file: ms.tex
\documentclass[a4paper]{article}

\usepackage[unicode, colorlinks]{hyperref} 
\input{packages.tex}

\usepackage{natbib}
\input{theoremenvs.tex}

\input{macros.tex}

\title{
	In almost all shallow analytic neural network optimization landscapes,
	efficient minimizers have strongly convex neighborhoods
}
\author{
	Felix Benning\thanks{Corresponding Author}\\
	University of Mannheim\\
	\texttt{felix.benning@uni-mannheim.de}
	\and
	Steffen Dereich\\
	University of Münster\\
	\texttt{steffen.dereich@uni-muenster.de}
}

\begin{document}
	\maketitle 

	\begin{abstract}
	\input{sections/abstract.tex}

	\smallskip
	\noindent \textbf{Key words and phrases.} \input{sections/keywords.tex}

	\end{abstract}

	{ \hypersetup{hidelinks} \tableofcontents }

	\input{sections/introduction.tex}

\input{sections/morse.tex}

\input{sections/efficient_param_characterization.tex}

	\input{sections/neighborhood_redundant_params.tex}

	\input{sections/existence_critical_points.tex}
	\input{sections/existence_redundancies.tex}

	\bibliography{references.bib}
	

\end{document}

%% file: packages.tex
\usepackage[utf8]{inputenc} 

\usepackage{enumitem} 

\usepackage{emptypage} 
\usepackage{xcolor}

\usepackage{mathtools} 
\usepackage{amsthm, amssymb} 
\usepackage{thmtools} 
\usepackage{aligned-overset} 


%% file: theoremenvs.tex
\theoremstyle{plain}
\newtheorem{prop}{Proposition}[section]
\newtheorem{lemma}[prop]{Lemma}

\newtheorem{theorem}[prop]{Theorem}

\theoremstyle{definition}
\newtheorem{definition}[prop]{Definition}

\newtheorem{example}[prop]{Example}

\theoremstyle{remark}
\newtheorem{remark}[prop]{Remark}

%% file: macros.tex
\newcommand{\dims}{d}

\newcommand{\cH}{\mathcal{H}}

\newcommand{\identity}{\mathbb{I}}

\newcommand{\transpose}{T}

\newcommand*{\real}{\mathbb{R}}
\newcommand*{\complex}{\mathbb{C}}
\newcommand*{\nat}{\mathbb{N}}
\newcommand*{\integer}{\mathbb{Z}}

\newcommand*{\Domain}{\mathcal{X}}
\newcommand*{\domain}{\mathbb{X}}
\newcommand*{\range}{\mathbb{Y}}

\newcommand*{\polyEfficient}{\mathcal{E}_P}
\newcommand*{\efficient}{\mathcal{E}}

\newcommand{\closure}[1]{\overline{#1}}

\newcommand{\metric}{\mathrm{d}}

\newcommand*{\problem}{\mathbf{m}}
\newcommand*{\Problem}{\mathbf{M}}
\newcommand*{\problemSpace}{\mathbb{M}}
\newcommand*{\target}{f}
\newcommand*{\kernel}{K}

\newcommand*{\Cost}{\mathbf{J}}
\newcommand*{\cost}{J}
\newcommand*{\loss}{\ell}

\newcommand*{\activation}{\psi}

\DeclareMathOperator{\sigmoid}{{\normalfont{sigmoid}}}

\newcommand*{\response}{\Psi}
\newcommand*{\regularizer}{R}

\newcommand*{\network}{\mathfrak{N}}
\newcommand*{\nodes}{\mathbb{V}}
\newcommand*{\param}{\theta}
\newcommand*{\paramSpace}{\Theta}
\newcommand*{\weight}{w}
\newcommand*{\bias}{\beta}
\newcommand*{\inSgt}{\blacklozenge}
\newcommand*{\outSgt}{*}
\newcommand*{\placeholder}{\bullet}

\newcommand*{\tin}{\text{in}}
\newcommand*{\tout}{\text{out}}

\newcommand*{\lebesgue}{\lambda}

\newcommand{\E}{\mathbb{E}}
\DeclareMathOperator{\cov}{Cov}
\DeclareMathOperator{\var}{Var}
\renewcommand{\Pr}{\mathbb{P}}
\newcommand*{\rf}{\mathbf{f}}
\newcommand*{\rg}{\mathbf{g}}

\newcommand*{\normal}{\mathcal{N}}
\DeclareMathOperator*{\supp}{supp}

\newcommand*{\C}{\mathcal{C}}
\newcommand*{\rP}{\mathbf{P}}

\newcommand*{\R}{\mathbb{R}}

\DeclareMathOperator{\mon}{mon}

\makeatletter 
\let\orgdescriptionlabel\descriptionlabel
\renewcommand*{\descriptionlabel}[1]{%
  \let\orglabel\label
  \let\label\@gobble
  \phantomsection
  \protected@edef\@currentlabel{#1\unskip}%
  \let\label\orglabel
  \orgdescriptionlabel{#1}%
}
\makeatother

%% file: sections/abstract.tex
Whether or not a local minimum of a cost function has a strongly convex
neighborhood greatly influences the asymptotic convergence rate of optimizers.
In this article, we rigorously analyze the prevalence of this property for the
mean squared error induced by shallow, 1-hidden layer neural networks with
analytic activation functions when applied to regression problems. 
The parameter space is divided into two domains: the \emph{efficient domain}
(all parameters for which the respective realization function cannot be
generated by a network having a smaller number of neurons) and the
\emph{redundant domain} (the remaining parameters).  In almost all regression
problems on the efficient domain the optimization landscape only features local
minima that are strongly convex. Formally, we will show that for certain
randomly picked regression problems the optimization landscape is almost surely a Morse
function on the efficient domain. The redundant domain has significantly smaller
dimension than the efficient domain and on this domain, potential local minima
are never isolated.

%% file: sections/keywords.tex
Artificial neural network, shallow network, analytic activation, Morse function,
strong convexity, regression problem

%% file: sections/introduction.tex

\section{Introduction}

Artificial neural networks (ANNs) define parametrized families of functions (the
realization functions) whose definition is inspired by biological neural
networks. Running optimization algorithms on these parametrized families (the
training of neural networks) has proven to be very efficient in various machine
learning tasks, including image recognition, natural language processing,
autonomous systems, protein folding, climate modelling.

The preferred method for the training of artificial neural networks (ANNs) are
Stochastic Gradient Descent (SGD) algorithms. The vanilla SGD algorithm
was first applied in \citet{rumelhartLearningRepresentationsBackpropagating1986}. Today, variants such as
momentum-based methods \citep{polyakMethodsSpeedingConvergence1964}, AMSProp
\citep{hintonNeuralNetworksMachine2012} and the Adam optimizer \citep{kingmaAdamMethodStochastic2015}
are more commonly used.

Generally, the efficiency of optimization algorithms is significantly
affected by the structure of the optimization landscape. 
The smoothing of updates in the momentum approaches seem to help with
saddle points and adaptive methods like RMSProp and Adam seem to adjust learning
rates better to navigate complex landscapes effectively. 

Mathematically rigorous approaches often assume that the SGD scheme
converges to a (local) minimum with a strongly convex neighborhood (meaning
that the Hessian of the landscape is strictly positive definite) or that a 
Polyak-Łojasiewicz inequality (in the strong sense with exponent \(2\)) applies.
SGD is typically applied with polynomially decaying step-sizes \(\gamma_n= c
n^{-\gamma}\) with $c,\gamma\in(0,\infty)$, $\gamma\le 1$ and, additionally, $c>
1/(2\rho)$, in the case of \(\gamma=1\),  where \(\rho\) is the spectral gap
between the spectrum of the Hessian and \(0\). In that case convergence of the
parameter occurs of order \(\sqrt{\gamma_n}\) in the number of steps~\(n\) and
in the loss we see convergence of order \(\gamma_n\).  In the original paper
introducing stochastic approximation techniques
\citep{robbinsStochasticApproximationMethod1951a} a first error analysis has been
conducted. Since then a variety of generalizations and extensions have been
proven, for instance CLTs \citep{sacksAsymptoticDistributionStochastic1958} and
non-asymptotic bounds \citep{moulinesNonAsymptoticAnalysisStochastic2011}.
When the order of convergence is \(\sqrt{\gamma_n}\), i.e.
\(\sqrt{n^{-\gamma}}\), the best order of convergence is obviously achieved with
the maximal decay rate \(\gamma=1\) resulting in step-sizes of order \(1/n\). But
since the spectral gap
\(\rho\) is typically not known in application it is hard to device such
algorithms. Moreover in practice, the choice \(1/n\) results in very slow convergence
in the first training phase when the algorithm is still far away from its limit
point. 
Polyak-Ruppert averaging (i.e. the use of Ces\`aro  average of the iterates of
the SGD scheme) overcomes these problems
\citep{polyakNewMethodStochastic1990,ruppertEfficientEstimationsSlowly1988} and
achieves convergence of order \(\sqrt{n^{-1}}\) even for smaller step size decay
\(\gamma\in(\frac12,1)\) under mild additional smoothness
assumptions.


\citet{kawaguchiDeepLearningPoor2016} shows that deep \emph{linear} networks have no poor
local minima, revealing the possibility of globally optimal solutions in
simplified settings. \citet{choromanskaLossSurfacesMultilayer2015} compare neural network losses
to spin-glass models and argue that so-called “bad” minima are rare in
high-dimensional settings. \citet{geEscapingSaddlePoints2015} introduce strict saddle
conditions that guarantee that SGD is not ``trapped'' in  saddle points.
\citet{nguyenConnectedSublevelSets2019} prove that for particular activation functions that  sublevel sets are connected provided that the data in the empirical risk minimization satisfies a non-degeneracy assumption. 
\citet{venturiSpuriousValleysOnehiddenlayer2019} examine the existence of spurious valleys in shallow
overparameterized ANNs and  \cite{freemanTopologyGeometryHalfRectified2017a} focus on half-rectified
networks. 

The referenced articles focus on the fact that SGD and similar numerical
methods in machine learning typically approach  “good” local minima. In this work, 
we aim to deepen the understanding of the \emph{second} training phase, when 
the numerical scheme has reached the vicinity of a local minimum and the crucial statistical properties are governed by the second order Taylor approximation of the loss-landscape around the local minimum. In this phase, we see fast convergence and know that averaging techniques are effective if the Hessian is strictly positive definite. Although many SGD schemes have been well analyzed under assumptions that imply the presence of 
fast convergence, proving that these assumptions actually hold is highly nontrivial.

Our approach is to examine the optimization landscapes of a broad class of regression 
problems with squared error loss for shallow ANNs using analytic activation 
functions. The main finding is that, in an appropriate sense, almost all such 
problems exhibit a “nice” optimization landscape. More precisely, we show that the optimization
landscape is typically Morse on the domain of \emph{non-degenerate} parameters. 
Conversely, the set of \emph{degenerate} parameters -- those for which the same response function can be accomplished with fewer neurons -- has 
significantly smaller Hausdorff dimension. Such parameters do not fully exploit 
the network's representational capacity, and inherently have redundancies that prevent 
the optimization landscape from being Morse at those points.

In order to state our results we start with a formal definition of shallow neural networks. The definition uses a graph structure that will prove to be useful later.


\begin{definition}[Shallow neural network]\label{def:ANN}
	\label{def: shallow neural network}
    A \emph{(dense) shallow neural network} (briefly called \emph{ANN}) is a tuple \(\network = (\mathbb V, \activation)\)  consisting of
\begin{itemize}
\item a tuple $\mathbb V=(V_0,V_1,V_2)$ of finite disjoint sets $V_0$, $V_1$ and
$V_2$ (the neurons of the input \(V_\tin := V_0\), hidden \(V_1\) and output
layer \(V_\tout := V_2\)) and
\item a measurable function \(\activation:\real\to \real\) (the activation
function).
\end{itemize}
\end{definition}

\begin{definition}Let \(\network=(\mathbb V, \psi)\) be an ANN.
\begin{enumerate}\item We call the directed graph $G=(V,E)$  given by 
\[ V=V_0\cup V_1\cup V_2\text{ \ and \ } E= \bigcup_{k=0}^1 V_k\times V_{k+1}\]
the \emph{ANN-graph} of \(\network\).
\item We call \(\Theta=\Theta_{\network}=\R^E\times \R^{V_1\cup V_2}\) \emph{parameter space of the network \(\network\)} and  
every tuple  \(\param=(\weight, \bias)\in \Theta=\R^E\times \R^{V_1\cup V_2}\) 
a \emph{parameter of the network~\(\network\)}. We refer to \(\weight\) as the 
 (edge) \emph{weights} and to \(\bias\) as the 
        \emph{biases}.  
        \item      
        For every parameter $\theta=(\weight,\bias)\in\Theta$ we call
        \begin{align}\label{eq:reali}
        \Psi_\param:\R^{V_{\tin}}\to \R^{V_{\tout}}  , \ x\mapsto  \Bigl(\bias_l
        + \sum_{j\in V_1} \activation\Bigl(
            \bias_j + \sum_{i\in V_{\tin}}  x_i\weight_{ij}\Bigr)\weight_{jl} 
        \Bigr)_{l\in V_{\tout}}
        \end{align}
the   \emph{response function} of the parameter $\param$.
\end{enumerate}\end{definition}
Typically, the underlying network is clear from the context and it is therefore omitted in the notation.

We study regression problems where the input data lies in \(\real^{V_\tin}\) and
the labels in \(\real^{V_\tout}\). These can be formally described by a
distribution \(\Pr_X\) on \(\real^{V_\tin}\) (the distribution of the input data)
and a probability kernel \(K\) from \(\real^{V_\tin}\)  to \(\real^{V_\tout}\)
(the conditional distribution of the label given the input data). Our aim is to
show that for a fixed distribution \(\Pr_X\) for ``most'' kernels $K$ the
respective optimization landscape is Morse on the efficient domain. For this we analyze
random regression problems where the kernel in the regression problem itself is
random.

\begin{definition}\label{def:873}
    Let \(\network\) be an ANN.  A \emph{measurable family of regression
    problems} is a tuple \(\mathfrak R=(\Pr_X,\kernel,\loss)\) consisting of
    \begin{enumerate}
        \item a distribution \(\Pr_X\) on \(\R^{V_\tin}\) (the distribution of the input data \(X\))
        \item a measurable set \((\problemSpace,\mathcal M)\) (the statistical model space)
        \item a probability kernel $\kernel$ mapping model and input data from
        \(\mathbb M\times \R^{V_\tin}\) to a probability distribution over labels in
        \(\R^{V_\tout}\) and
                \item a measurable function \(\loss:\real^{V_\tout}\times\real^{V_\tout}\to [0,\infty]\) (the \emph{loss}).
    \end{enumerate}
\end{definition}

When dealing with measurable families of regression problems we will always
associate the setting with a measurable space that is equipped with a family of
distributions (\(\Pr_\problem)_{\problem\in\problemSpace}\) together with a
\(\real^{V_{\tin}}\)-valued random variable \(X\) (the input data) and a
\(\real^{V_{\tout}}\)-valued random variable \(Y\) (the label) such that
under  every distribution \(\Pr_\problem\) with \(\problem\in\problemSpace\),
\(\Pr_X\)  is the distribution of \(X\) and \(K(\problem,\cdot;\cdot)\) is the 
conditional distribution of \(Y\) given \(X\), i.e.,
\[
    \Pr_\problem(Y\in B \mid X) = \kernel(\problem, X; B), \text{ \ a.s.}
\]

Then the loss that we incur when using a shallow ANN for the prediction defines
a optimization landscape in the sense of the following definition.

\begin{definition}[Cost function]\label{def:J_and_f}
    Let \(\network\) be an ANN as in Definition \ref{def:ANN}
   and \(\mathfrak R\) a measurable family of regression problems as in Definition~\ref{def:873} and a function \(R:\Theta\to\R\) (regularization). The family of functions  $(J_{\mathbf m})_{\mathbf m\in\mathbb M}$ given by
    \[
        J_{\problem}\colon \Theta \to (-\infty,\infty],
        \  \param\mapsto
        \E_\problem\bigl[\loss(\response_\param(X), Y)\bigr]+ \regularizer(\param) 
\]
the \emph{(regularized) cost functions of \((\network, \mathfrak R,R)\)}. 
\end{definition}

A useful concept for the analysis of the MSE cost function is the `target
function' representing the best possible predictor.

\begin{definition}[Family of \(L^p\)-integrable regression problems, target function] Let \(p\in[1,\infty)\). 
A measurable family of
    regression problems  \(\mathfrak R\) is said to be \emph{\(L^p\)-integrable}, if for every   \(\problem\in \problemSpace\) the label   is \(L^p\)-integrable, i.e.,
    \[
        \forall \mathbf m\in\mathbb M:\; \E_\problem\bigl[\|Y\|^p\bigr]<\infty.
    \]
For a family of \(L^1\)-integrable
    regression problems \(\mathfrak R\), for every   \(\problem\in \problemSpace\) the function
    \[
        \target_\problem(x) := \int  y\,K(\problem,x; dy) \overset{\Pr_X\text{-a.s.}}= \E_\problem[Y \mid X=x]
    \]
    is well-defined for \(\Pr_X\)-almost all \(x\in \real^{V_\tin}\).
    We call \(\target_\problem\) the \emph{target function of \(\problem\)}.
\end{definition}

The conditions assumed for our main result are collected in the following
definition.
\begin{definition}\label{def: standard model}
    The \emph{standard setting} is a tuple \((\network,  \mathfrak R, \regularizer,\Problem)\)
    consisting of 
    \begin{itemize}
    \item an ANN $\network$ with one dimensional output $\#V_\tout=1$ and
    analytic activation function \(\activation\),
    \item a  family
    of  \(L^2\)-integrable regression problems \(\mathfrak R\) with squared-error loss
    \(\loss(\hat{y}, y) = (\hat{y}-y)^2\) and compact support 
    \(\Domain:=\mathrm{supp}(\Pr_X)\) of the input distribution \(X\),
    \item  analytic convex (regularization) function \(\regularizer:\Theta\to \R\) and
    \item an $\problemSpace$-valued random variable $\Problem$ such that
    the random target function \(\rf:=\target_\Problem\) is an \emph{weakly universal Gaussian random function} in the sense that for every continuous test function \(\phi:\real^{V_\mathrm{in}}\to\real\) the random variable
    \[ 
    \langle \phi, f_\mathbf{M}\rangle_{\Pr_X}:=\int \phi(x) \,\target_\Problem(x) \,\Pr_X(dx)
    \]
    is Gaussian and has strictly positive variance whenever \(\phi\not\equiv 0\)
    on \(\Domain\).\footnote{
        Note that for every \(\problem\in\problemSpace\),
        \(\target_\problem\) is in \(L^2(\Pr_X)\subseteq L^1(\Pr_X)\) and \(\phi\) is uniformly
        bounded on the compact support of \(\mathbb P_X\) so that the integral
        \(\langle \phi, \target_\Problem\rangle_{\Pr_X}\) is for \emph{all} realizations
        of \(\mathbf M\) well-defined. It  is further  measurable since
        \(f_\cdot(\cdot)\) is product measurable by Fubini's theorem.
    }
    \end{itemize}
\end{definition}

To understand the optimization landscape of models $\problem\in\problemSpace$  we need
to divide the space of all parameters $\param\in \paramSpace=\R^E\times \R^{V_1\cup
V_2}$ into two domains. For the activation functions \(\sigmoid\) and \(\tanh\)
we define
\begin{itemize}
\item the \emph{efficient domain} by
\begin{equation}
    \label{eq: defintion efficient 0} 
    \efficient_0
    := \Biggl\{
        \param =(\weight, \bias)\in \real^{E \times (V\setminus V_0)}:
        {
        \begin{aligned}[c]
            \scriptstyle \weight_{j\placeholder}
            &\scriptstyle\neq 0
            &&\scriptstyle \forall j \in V_1,
            \\[-0.5em]
            \scriptstyle\weight_{\placeholder j}
            &\scriptstyle\neq 0
            &&\scriptstyle\forall j\in V_1, 
            \\[-0.5em]
            \scriptstyle(\weight_{\placeholder i}, \bias_i)
            &\scriptstyle\neq \pm(\weight_{\placeholder j}, \bias_j)
            &&\scriptstyle\forall i,j\in V_1\text{ with }i\not=j
        \end{aligned}
        }
    \Biggr\},
\end{equation}
\item the \emph{redundant domain} by \((\real^E\times \real^{V_1\cup V_2})\backslash \efficient_0\).
\end{itemize}

Our main structural result, namely the fact that the cost is typically Morse, is
only true on the efficient domain. The restriction onto the efficient domain is
natural since any redundant parameter lies on a path of constant response
such that no local minimum in the redundant domain can be a strict local
minimum. In particular the cost cannot be Morse on the whole set of parameters.

Our main result states that for ``most'' statistical models $\problem$
the realization of the MSE is Morse on the efficient domain. 

\begin{theorem}[Almost all optimization landscapes are Morse on the efficient domain]\label{thm:main1}
	Let \((\network, \mathfrak R,R, \Problem)\) be a standard setting
    (Definition~\ref{def: standard model}). Assume \(\activation\in \{\sigmoid,
    \tanh\}\) about the activation function and that the support of
    \(\Pr_X\) contains a non-empty open set.
	Almost surely, the regularized cost \(\cost_\Problem\colon \efficient_0\to \R\) with
	\[  
        \cost_\Problem(\param) = \E_\Problem\bigl[\loss(\response_\param(X), Y)\bigr]+ \regularizer(\param) 
	\]
	is a Morse function. Equivalently, it holds that 
	\[
		\Pr\Bigl(
			\exists \param \in \efficient_0:
			\nabla J_{\mathbf M}(\param) = 0, \,\det(\nabla^2 J_{\mathbf M}(\param)) = 0
		\Bigr) = 0.
	\]
\end{theorem}

In Section~\ref{sec:Morse} we actually prove a version of this theorem for general
analytic activation functions (see Theorem~\ref{thm:main1.1}). This requires
a notion of the efficient domain that is an implicitly defined set. In
Section~\ref{sec: characterization of efficient params} we then prove that for the activation functions
\(\mathrm{sigmoid}\) and \(\mathrm{tanh}\) the implicitly defined version of the
efficient domain agrees with the one used in the latter theorem.

\begin{remark}[Generalization of the Gaussian assumption]
    While we assume that the target function is weakly universal Gaussian in the
    standard setting (Definition~\ref{def: standard model}), our main theorem
    (Theorem~\ref{thm:main1}) is a statement about null sets. Since null sets
    remain null sets for measures that are absolutely continuous with respect to
    such Gaussian measures and mixtures thereof, it is straight-forward to
    generalize the statement to significantly more general distributions of the
    random target function \(\target_\Problem\). That is, the statement remains
    true if the distribution of \(\target_\Problem\) can be written as a mixture
    of measures that are absolutely continuous with respect to weakly universal
    Gaussian measures!
\end{remark}
\begin{remark}[Weak universality]
    \label{rem: weak universality}
    For a better understanding of weak universality consider the stronger
    assumption\footnote{
        On the positive probability event in \eqref{eq: positive ball probability} we have
        \[
            |\langle \phi, \target_\Problem\rangle_{\Pr_X}-  \|\phi\|_{\Pr_X}^2| =   |
            \langle \phi, \target_\Problem- \phi \rangle_{\Pr_X}| \le  \epsilon \|\phi\|_{\Pr_X}.
        \]
        Since \(\phi\) is continuous and non-zero on the support of \(\Pr_X\) we
        have \(\|\phi\|_{\Pr_X}>0\). Choosing
        \(\epsilon\in (0,\|\phi\|_{\Pr_X})\) we conclude that \(\langle \phi,
        f_\mathbf{M}\rangle_{\Pr_X}>0\) with strictly positive probability. The same
        argument applied to \(-\phi\) gives that \(\langle \phi,
        f_\mathbf{M}\rangle_{\Pr_X}<0\) with strictly positive probability.
        Consequently, \(\langle\phi, f_\mathbf{M}\rangle_{\Pr_X}\) has positive variance.
    } that all continuous functions
    \(\phi\colon \real^{V_\tin}\to\real^{V_\tout}\) lie in the support of
    \(\Pr_{\target_\Problem}\), when \(\target_\Problem\) is a random element
    in \(L^2(\Pr_X)\). I.e. for every continuous
    \(\phi\colon \real^{V_\tin}\to\real^{V_\tout}\) and \(\epsilon>0\), one
    has that 
    \begin{equation}
        \label{eq: positive ball probability}    
        \Pr\bigl(\|f_\mathbf{M}-\phi\|_{\Pr_X}<\epsilon\bigr)>0.
    \end{equation}
    This is a \emph{universality} assumption \citep[Thm.\@
    3.6.1]{micchelliUniversalKernels2006, carmeliVectorValuedReproducing2010,
    bogachevGaussianMeasures1998} and intuitively means that no continuous
    function \(\phi\) can be ruled out as the target function
    \(\target_\Problem\) ex ante. We believe this is a natural assumption for
    a learning problem.

    In the proof of Theorem~\ref{thm:main1}, we will actually work with
    an even weaker assumption than weak universality: It would suffice to assume
    the non-degeneracy for real-analytic test functions \(\phi\) only.
\end{remark}

A natural question is whether local minima on the efficient domain exist and whether the restriction to the efficient domain in
Theorem \ref{thm:main1} is an artefact of our proof. This will be the content of
Sections \ref{sec: neighborhood of redundant parameters}-\ref{sec: existence of
redundant critical points}.  Intuitively we show, for the standard
unregularized setting with activation \(\activation\in \{\sigmoid, \tanh\}\) and
the additional regularity assumption \eqref{eq: positive ball probability} in
Remark \ref{rem: weak universality}, that we have the following:
\begin{itemize}
    \item For every open set \(U\subset \paramSpace\) containing an efficient
    point, the probability is strictly positive that the  loss has a local
    minimum in \(U\), see Theorem \ref{thm: existence of efficient minima}.
    
    \item With strictly positive probability, there exist critical points in the
    set of redundant parameters (Theorem \ref{thm: redundanct critical points exist}) and all redundant critical points have
    a direction of zero curvature (the determinant of the Hessian is zero), see Theorem \ref{thm: critical points of redundant type}.
\end{itemize}

   
It is therefore \emph{impossible} to prove the MSE to be a Morse function
on the redundant domain, since critical points may exist and those always
violate the Morse condition.

\paragraph*{Outline}

In Section \ref{sec:Morse} we prove a more general version of Theorem \ref{thm:main1}
which is applicable to all analytic activation functions. However, for
general analytic activation functions the efficient domain has to be defined in
an implicit way. Specifically, we will prove in Theorem~\ref{thm:main1.1} for
the standard setting that the optimization landscape is almost surely Morse on the set
of \emph{polynomially efficient parameters} (Definition~\ref{def: polynomial
independence}). In Section~\ref{sec: characterization of efficient params} we
show that the various
definitions of efficient parameter domains coincide for \(\activation\in
\{\sigmoid, \tanh\}\) (Theorem~\ref{thm: characterization of efficient networks}).
With this result Theorem~\ref{thm:main1} becomes a direct corollary of
Theorem~\ref{thm:main1.1}. In Section \ref{sec: neighborhood of redundant parameters}
we prove for any redundant parameter \(\param\) that there exists a straight line 
of parameters \((\param(t))_{t\in\real}\) passing \(\param\), where the response, and
therefore the cost in the unregularized setting, remains unchanged. In Section \ref{sec: existence of
efficient critical points} we prove that efficient local minima exist
with positive probability. We use this fact in Section \ref{sec: existence of
redundant critical points} to prove that redundant critical points exist with positive
probability. To show this we extend an efficient critical parameter of a smaller
network to a redundant critical parameter of a larger network.

%% file: sections/morse.tex
\section{MSE is Morse on efficient domain}
\label{sec:Morse}

In this section, we will prove that for a standard model the random
loss-landscape is Morse on the \emph{polynomially}, efficient domain. For
general activation functions~\(\activation\) we have to work with a different
notion of the efficient
domain than \(\efficient_0\) introduced in \eqref{eq: defintion efficient 0}. As we will show in Section~\ref{sec: characterization of efficient
params} the definition coincides with \(\efficient_0\) whenever
\(\activation\in\{\sigmoid, \tanh\}\) and the support of \(\Pr_X\)
contains an open set.

\begin{definition}[Polynomial efficiency]
	\label{def: polynomial independence}
    Let \(\network = (\nodes, \activation)\) be an ANN (Definition~\ref{def: shallow neural network}), \(n\in\nat_0\) and \(m=(m_{\emptyset}, m_0, \dots, m_n)\in\nat_0^{n+2}\).
	\begin{enumerate}[label=(\roman*)]
        \item
        A parameter \(\param\in\Theta\) is called \emph{\(m\)-polynomially
        independent on \(\Domain\)} if \(\activation\) is
        \(n\)-times differentiable and the equation
        \[
            0 = P^{(\emptyset)}(x)
            + \sum_{j\in V_1}
            \sum_{k=0}^n P^{(k)}_j(x) \activation^{(k)}\Bigl(\bias_j + \sum_{i\in V_0}x_i\weight_{ij} \Bigr)
            \quad \forall x\in \Domain
        \]
        considered in all polynomials \(P^{(\emptyset)}\) and \((P^{(k)}_j:j\in V_1,
        k\in\{0,\dots,n\})\) of at most degree \(m_\emptyset\) and  \(m_k\),
        respectively, has only the trivial solution where all polynomials are
        identically zero.
        Here, \(\activation^{(k)}\) denotes the \(k\)-th derivative of the activation function \(\activation\).
        \item
        A parameter \(\param\in\Theta\) is called \emph{\(m\)-polynomially efficient on \(\Domain\)}, if 
        \begin{enumerate}[label={(\alph*)}]
            \item
            all neurons are used meaning that for all  \(k\in V_1\) one has
            \[
                \weight_{k \placeholder} = (\weight_{kl})_{l\in V_\tout} \not\equiv
                0,
            \]
            and
            \item it is \(m\)-polynomially independent.
        \end{enumerate}
	We denote by \(\polyEfficient^m=\polyEfficient^m(\Domain)\) the set of all
	\(m\)-polynomially efficient parameters.
    \end{enumerate}
\end{definition}

\begin{theorem}[MSE is a Morse function on polynomially efficient parameters]
	\label{thm: morse function on efficient networks}\label{thm:main1.1}
	Let \((\network, \mathfrak R, R, \Problem)\) be the standard setting
    (Definition~\ref{def: standard model}).
	Then the MSE cost is almost surely a Morse function on the set
	\(\polyEfficient:=\polyEfficient^{(0,0,1,2)}(\Domain)\) of \((0,0,1,2)\)-polynomially
	efficient parameters on the support \(\Domain\) of \(\Pr_X\), i.e.
	\[
		\Pr\Bigl(
			\exists \param \in \polyEfficient:
			\nabla\Cost(\param) = 0, \det(\nabla^2\Cost(\param)) = 0
		\Bigr) = 0
	\]
\end{theorem}

Before we explain the methodology of our proof we first derive a crucial
representation for the MSE cost \(\cost_\problem\) given by 
\[
    \cost_\problem(\param)
    = \E_\problem\bigl[ \|\response_\param(X) - Y\|^2\bigr] +\regularizer(\param) 
\]
with convex regularizer \(R\). Recall that \(\Psi_\param\) is the realization function of the ANN as introduced in~(\ref{eq:reali}).

\begin{prop}[Decomposition of the MSE cost]
    \label{prop: decomposition}
    For \(\problem\in \problemSpace\) and \(\param\in\paramSpace\)
    one has
    \begin{equation}\label{eq: decomposition}
        \cost_\problem(\param)
        = \regularizer(\param) + \|\response_\param\|_{\Pr_X}^2
        - 2\underbrace{
            \langle \response_\param, \target_\problem\rangle_{\Pr_{X}}
        }_{=:\hat{\cost}_\problem(\param)}
        + \E_\problem[\|Y\|^2].
    \end{equation}
    Here \(\|\cdot\|_{\Pr_X}\) is induced by \(\langle \phi, \varphi\rangle_{\Pr_X} := \int \langle\phi(x), \varphi(x)\rangle \Pr_X(dx)\).
\end{prop}
\begin{proof}
    With the Pythagorean formula we have
    \begin{align*}
        \cost_\problem(\param)
        &= \regularizer(\param) + \E_\problem[\|\response_\param(X) - Y\|^2]
        \\
        &= \regularizer(\param) + \E_\problem[\|\response_\param(X)\|^2] - 2\E_\problem[\langle \response_\param(X), Y\rangle] + \E_\problem[\|Y\|^2].
    \end{align*}
    Since \(\target_\problem(X)=\E_\problem[Y|X]\) we conclude with the tower property
    \[
        \E_\problem[\langle \response_\param(X), Y\rangle] 
        = \E_\problem[\langle \response_\param(X), \target_\problem(X)\rangle]
        \overset{\text{def.}}= \langle \response_\param, \target_\problem \rangle_{\Pr_X}
        =\hat{\cost}_\problem(\param).
        \qedhere
    \]
\end{proof}

In view of  \eqref{eq: decomposition} we observe that the term
\(\E_\problem[\|Y\|^2]\) does not depend on the parameter \(\param\) and it is
thus irrelevant when it comes to deciding whether \(\cost_\problem\) is Morse or
not. In the proof we then argue that the event where the stochastic process
\((R(\param)+\|\response_\param\|_{\rP_X}^2-2\hat{\cost}_\Problem(\param))_{\param\in\Theta}\) is not Morse on the
polynomially efficient parameters is a \emph{``thin set''}.

Unfortunately, our setting is not immediately covered by the arguments of
\citet{adlerRandomFieldsGeometry2007}. Roughly speaking, their approach is as
follows. If there is a parameter \(\param\) that is critical with its Hessian
having a zero eigenvalue, then this satisfies
\begin{align}\label{eq: morse violation}
    \nabla \Cost(\param)=0 \quad\text{and}\quad
    \det(\nabla^2  \Cost(\param))=0.
\end{align}
Note that the latter is a collection of \(\dim(\paramSpace)+1\) real equations in
\(\dim(\paramSpace)\) real variables and intuitively one would expect that, under
appropriate non-degeneracy assumptions, the equation does not have solutions.
The equations in~\eqref{eq: morse violation} depend on the collection of first
order differentials \(\rg_1(\param)\) and of second order differentials
\(\rg_2(\param)\). As shown in Lemma~11.2.10 of
\citet{adlerRandomFieldsGeometry2007} solutions of~\eqref{eq: morse violation} would
not exist, if for every \(\param\) under consideration (in our
case the efficient domain) the combined vector \((\rg_1(\param),\rg_2(\param))\)
has locally uniformly bounded Lebesgue density. 

Unfortunately,  in our situation, many second order differentials are degenerate and the
result is not applicable. To bypass this problem we proceed as follows.  In the
following  subsection, we will first analyze the stochastic process 
\[
    \hat{\Cost}
    =(\hat{\cost}_\Problem(\param))_{\param\in \paramSpace}
    =(\langle \response_\param,\target_\Problem\rangle_{\Pr_X})_{\param \in \paramSpace} .
\]
This process is obtained by applying a \(\param\)-dependent linear functional on
the random target function \(\rf=\target_\Problem\) and thus \(\hat{\Cost}\) is
a Gaussian process since \(\rf\) is Gaussian by assumption. We will
collect in \(\rg_1(\param)\) all first order differentials and in
\(\rg_2(\param)\) the \emph{`centered'}\footnote{
    \label{footnote: centered}
    This is only true if \(\target_\Problem\) is centered, which we are not
    willing to assume. But this provides the right intuition, since we subtract a
    deterministic term (which is not necessarily the mean).
} and \emph{non-degenerate} second order differentials. The
non-degeneracy of the combined collection \(\rg= (\rg_1,\rg_2)\) is shown in
Proposition~\ref{prop: non-degenrate g} and follows from the polynomial
independence that is assumed in the polynomially efficient domain (cf.~Definition~\ref{def:
polynomial independence}).

In Proposition~\ref{prop: graph of g is not in U}
we show that the generalization of the volume argument of
\citet[Lemma~11.2.10]{adlerRandomFieldsGeometry2007}
given in Lemma~\ref{lem: generalization 11.2.10} is
applicable to \(\rg\). The generalization of the volume argument
is necessary since we want to show that the process
\[
    \bigl(\regularizer(\param) + \|\response_\param\|_{\Pr_X}^2-2 \hat{\Cost}(\param)\bigr)_{\param\in\paramSpace}
\]
never satisfies \eqref{eq: morse violation} on \(\polyEfficient\).
While \citet{adlerRandomFieldsGeometry2007} considered level sets, we move
the model indepeendent term \(\regularizer(\param) + \|\response_\param\|_{\Pr_X}^2\) to the other side in \eqref{eq:
morse violation} and therefore need to consider function graph intersections.
Proposition~\ref{prop: graph of g is not in U} would then immediately yield the
Morse property if all second order derivatives would be contained in
\(\rg_2(\param)\).

The subsequent subsection (Section~\ref{sec: proof of morse result}) finishes
the proof of Theorem~\ref{thm: morse function on efficient networks}.
To do so we carefully craft a thin set \(U\) as the zero set of a function \(F\)
which \(\rg\) may not intersect. Although \(\hat {\mathbf J}(\param)\) has
degenerate second order differentials in the last layer,
the additional deterministic term \(\regularizer(\param) + \|\response_\param\|_{\Pr_X}^2\) that
is strictly convex in the last layer helps us out. More explicitly, we
will design a real analytic function \(F\) taking an outcome of \((\param,
\rg_2(\param))\) to a real value in such a way that for all
\(\param\in\polyEfficient\)
\[
    \nabla \mathbf J(\param)=0
    \; \implies \;
    \det(\nabla^2 \mathbf J(\param))= F(\param,\rg_2(\param)).\ 
\]
Consequently, if there is a parameter \(\param\in\polyEfficient\) with 
\[ \nabla \mathbf J (\param)=0 \text{ \ and \ } \det(\nabla^2  \mathbf J (\param))=0,
\]
then we also found a solution to 
\[ \rg_1 (\param)=0 \text{ \ and \ } F(\param , \rg_2 (\param))=0.
\]
This is again a collection of \(\dim(\Theta)+1\) real equations in \(\dim(\Theta)\) variables and we formally conclude with Proposition~\ref{prop: graph of g is not in U} that, almost surely, no solutions exist. Note that we are now able to proceed since the latter equations only make use of the non-degenerate differentials of first and second order of \(\hat{\mathbf J}(\param)\).

%
%
%

\subsection{Analysis of \texorpdfstring{\((\hat{\Cost}(\param))\)}{(ˆJ(θ))}}

In this section we analyze the stochastic process \((\hat{\Cost}(\param))_{\param\in \paramSpace}\). We will
\begin{itemize}
    \item derive representations for differentials of \((\hat{\Cost}(\param))_{\param\in \paramSpace}\) (Lemma~\ref{lem: differentiability})

    \item show non-degeneracy of the combined vector
    \((\rg_1(\param),\rg_2(\param))\) for all \(\param\in\polyEfficient\), where
    \(\rg_1(\param)\) and \(\rg_2(\param)\) are constituted by all first order
    differentials and certain second order differentials  of \(\hat{\mathbf
    J}(\param)\), respectively (Proposition~\ref{prop: non-degenrate g})

    \item generalize the volume argument of \citet{adlerRandomFieldsGeometry2007} to our needs
    (Lemma~\ref{lem: generalization 11.2.10}).
\end{itemize}
The combination of all these results leads to Proposition~\ref{prop: graph of g is not in U} which will allow us to show the Morse property in the subsequent section.

\begin{lemma}[Differentiability]
    \label{lem: differentiability}
    Let \(k\in\nat\), \(\network = (\nodes, \activation)\) be an ANN with a \(C^k\) activation function
    \(\activation\) and \(\# V_\mathrm{out}=1\), let \(\mathfrak R\) be a family of \(L^1\)-integrable regression problems with  \(\Pr_X\) having compact domain and let \(\mathbf m\in \mathbf M\).
    Then \[\hat{\cost}_\problem(\param) = \langle \response_\param,
    \target_\problem \rangle_{\Pr_X}\] is in \(C^k\) and its partial derivatives satisfy
    \[
        \partial_\param^\alpha \hat{\cost}_\problem(\param)
        = \langle \partial_\param^\alpha \response_\param, \target_\problem\rangle_{\Pr_X}
    \]
    for all multi-indices \(|\alpha|\le k\).
\end{lemma}
Note that the lemma implies that in the standard setting all differentials of  \((\hat{\Cost}(\param))_{\param\in \paramSpace}\) define again Gaussian processes. This fact together with the representations for the differentials will be the basic tool in the analysis of  \((\hat{\Cost}(\param))_{\param\in \paramSpace}\) and we mostly will not give reference to the lemma when using it.

\begin{proof}
By assumption, one has that \(\E_\problem[\|Y\|] < \infty\). Recall that \(\target_\problem(x) = \E_\problem[Y \mid X=x]\) so that with the \(L^1\)-contraction property of the conditional expectation
    \begin{align}\label{eq8723}
        \int |f_\mathbf{m}(x)|\, \Pr_X(dx)
        = \E_\problem\bigl[|\E_\problem[Y\mid X]|\bigr]
        \le
        \E_\problem\bigl[\E_\problem[|Y| \mid X]\bigr]
        = \E_\problem[|Y|^2] < \infty.
    \end{align}
     
    Fix \(\param\in\Theta\) and \(v\in \Theta\).  By assumption, \(\Pr_X\) has
    compact support \(\Domain\) and the directional derivative \(D_v^\param
    \response\) in direction \(v\) in the \(\param\) component is a continuous
    mapping on \(\Theta\times \Domain\). In particular, it is  uniformly bounded
    on the compact set \(\overline{B(\param,\|v\|)}\times\Domain\), say by the
    constant \(C\). Consequently, for every \(t\in(0,1]\), one has that
     \begin{align}
        \nonumber
        \frac 1t(\hat{\cost}_\problem(\param + tv) - \hat{\cost}_\problem(\param))
        &= \frac1t\int (\response_{\param + tv}(x)- \response_\param(x)) \target_\problem(x)\, \Pr_X(dx)
        \\
        \nonumber
        \overset{\text{FTC}}&= \int \int_0^1 D_v^\param\response_{\param + stv}(x)  \, \target_\problem(x)\, ds\,\Pr_X(dx)
    \end{align}
    Now note that \(C |f_\mathbf{m}|\) is an integrable majorant due to~(\ref{eq8723}). Using that for every \(s\in [0,1]\) and \(x\in\domain\), \(\lim _{t\downarrow 0} D_v^\param\response_{\param + stv}(x)= D_v^\param\response_{\param}(x)\) by continuity of the differential it follows with dominated convergence that
\[\lim_{t\downarrow 0}
\frac 1t(\hat{\cost}_\problem(\param + tv) - \hat{\cost}_\problem(\param))= \int  D_v^\param\response_{\param}(x)  \, \target_\problem(x)\,\Pr_X(dx).
\]
Recall that \((\param,v)\mapsto D_v^\param\response_{\param}(x)\) is continuous and we get again with dominated convergence that the latter integral is continuous in the parameters \(\theta\) and \(v\). This proves that \(\hat J_\mathbf{m}\) is \(C^1\) and that the upper identity holds.

By induction, one obtains the general statement. The induction step can be carried out exactly as above by using that the assumptions imply that \(\Psi\) is \(k\)-times continuously differentiable as mapping on \(\Theta\times \domain\).
\end{proof}

As indicated before there are second order differentials that degenerate.
However, for all first order and some second order differentials this is not the
case. In the next step we will show this. 
We consider the stochastic processes
\(\rg_1=(\rg_1(\param))_{\param\in\param}\) and
\(\rg_2=(\rg_2(\param))_{\param\in\param}\) defined by
\begin{align}
        \label{eq: definition of g1}
    \rg_1(\param)&:=\nabla\Cost(\param) \qquad\text{and}\\
    \label{eq: definition of g2}
    \rg_2(\param)
    &:=\Bigl(
        \bigl(\partial_{\bias_j}^2\hat{\Cost}(\param)\bigr)_{j\in V_1},
        \bigl(\partial_{\bias_j}\partial_{\weight_{ij}}\hat{\Cost}(\param)\bigr)_{\substack{j\in V_1\\i\in V_0}},
        \bigl(\partial_{\weight_{ij}}\partial_{\weight_{kj}}\hat{\Cost}(\param)\bigr)_{\substack{j\in V_1\\ i,k\in V_0\\i\le k}}
    \Bigr),
\end{align}
where we assume some total order on the input neurons \(V_0\) such that \(i\le k\)
makes sense for \(i,k\in V_0\).
Note that \(\rg_1\) utilizes the un-centered \(\Cost\) to ensure \(\nabla\Cost(\param) = 0\)
translates to \(\rg_1(\param)=0\) whereas \(\rg_2(\param)\) utilizes the
`centered'\footref{footnote: centered}
\(\hat{\Cost}\). This is because \(\rg_2\) does not contain all second order differentials
and a translation function \(F\) is necessary to get from \(\rg_2\)
to \(F(\param, \rg_2(\param)) = \det(\nabla^2\Cost(\param))\). Constructing
\(F\) in turn is more straightforward with the `mean'
\(\|\response_\param\|_{\Pr_X}^2\) built into \(F\) (cf.~Section~\ref{sec: proof
of morse result}).

\begin{prop}
    \label{prop: non-degenrate g}
    For an ANN \(\network = (\nodes, \activation)\) let \(\# V_\tout=1\).
    We consider \(\rg_i\) as defined in \eqref{eq: definition of g1} and \eqref{eq: definition of g2}
    based on the standard Gaussian setting (Definition~\ref{def: standard model}).
    Then for every parameter \(\param\in\mathcal \polyEfficient\) the  Gaussian
    random vector \((\rg_1(\param), \rg_2(\param))\) is
    non-degenerate meaning that its covariance has full rank. 
\end{prop}

\begin{proof}
    Recall that \(\hat{\Cost}(\param) = \langle \response_\param, \rf\rangle_{\Pr_X}\). 
    Since the variance does not depend on the mean we can assume without loss of
    generality \(\rg_1(\param) = \nabla \hat{\Cost}(\param)\) in this proof. 
    Let \(I_1\) and \(I_2\) be  index sets such that\footnote{
        With slight misuse of the notation, we ignore the ordering of the
        differentials in the representation of \(\rg_2(\param)\).
    }
    \[
        \rg_1(\param)=\nabla \hat{\Cost}(\param)
        = (\partial_{\param_i}\hat{\Cost}(\param))_{i\in I_1}
        \text{ \ and \ }
        \rg_2(\param)
        = (\partial_{\param_i}\partial_{\param_j}\hat{\Cost}(\param))_{(i,j)\in I_2}.
    \] 
    To show that \((\rg_1(\param),\rg_2(\param))\) is non-degenerate it suffices to show that the only vector \((\lambda_\mathbf {i})_{\mathbf i\in I_1\cup I_2}\in \real^{I_1\cup I_2}\) for which the linear combination
    \[
        \sum_{i\in I_1}\lambda_i \partial_{\param_i}\hat{\Cost}(\param)
        + \sum_{(i,j)\in I_2}\lambda_{i,j}\partial_{\param_i}\partial_{\param_j}\hat{\Cost}(\param)
    \]
    has zero variance is \((\lambda_\mathbf {i})_{\mathbf i\in I_1\cup I_2}\equiv 0\). 
    Lemma~\ref{lem: differentiability} ensures that the differentials exist and
    that they can be moved into the inner product defining \(\hat{\Cost}\).
    This implies
    \begin{align*}
        &\var\Biggl(
            \sum_{i\in I_1}\lambda_i \partial_{\param_i}\hat{\Cost}(\param)
            + \sum_{(i,j)\in I_2}\lambda_{i,j} \partial_{\param_i}\partial_{\param_j}\hat{\Cost}(\param)
        \Biggr)
        \\
        &= \var\Biggl(
            \Biggl\langle
            \underbrace{
                \sum_{i\in I_1}\lambda_i\partial_{\param_i}
                \response_\param
                + \sum_{(i,j)\in I_2}\lambda_{i,j}\partial_{\param_i}\partial_{\param_j}
                \response_\param
            }_{
                =:\phi
            },\;
            \rf
        \Biggr\rangle_{\Pr_X}\Biggr).
    \end{align*}
    By weak universality (Definition~\ref{def: standard model}) of the Gaussian
    process \(\rf\) on \(\Domain\) it follows that the latter variance is zero
    if and only if \(\phi\equiv 0\) on the support \(\Domain\) of \(\Pr_X\).
    It is therefore sufficient to prove that
    there exists no non-trivial linear combination of
    \[
        \Bigl(  \nabla \response_\param,
            \bigl(\partial_{\bias_j}^2\response_\param\bigr)_{j\in V_1},
            \bigl(\partial_{\bias_j}\partial_{\weight_{ij}}\response_\param\bigr)_{j\in V_1,i\in V_0},
            \bigl(\partial_{\weight_{ij}}\partial_{\weight_{kj}}\response_\param\bigr)_{j\in V_1, i,k\in V_0, i\le k}
        \Bigr),
    \]
    which is zero on \(\Domain\). We need to ensure that in \(\param\) all
    derivatives \(\partial_{\param_i}\) and
    \(\partial_{\param_i}\partial_{\param_j}\) (\(i\in I_1, (i,j)\in I_2\)) of
    the response \(\response_\param\)  are linearly independent as functions
    on~\(\Domain\). We recall that 
    \[
        \response_\param(x)
        = \bias_{\outSgt}
        + \sum_{j\in V_1}
        \activation\bigl(\bias_j + \langle x, \weight_{\placeholder j}\rangle\bigr)
        \weight_{j\outSgt}
    \]
   and note that we need to ensure linear independence of the following
    derivatives of the response \(\response_\param\)
    \begin{align}
        &x\mapsto 1 
        \tag{\(\partial \bias_\outSgt\)}
        \\
        &x \mapsto \activation\bigl(\bias_j + \langle x, \weight_{\placeholder j}\rangle \bigr)
        && j\in V_1
        \tag{\(\partial \weight_{j\outSgt}\)}
        \\
        &x \mapsto
        \activation'\bigl(\bias_j + \langle x, \weight_{\placeholder j} \rangle\bigr)
        \weight_{j\outSgt}
        && j\in V_1
        \tag{\(\partial \bias_j\)}
        \\
        &x \mapsto \activation'\bigl(\bias_j + \langle x, \weight_{\placeholder j} \rangle\bigr)
        \weight_{j\outSgt}x_i
        &&  j\in V_1,\; i\in V_0 
        \tag{\(\partial \weight_{ij}\)}
        \\
        &x \mapsto \activation''\bigl(\bias_j + \langle x, \weight_{\placeholder j} \rangle\bigr)
        \weight_{j\outSgt}
        && j\in V_1
        \tag{\(\partial \bias_j^2\)}
        \\
        &x \mapsto \activation''\bigl(\bias_j + \langle x, \weight_{\placeholder j} \rangle\bigr)
        \weight_{j\outSgt}x_i
        && i\in V_0, j\in V_1
        \tag{\(\partial \bias_j\partial \weight_{ij}\)}
        \\
        &x \mapsto \activation''\bigl(\bias_j + \langle x, \weight_{\placeholder j} \rangle\bigr) 
        \weight_{j\outSgt}x_i x_k
        && j\in V_1,\; i,k\in V_0
        \tag{\(\partial \weight_{ij}\partial\weight_{kj}\)}
    \end{align}
    Recall that \(\phi=0\) on \(\Domain\) is a linear combination of the derivatives above with the
    prefactors \((\lambda_\mathbf{i})\). To distinguish between the
    types of the indices we write \(\lambda_{\bias_*}\),
    \(\lambda_{\bias_j}\), \(\lambda_{\bias_j,\bias_j}\), \(\lambda_{w_{i,j}}\)
    and so on to refer to the coefficients in front of the respective
    differentials.  Note that in the above list all but the first differential
    can all be assigned to a particular neuron \(j\) in the first layer \(V_1\).
    For a fixed neuron \(j\in V_1\) we denote these contributions to \(\phi\)
    by \(\phi_j\) defined as
        \begin{align}\begin{split}\label{eq: polynomial neuron differentials}
        \phi_j(x)
        := 
            &\underbrace{\lambda_{\weight_{j\outSgt}}}_{=:P_j^{(0)}}
            \activation\bigl(\bias_j + \langle x, \weight_{\placeholder j}\rangle\bigr)
           + \underbrace{\Bigl(\lambda_{\bias_j} + \sum_{i\in V_0} \lambda_{\weight_{ij}}x_i\Bigr)}_{=:P_j^{(1)}(x)}
            \activation'\bigl(\bias_j + \langle x, \weight_{\placeholder j}\rangle\bigr)
            \\
            &+ \underbrace{\Bigl(\lambda_{\bias_j^2}
            + \sum_{i\in V_0}\lambda_{\bias_j\weight_{ij}}x_i
            + \sum_{i,k\in V_0:i\le k}\lambda_{\weight_{ij}\weight_{kj}}x_i x_k\Bigr)}_{=:P_j^{(2)}(x)}
            \activation''\bigl(\bias_j + \langle x, \weight_{\placeholder j}\rangle\bigr).
        \end{split}
    \end{align}  
   Together with \(P^{(\emptyset)}(x):=\lambda_{\bias_\outSgt}\) we therefore obtain 
   \[
        \phi(x)
        = \lambda_{\bias_\outSgt} + \sum_{j\in V_1}\phi_j(x)
        = P^{(\emptyset)}(x)+\sum_{j\in V_1}\sum_{k=0}^2 P_j^{(k)}(x)
        \activation^{(k)}\bigl(\bias_j+\langle x,\weight_{\placeholder j}\rangle\bigr).
    \]
    Since \(\phi(x)=0\) for all \(x\in \Domain\) we can conclude with polynomial
    efficiency of \(\param\) that all the polynomials \(P^{(\emptyset)}\) and
    \(P_j^{(k)}\) (\(j\in V_1,k=0,1,2)\) are zero.
    
    Now inspect the definition of the polynomials in \eqref{eq: polynomial neuron differentials} again.
    In the definition of every polynomial no monomial appears twice so that all
    coefficients have to be equal to zero. Since all coefficients appear in the
    polynomials, indeed all coefficients have to be equal to zero. This finishes
    the proof.
\end{proof}

We will combine the latter statement with a  generalization of a result of
\citet{adlerRandomFieldsGeometry2007}.  It will allow us to conclude that the non-degeneracy of the
distributions of \(\rg(\param):=(\rg_2(\param),\rg_1(\param))\) will allow us to
show that certain ``thin'' events (represented in terms of properties of the
function graph of \(\rg\)) have probability zero. In the final step, we will   define
appropriate ``thin'' events and verify the assumptions of the next lemma.

\begin{lemma}[Generalization {\citep[Lemma~11.2.10]{adlerRandomFieldsGeometry2007}}]
    \label{lem: generalization 11.2.10}
    Let \(\dims,\dims'\in\nat\), \(U\subseteq \real^{d+d'}\) and
    \(W\subseteq\real^\dims\)  be measurable sets. Let \((\rg(w))_{w\in W}\) be
    an \(\real^{\dims'}\)-valued, Lipschitz-continuous stochastic process and suppose
    that for some constants \(C,\rho\in(0,\infty)\) one has for every \((w,v)\in
    U\cap(W\times \real^{\dims'})\) that the distribution of \(\rg(w)\) confined to
    \(B_{\real^{\dims'}}(w,\rho)\) is absolutely continuous w.r.t.\ Lebesgue measure
    with the density being bounded by \(C\). If
    \[
        \mathcal H_{\dims'}(U) =0,
    \]
    then the probability of the graph of \((\rg(w))_{w\in W}\) intersecting \(U\) is zero,
    i.e.
    \[
        \Pr\bigl(\exists w\in W: (w,\rg(w))\in U\bigr) = 0.
    \]
\end{lemma}
\begin{proof}
    Without loss of generality we can assume that \(U\subseteq W\times
    \real^{\dims'}\) (otherwise we replace \(U\) by  \(U\cap( W\times
    \real^{d'})\)).  Let \(L,\epsilon\in(0,\infty)\). Since by assumption
    \(\mathcal H_{d'}(U)=0\) there exist a \(U\)-valued sequence
    \((w_i,v_i)_{i\in\nat}\) and a \((0,\epsilon)\)-valued sequence
    \((r_i)_{i\in\nat}\) such that
    \[
        U\subseteq \bigcup_{i\in\nat} B((w_i,v_i),r_i) \text{ \ and \ } \sum_{i\in\nat} r_i^{d'} \le \epsilon.
    \] 
    Now note that for an arbitrary  Lipschitz function \(g:W\to \real^{\dims'}\)
    with Lipschitz constant \(L\) we have the following: if there exists \(w\in
    W\) with \((w,g(w))\in U\), then there exists an index \(i\in\nat\) with
    \(\|g(w_i)- v_i\|<(1+L)r_i\).  Indeed, in that case there exists an
    index \(i\in\nat\) with \(\|(w,g(w))-(w_i,v_i)\|<r_i\), since balls of
    radius \(r_i\) around \((w_i, v_i) \) cover \(U\), and
    together with the Lipschitz continuity we get that
       \begin{align*}
        \|g(w_i) - v_i\|
        &\le \|g(w_i)- g(w)\| + \|g(w)- v_i\|
        \\
        &\le  L\|w_i - w\| + \|g(w)- v_i\|
        \le (1+L) r_i.
    \end{align*}
    Consequently, we get that for the events
    \[
    \mathbb U=\{\exists w\in W: (w,\rg(w))\in U\}\text{ \ and \ } \mathbb L^{(L)}=\{\rg \text{ is $L$-Lipschitz cont.}\} 
    \]
    one has that
    \[ \mathbb U\cap \mathbb L^{(L)} \subseteq \bigcup_{i\in\nat} \bigl\{ \metric (\rg(w_i), v_i)<(1+L) r_i\bigr\}.
    \]
    Now suppose that \(\epsilon\in(0,\infty)\) was chosen sufficiently small to
    guarantee that \((1+L)\epsilon<\rho\) and conclude using the Lebesgue
    measure \(\lebesgue^{\dims'}\) on \(\real^{\dims'}\) that
    \begin{align*}
        \Pr(\mathbb U\cap \mathbb L^{(L)} )
        &\le   \sum_{i=1}^\infty \Pr\bigl(\rg(w_i)\in B_{(1+L)r_i}(v_i) \bigr)
        \\
        &\le \sum_{i=1}^\infty \int_{ B_{(1+L)r_i}(v_i)} \frac{d\Pr_{\rg(x_i)}}{d\lebesgue^{d'}} d\lebesgue^{d'}
        \\
        &\le C\sum_{i=1}^\infty \lebesgue^{d'}(B_{(1+L)r_i}(v_i))
        \le  C \lebesgue^{d'}(B_{1}(0)) (1+L)^{d'} \epsilon.
    \end{align*}
    By letting \(\epsilon\) go to zero we conclude that \(\Pr(\mathbb U\cap
    \mathbb L^{(L)} )=0\). This is true for every \(L\in(0,\infty)\) and a
    union over rational \(L\) finishes the proof.
\end{proof}

\begin{prop}\label{prop: graph of g is not in U}
    Assume the standard setting (Definition~\ref{def: standard
    model}). Let \(\rg=(\rg(\param))_{\param\in\param}\) be the process given by
    \[
        \rg(\param)=(\rg_2(\param),\rg_1(\param)),
    \]
    where \(\rg_1\) and \(\rg_2\) are as defined in~\eqref{eq: definition of g1}
    and~\eqref{eq: definition of g2}. Let \(\dims,\dims'\in\nat\) be the dimensions of
    \(\param\) and  the target space of \(\rg\). Moreover, let \(U\subset
    \R^{\dims+\dims'}\) be a measurable set
    with
    \[
        \mathcal H_{d'}(U)=0,
    \]
    then 
    \[
        \Pr(\{\exists \param\in\polyEfficient: (\param,\rg(\param))\in U\})=0.
    \] 
\end{prop}

\begin{proof}
    Since \(\polyEfficient\) is an open set in \(\paramSpace\) which is
    separable as a finite dimensional real vector space, we can cover it by a
    countable number of compact balls contained in \(\polyEfficient\).
    Specifically, about any rational point in \(\polyEfficient\) we take a
    closed ball contained in \(\paramSpace\). Then it is sufficient to show
    the claim for any such ball as the countable union of zero sets is still
    a zero set. We thus consider such a compact subset \(W\subseteq
    \polyEfficient\) and aim to show
    \[
        \Pr(\{\exists \param\in W: (\param,\rg(\param))\in U\})=0.
    \]
    Since we have that \(\hat{\Cost}\) has continuous differentials up
    to third degree (Lemma~\ref{lem: differentiability}),
    the process \(\rg\) is continuous as it only consists of first and second
    order differentials (and a continuous mean in the case of \(\rg_1\)). This
    then implies that the covariance kernel of \(\rg\) is continuous
    \citep[e.g.][Theorem~3]{talagrandRegularityGaussianProcesses1987}.
    Moreover the third order differentials are continuous and thereby bounded on
    compact sets, which implies \(\rg\) is almost surely Lipschitz continuous on \(V\) and
    since the covariance kernel of \(\rg\) is continuous, the function
    \[
        \gamma(\param) = \det(\cov(\rg(\param)))
    \]
    is continuous on \(W\) and therefore assumes a minimum in \(W\)
    as \(W\) is compact.  Since the covariance is positive definite, this
    minimum must be greater or equal than zero and by Proposition~\ref{prop:
    non-degenrate g} it cannot be zero since \(W \subseteq
    \polyEfficient\). And since \(\rg(\param)\) is Gaussian by Lemma~\ref{lem:
    differentiability} and assumption on \(\target_\Problem\) its Lebesgue density is bounded by the density at its
    mean given by
    \[
        (2\pi)^{-\dim(\paramSpace)/2} \det(\cov(\rg(\param)))^{-1/2} \le (2\pi)^{-\dim(\paramSpace)/2} (\inf_{\param \in W}\gamma)^{-1/2}=:C< \infty.
    \]
    We can now finish the proof by application of Lemma~\ref{lem: generalization
    11.2.10}.
\end{proof}
\begin{remark}\label{rem: C3 too strong}
    While we have assumed analytic activation functions in the standard
    setting (Definition~\ref{def: standard model}) we only required
    the activation functions to be in \(C^3\) so far.  As we only work with the
    gradient and Hessian even the assumption of \(C^3\) activation functions
    appears too strong. And indeed in the proof above we only used this fact to
    show that \(\rg\) is almost surely Lipschitz.
    \citet{adlerRandomFieldsGeometry2007} highlights a similar issue after the
    proof of their Lemma~11.2.10, which we generalized in Lemma~\ref{lem:
    generalization 11.2.10}. \citet{adlerRandomFieldsGeometry2007}
    proceed to generalize their result using a growth condition on the modulus of
    continuity in place of Lipschitz continuity (cf. Lemma 11.2.11). A similar
    generalization should be possible for Lemma~\ref{lem: generalization 11.2.10}.
    But since we need analytic activation functions anyway for Lemma~\ref{lem: F
    definition} and therefore Theorem~\ref{thm: morse function on efficient
    networks}, we avoid the complications of this generalization.
\end{remark}

\subsection{Proof of Thm~\ref{thm: morse function on efficient networks}}
\label{sec: proof of morse result}


The main task of this section is to prove existence of the function \(F\) announced in the end of the introduction to Section~\ref{sec:Morse}.
We will show the following.

\begin{lemma}[Definition of \(F\)]
    \label{lem: F definition}
    In the standard setting (Definition~\ref{def: standard model})
    let  \(\rg_2=(\rg_2(\param))_{\param\in\param}\) be the
    \(\real^{I_2}\)-valued process as defined in~\eqref{eq: definition of g2},
    with \(I_2\) being the respective index set. Then there exists a non-zero,
    real-analytic function
    \[ 
        F:
            \polyEfficient \times \real^{I_2}
            \to \real,
    \]
    such that for every \(\param\in\polyEfficient =\polyEfficient(\Domain)\) with \(\nabla\Cost(\param) = 0\) we have that
    \[
        \det(\nabla^2 \Cost(\param))
        = F(\param, \rg_2(\param)).
    \]
%
\end{lemma}

Before we prove this lemma, we show that it finishes the proof of
Theorem~\ref{thm: morse function on efficient networks}.
We have that
\begin{align*}
    &\Pr\Bigl(\exists \param \in \polyEfficient: \nabla\Cost(\param) = 0,\; \nabla^2\Cost(\param)=0\Bigr)
    \\
    \overset{\text{Lemma~\ref{lem: F definition}}}&\le 
    \Pr\Bigl(\exists \param \in \polyEfficient: \nabla\Cost(\param) = 0,\; F(\param, \rg_2(\param))=0\Bigr)
    \\
    \overset{\rg_1(\param)=\nabla\Cost(\param)}&=
    \Pr\Bigl(\exists \param \in \polyEfficient: (\param, \rg_2(\param)) \in F^{-1}(0),\; \rg_1(\param) \in \{0\} \Bigr)
    \\
    &= \Pr\Bigl(\exists \param \in \polyEfficient: (\param, \rg(\param)) \in \underbrace{F^{-1}(0)\times \{0\}}_{=: U} \Bigr).
\end{align*}
To apply Proposition~\ref{prop: graph of g is not in U} we only need that \(U\) has
sufficiently small Hausdorff dimension (specifically smaller dimension than
the target space of \(\rg\)). And indeed, since \(F\) is a non-zero,
real-analytic function its zero set is one dimension smaller than the dimension \(d':=\#I_1+\# I_2\) of its domain, see
\citep{mityaginZeroSetReal2020}. This entails that
\[
\mathcal H_{d'}(F^{-1}(0))=0\text{ \ \ and \ \ } \mathcal H_{d'}(U)=0.
\]
By definition, \(d'\) is also the dimension of the target space of \(\rg\)   and Proposition~\ref{prop: graph of g is not in U} entails that
\[
\Pr\bigl(\exists \param \in \polyEfficient: (\param, \rg(\param)) \in U\bigr).
\]
This finishes the proof of Theorem~\ref{thm: morse
function on efficient networks}.

\begin{remark}[Analytic activation]
    Observe that the analytic activation function was only used to ensure
    \(F\) is analytic (cf.~Remark \ref{rem: C3 too strong}). Lemma~\ref{lem: F definition} is
    therefore the appropriate place to search for generalizations.
\end{remark}

\begin{remark}[Efficient is sufficient]
    The function in Lemma~\ref{lem: F definition} can be defined on
    the efficient domain \(\efficient=\efficient(\Domain)\) (cf.~Definition~\ref{def: efficient}),
    i.e.  \(F:\efficient\times\real^{I_2}\to\real\). This is
    a superset of the polynomially efficient domain \(\polyEfficient\)
    in general and coincides with the efficient domain for certain activation
    functions (cf.~Theorem~\ref{thm: characterization of efficient networks}).
    We conduct the proof with \(\efficient\) but readers may replace
    this with \(\polyEfficient\).
\end{remark}
        
\begin{proof}[Proof of Lemma~\ref{lem: F definition}]
    To prove Lemma~\ref{lem: F definition} we need to construct a function \(F\)
    with the following properties
     \begin{enumerate}[label=\text{\normalfont (P\arabic*)}]
        \item\label{item: F is det hessian}
        at any \(\param\in\efficient\) with \(\nabla\Cost(\param) = 0\) we have
        \[
            \det(\nabla^2 \Cost(\param))
            = F(\param, \rg_2(\param)),
        \]
        \item\label{item: F is analytic}
        \(F\) is analytic, and

        \item\label{item: F non-zero}
        \(F\) is non-zero, i.e. there exists an input to \(F\) where \(F\) is non-zero.
    \end{enumerate}
    Recall that we have by definition for some total order on \(V_0\)
    \[
        \rg_2(\param)
        =\Bigl(
            \bigl(\partial_{\bias_j}^2\hat{\Cost}(\param)\bigr)_{j\in V_1},
            \bigl(\partial_{\bias_j}\partial_{\weight_{ij}}\hat{\Cost}(\param)\bigr)_{\substack{j\in V_1\\i\in V_0}},
            \bigl(\partial_{\weight_{ij}}\partial_{\weight_{kj}}\hat{\Cost}(\param)\bigr)_{\substack{j\in V_1\\ i,k\in V_0\\i\le k}}
        \Bigr).
    \]
    In order to be able to plug \(\rg_2\) into \(F\) it must thus be of the form
    \[
        F:
        \efficient \times V_1 \times (V_1\times V_0) \times (V_1\times \text{Sym}(V_0^2)) \to \real,
    \]
    where \(\text{Sym}(V_0^2) = \{(i,k) \in V_0^2 : i\le k\}\).
    To satisfy \ref{item: F is det hessian} we need to reconstruct the
    determinant of the Hessian \(\nabla^2\Cost(\param)\) on the basis of the differentials in  \(\rg_2(\param)\) (that originate from the Hessian \(\nabla^2\hat{\Cost}(\param)\)). 
    First recall that by Proposition~\ref{prop: decomposition}, one has that
    \begin{align*}
        \det(\nabla^2 \Cost(\param))
        &= \det\Bigl(\nabla^2 \Bigl[\regularizer(\param) + \|\response_\param\|_{\Pr_X}^2 - 2\hat{\Cost}(\param) + \E_\Problem[\|Y\|^2]\Bigr]\Bigr)
        \\
        &= \det\bigl(\nabla^2[\regularizer(\param)+ \|\response_\param\|_{\Pr_X}^2] - 2\nabla^2\hat{\Cost}(\param)\bigr).
    \end{align*}
    Since \(\param\mapsto \nabla^2[\regularizer(\param) +
    \|\response_\param\|_{\Pr_X}^2]\) is a deterministic function, we can absorb
    it into the definition of \(F\) and
    construct a function \(\tilde{F}\) that reproduces
    \(\nabla^2\hat{\Cost}(\param)\) from \(\rg_2(\param)\). That is, assuming we had
    a function \(\tilde{F}\) with \(\tilde{F}(\param, \rg_2(\param)) =
    \nabla^2\hat{\Cost}(\param)\) in all critical points, we can define
    \[
        F(\param, x)
        := \det\bigl(\nabla^2[\regularizer(\param) + \|\response_\param\|_{\Pr_X}^2] - 2\tilde{F}(\param, x)\bigr)
    \]
    If all entries of the matrix valued function \(\tilde{F}\) are analytic
    functions it is therefore sufficient for all entries of \(\param\mapsto\nabla^2 [\regularizer(\param) +
    \|\response_\param\|_{\Pr_X}^2]\) to be analytic for \(F\) to be
    analytic, because the determinant is a sum and product of these analytic
    components.
    And by the assumption that \(X\) has compact support in the
    standard setting (Definition~\ref{def: standard model}) and Theorem~5.1 in
    \citet{dereichConvergenceStochasticGradient2024}, \(\param\mapsto
    \|\response_\param\|_{\Pr_X}^2\) is analytic, and
    so are its differentials.
    \ref{item: F is analytic} thus follows if all entries of
    \(\tilde{F}\) are analytic since we assumed \(\response\) to be analytic in
    the standard setting (Definition~\ref{def: standard model}).

    Our strategy is therefore to show that all entries/differentials of
    \(\nabla^2\hat{\Cost}(\param)\) fall into one of the following categories:
    \begin{enumerate}[label=\text{\normalfont (\alph*)}]
        \item \label{item: regular}
        the partial differential is contained in  \(\rg_2\) in which case  the respective matrix entry in \(\tilde{F}\) is identical to the related input  (analytic)
        \item\label{item: second order is zero}
        the partial differential is zero (analytic), or
        \item\label{item: deterministic functions}
        there is an (analytic) deterministic function of \(\param\) (that we
        still need to construct) such that the partial differential coincides
        with the function value whenever
        \(\nabla\Cost(\param)=0\).
    \end{enumerate}
    
    To enact this strategy, we now consider all the second order derivatives in
    \(\nabla^2\hat{\Cost}(\param)\) that are not contained in \(\rg_2\) and
    categorize them into \ref{item: second order is zero} or \ref{item:
    deterministic functions}. Recall that by Lemma~\ref{lem: differentiability}
    \[
        \hat{\Cost}(\param)
        = \langle \response_\param, \rf\rangle_{\Pr_X},
        \quad
        \nabla \hat{\Cost}(\param)
        = \langle \nabla\response_\param, \rf\rangle_{\Pr_X},
        \quad
        \nabla^2 \hat{\Cost}(\param)
        = \langle \nabla^2\response_\param, \rf\rangle_{\Pr_X}.
    \]
    We therefore need to consider the derivatives of the
    response function. Since the response \(\response_\param\) is essentially a
    weighted sum over the index set \(V_1\) with all parameters appearing only
    in one of the summands, we have that second order partial derivatives that
    belong to two different neurons \(j\neq l\) of the hidden layer \(V_1\)
    vanish. This then directly implies that these differentials also vanish for
    \(\hat{\Cost}\).
    Specifically, we have that
    \begin{align*}
        \partial_{\bias_j}\partial_{\bias_l} \response_\param &= 0
        &&\forall j \neq l\in V_1
        \\
        \partial_{\bias_j}\partial_{\weight_{il}}\response_\param &= 0
        &&\forall j \neq l\in V_1,\; \forall i\in V_0
        \\
        \partial_{\weight_{ij}}\partial_{\weight_{kl}}\response_\param &= 0
        &&\forall j \neq l\in V_1,\; \forall i,k\in V_0 
        \\
        \partial_{\weight_{ij}}\partial_{\weight_{l\outSgt}}\response_\param &= 0
        &&\forall j \neq l\in V_1,\; \forall i\in V_0
        \\
        \partial_{\weight_{ij}}\partial_{\bias_\outSgt}\response_\param &= 0
        &&\forall j\in V_1,\; \forall i\in V_0
        \\
        \partial_{\bias_j}\partial_{\bias_\outSgt} \response_\param &= 0
        &&\forall j \in V_1
    \end{align*}
    Let us refer to ``inner parameters'' as the parameters that appear inside the activation
    functions (the connections from the input layer to the hidden layer and
    the biases of neurons in the hidden layer). All second order
    derivatives of these inner parameters are either included in \(\rg_2\)
    or mix derivatives of two different hidden neurons and therefore vanish.
    All second order differentials with respect to inner parameters
    (exclusively) are thus of type \ref{item: regular} or \ref{item: second
    order is zero}.

    Since the response is linear in the outer (remaining)
    parameters \(\weight_{j\outSgt}\) and \(\bias_\outSgt\) taking two
    derivatives in these direction also results in zero, i.e.,
     \begin{align*}
        \partial_{\beta_{\outSgt}}^2 \response_\param = 0,
        \quad \partial_{\beta_{\outSgt}}\partial_{\weight_{j\outSgt}} \response_\param = 0
        \qquad \text{and}\qquad
        \partial_{\weight_{j\outSgt}}^2 \response_\param = 0
        \quad \forall j \in V_1.
       \end{align*}
    Most derivatives are thus in category \ref{item: second order is zero}.

    The only derivatives left are therefore the mixtures of outer derivatives
    \(\weight_{j\outSgt}\) with inner derivatives \(\bias_j\) and
    \(\weight_{ij}\) of the same hidden neuron \(j\in V_1\). These will be in category
    \ref{item: deterministic functions}. For those observe:
    \[
        \partial_{\weight_{ij}} \response_\param(x)
        = \weight_{j\outSgt}
        \activation'\bigl(
            \bias_j
            + \langle x, \weight_{\placeholder j}\rangle
        \bigr) x_i.
        \]
    Using that \(w_{j\outSgt}\) is non-zero as \(\param \in \efficient\) we get that
    \[
        \partial_{\weight_{j\outSgt}} \partial_{\weight_{ij}} \response_\param(x) =   \activation'\bigl(
            \bias_j
            + \langle x, \weight_{\placeholder j}\rangle
        \bigr) x_i
        =\tfrac 1{w_{j\outSgt}} 
        \partial_{\weight_{ij}}
        \response_\param(x).
    \]
    Since we are allowed to move differentiation into the
    inner products by Lemma \ref{lem: differentiability} we get
    \[
        \partial_{\weight_{j\outSgt}}\partial_{\weight_{ij}}
        \hat{\Cost}(\param)
        =
        \bigl\langle
            \partial_{\weight_{j\outSgt}} \partial_{\weight_{ij}} \response_\param(x),
            \rf
        \bigr\rangle
        = \tfrac 1{\weight_{j\outSgt}}
        \bigl\langle
            \partial_{\weight_{ij}} \response_\param(x),
            \rf
        \bigr\rangle= \tfrac 1{\weight_{j\outSgt}}
        \partial_{\weight_{ij}}
        \hat{\Cost}(\param).
    \]
    Now recall by Proposition~\ref{prop: decomposition} we have that
    \begin{equation}
        \label{eq: gradient decomposition} 
        \nabla\Cost(\param)
        = \nabla_\param[\regularizer(\param) +\|\response_\param\|_{\Pr_X}^2] - 2\nabla\hat{\Cost}(\param)
    \end{equation}
    so that in every critical point \(\param\) of \(\Cost\) with \(\nabla\Cost(\param)=0\)
    we get that
    \[
        \frac1{2\weight_{j\outSgt}} \partial_{\weight_{ij}}[
            \regularizer(\param) + \|\response_\param\|_{\Pr_X}^2
        ]
        = \frac1{\weight_{j\outSgt}} \partial_{\weight_{ij}}\hat{\Cost}(\param)
        = \partial_{\weight_{j\outSgt}}\partial_{\weight_{ij}}
        \hat{\Cost}(\param).
    \]
    For the definition of \(\tilde F\) we therefore use the deterministic
    function
    \[
        \param \mapsto \frac1{2\weight_{j\outSgt}}
        \partial_{\weight_{ij}}[\regularizer(\param) +
        \|\response_\param\|_{\Pr_X}^2]
    \]
    for the component where the differential
    \(\partial_{\weight_{j\outSgt}}\partial_{\weight_{ij}}\) should be.
    
      We proceed in complete analogy with the differential \(\partial_{\weight_{j\outSgt}}\partial_{\bias_j}\) and obtain that for every  critical efficient parameter \(\param\)
    \[
        \partial_{\weight_{j\outSgt}}\partial_{\bias_j}
        \hat{\Cost}(\param)
        = \frac{1}{2\weight_{j\outSgt}}\partial_{\bias_j}
        [\regularizer(\param)+ \|\response_\param\|_{\Pr_X}^2]
    \]
    and we use the respective deterministic function to define the corresponding component of \(\tilde F\).
    
	Note that  \(\param \mapsto \|\response_\param\|_{\Pr_X}^2\) is
    analytic, the deterministic functions that we use as substitutes for the
    remaining second order differentials in \(\tilde F\) are therefore analytic
    on \(\efficient\) (where \(\weight_{j\outSgt} \neq 0\)). Thus we constructed
    a function \(F\) satisfying properties
    \ref{item: F is det hessian} and \ref{item: F is analytic}.
    
    It remains to show  that \(F\) is a non-zero function \ref{item: F
    non-zero}.   
    To show this we will arrange the second order differentials appropriately.     
    We put the outer parameters
    \(\gamma := (\bias_\outSgt, (\weight_{j\outSgt})_{j\in V_1})\)
    at the end of the vector. Recall that all these components fall into
    category \ref{item: second order is zero} so that, in particular,
    \[
        \nabla_\gamma^2[\regularizer(\param) + \|\response_\param\|_{\Pr_X}^2] -2 \tilde F_\gamma(\theta,x)
        = \nabla_\gamma^2 [\regularizer(\param) + \|\response_\param\|_{\Pr_X}^2],
    \]
    where \(F_\gamma\) is \(F\) restricted to the output components with pairs from the \(\gamma\) coordinates.

    Let the remaining inner parameters be given by
    \[
        \alpha := \Bigl(
            (\bias_j)_{j\in V_1},
            (\weight_{ij})_{\substack{i\in V_0\\j\in V_1}}
        \Bigr).
    \]
    Recall that the second order differential with respect to two parameters
    from \(\alpha\) belongs to category \ref{item: regular} or \ref{item: second
    order is zero} and that the diagonal belongs to \ref{item: regular}. The
    diagonal can therefore be fully controlled and the other entries are either
    zero naturally or can be set to zero. Thus, for every parameter
    \(\param\in\efficient\) and any given \(\lambda \in \real\) we can find
    $x_\lambda$ with
    \[
        \tilde F_\alpha(\theta,x_\lambda)
        =  -\tfrac\lambda2 \identity,
    \]
    where \(\identity\) is the identity matrix and \(F_\alpha\) is \(F\)
    restricted to the output components with pairs from the \(\alpha\)
    coordinates. Consequently, for this choice of \(x\) we have that
    \[
        \nabla^2 \|\response_\param\|_{\Pr_X}^2-2 \tilde F(\theta,x_\lambda)
        =\begin{pmatrix}
            \nabla_\alpha^2[\regularizer(\param) + \|\response_\param\|_{\Pr_X}^2] + \lambda \identity & B(\param)\\
            B(\param)^\transpose & \nabla_\gamma^2[\regularizer(\param) + \|\response_\param\|_{\Pr_X}^2]
        \end{pmatrix}.
    \]
    Marked with a \(B(\param)\) are the mixed derivatives with parameters from
    \(\alpha\) and \(\gamma\). These are either of type \ref{item: second order
    is zero} or \ref{item: deterministic functions} and in particular they 
    are functions of \(\param\). More than \(F\) being non-zero, we will
    show for any fixed \(\param\) there exists \(\lambda\) and thus
    \(x_\lambda\) such that \(F(\param, x_\lambda) \neq 0\). For this note
    that \(F(\param, x_\lambda)\) is given by the determinant of the equation
    above and the determinant of a block matrix is given by
    \[
        \det\begin{bmatrix}
            A & B\\
            B^\transpose & D
        \end{bmatrix}
        = \det(D)\det(A - BD^{-1}B^\transpose)
    \]
    We will show that \(D := D(\param):=\nabla_\gamma^2 [\regularizer(\param)+ \|\response_\param\|_{\Pr_X}^2]\) is a
    strictly positive definite matrix and by doing so we show that it has full
    rank and therefore non-zero determinant. Since the eigenvalue \(\lambda\) of
    \(\lambda \identity\) can be directly controlled with the selection of
    \(x_\lambda\), selecting a sufficiently large \(\lambda \gg 0\) ensures
    that the eigenvalues of
    \[
        A- BD^{-1}B^\transpose
        = \lambda \identity + \underbrace{\nabla_\alpha^2[\regularizer(\param) + \|\response_\param\|_{\Pr_X}^2]
        - B(\param) D^{-1}B(\param)}_{\text{some matrix}}
    \]
    are all strictly positive. Thus we can ensure the second determinant is
    non-zero.
    
    What is left is thus the proof that \(\nabla_\gamma^2[\regularizer(\param)+
    \|\response_\param\|_{\Pr_X}^2]\)
    strict positive definite. Since \(\regularizer(\param)\) is assumed to
    be convex \(\nabla_\gamma^2\regularizer(\param)\) is positive semi-definite.
    It is thus sufficient to prove strict positive definiteness for
    \(\nabla_\gamma^2\|\response_\param\|_{\Pr_X}^2\).
    
    Let \(c\in \real^{\# V_1+1 }\) and recall \(\gamma=(\bias_\outSgt,
    (\weight_{j\outSgt})_{j\in V_1})\in\real^{\# V_1+ 1}\). Using that the
    iterated differentials from \(\gamma\) belong to category \ref{item: second
    order is zero} we conclude that
    \begin{align*}
       c^\transpose\nabla_{\gamma}^2\|\response_\param\|_{\Pr_X}^2c
       &= \sum_{i,j=0}^{\#V_1} c_i c_j \partial_{\gamma_i}\partial_{\gamma_j}\|\response_\param\|_{\Pr_X}^2
       \\
       &= \sum_{i,j=0}^{\# V_1} c_i c_j \partial_{\gamma_i}\partial_{\gamma_j}
       \int \response_\param^2(x) \,\Pr_X(dx)
       \\
       &= \int\sum_{i,j=0}^{\# V_1} c_i c_j \partial_{\gamma_i}\partial_{\gamma_j} \response_\param(x)^2 \,\Pr_X(dx).
       \\
       \overset{\partial_{\gamma_i}\partial_{\gamma_j}\response_\param \equiv 0}&= 2 \int\sum_{i,j=0}^{\# V_1} c_i c_j (\partial_{\gamma_i}\response_\param(x))(\partial_{\gamma_j} \response_\param(x))\, \Pr_X(dx).
       \\
       &= 2 \Bigl\|\sum_{i=0}^{\# V_1}c_i \partial_{\gamma_i} \response_\param\Bigr\|_{\Pr_X}^2.
    \end{align*}
   Consequently, \(\nabla_\gamma^2\|\response_\param\|_{\Pr_X}^2\) is always 
    positive definite. 
   To see strict positive definiteness we analyze solutions \(c\) for which the latter norm is zero.
    This requires that
    \begin{equation}
        \label{eq: sum of weighted outer parameter derivatives}    
        \sum_{i=0}^{\# V_1} c_i \partial_{\gamma_i} \response_\param = 0,
        \qquad \text{\(\Pr_X\)-almost surely.}
    \end{equation}
    Identifying \(V_1\) with \(\{1,\dots, \# V_1\}\) this implies that
    \[
        \partial_{\gamma_0} \response_\param
        = \partial_{\bias_\outSgt} \response_\param = 1
        \quad\text{and}\quad 
        \partial_{\gamma_i} \response_\param
        = \partial_{\weight_{i\outSgt}} \response_\param
        = \activation\bigl(\bias_i + \langle \weight_{\placeholder i}, x\rangle\bigr),
    \]
    for every \(i\in V_1\), 
   so that (\ref{eq: sum of weighted outer parameter derivatives}) implies that 
    \[
        c_0 +\sum_{i=1}^{\# V_1} c_i \,\activation\bigl(\bias_i + \langle \weight_{\placeholder i}, x\rangle\bigr)= 0
        \qquad \text{\(\Pr_X\)-almost surely.}
   \]
   Since the term above is continuous in \(x\) it is thus zero for all \(x\) in
   the support \(\Domain\) of \(\Pr_X\).
   Efficiency (Definition~\ref{def: efficient}) of \(\param\) then implies that
   \(c\equiv0\) is the unique solution of~(\ref{eq: sum of weighted outer
   parameter derivatives}).
    This proves that
    \(\nabla_\gamma^2\|\response_\param\|_{\Pr_X}^2\) is strictly positive
    definite and therefore finishes the proof of \ref{item: F non-zero}.
\end{proof}

%% file: sections/efficient_param_characterization.tex
\section{Characterization of the efficient domain}
\label{sec: characterization of efficient params}

In the following \(\network=(\mathbb V,\activation)\) is a fixed ANN.
We start with a more natural axiomatic definition of efficient parameters
than the polynomial efficiency we required for our main result.

\begin{definition}[Efficient and redundant parameters] 
	\label{def: efficient}
	A parameter \(\param=(\weight,\bias)\) is called
	\emph{efficient}, if 
	\begin{enumerate}[label={(\alph*)}]
		\item \label{it: disused neuron}
		all neurons are used meaning that for all \(k\in V_1\) one has
        \[
			\weight_{k \placeholder} = (\weight_{kl})_{l\in V_\tout} \not\equiv 0,
		\]

		\item \label{it: linear combination of neurons}
		the equation
	\[
		\lambda_\emptyset + \sum_{j\in V_1} \lambda_j
		\activation\Bigl(\bias_j + \sum_{i\in V_0} x_i\weight_{ij}\Bigr)
		= 0 
        \qquad \forall x\in \Domain
    \]
	has the unique solution \(\lambda_j = 0\) for all \(j \in V_1 \uplus \{\emptyset\}\).
	\end{enumerate}
	We denote by \(\efficient=\efficient(\Domain)\) the \emph{efficient domain}, which is the set of
	all parameters \(\param=(\weight, \bias)\) that are efficient.	A
	parameter that is not efficient is called \emph{redundant}.
\end{definition}

In the remark below we introduce categories of redundant parameters
and hope to convey the intuition of this definition of `efficiency'.

\begin{remark}[Taxonomy of redundant parameters]
    \label{rem: taxonomy of redundant parameters}
    A parameter can be redundant for various reasons:
    If property \ref{it: disused neuron} does not hold we call it
    a \emph{deactivation redundancy} since the output of the hidden neuron
    \(k\) is ignored.
    If the property \ref{it: linear combination of
    neurons} does not hold there exists a neuron which can be
    linearly replicated by other
    neurons meaning that  there exists a neuron \(k\in V_1\) and  \(\lambda_j\in \real\) for all \(j\in V_1\uplus \{\emptyset\}\) with
    \[
        \activation\Bigl(\bias_k + \sum_{i\in V_0}  x_i\weight_{ik} \Bigr)
        = \lambda_\emptyset + \sum_{j\in V_1\setminus \{k\}}\lambda_j
        \activation\Bigl(\bias_j + \sum_{i\in V_0} x_i\weight_{ij}\Bigr)
        \quad \forall x\in \Domain.
    \]
    If  \(\lambda_\emptyset\) is the only \(\lambda_j\) not equal to zero in
    this linear combination, we speak of a \emph{bias redundancy}, as the
    neuron \(k\) is constant like the bias (typically \(\weight_{\placeholder k}
    = 0\), cf. Lemma~\ref{lem: bias redundancy characterization}). If there is another neuron \(j\in V_1\) such that \((\bias_j,
    \weight_{\placeholder j}) = (\bias_k, \weight_{\placeholder k})\)
    the neuron \(k\) can be trivially linearly combined from the others and we speak of a
    \emph{duplication redundancy}. We call all other cases a \emph{generalized duplication redundancy}.
    While deactivation, bias and duplication redundancies occur independently of
    the activation function (cf.~Lemma~\ref{lem: general redundancies}),
    generalized duplication redundancies only occur due to symmetries of
    the activation function (cf. Lemma \ref{lem: sigmoid sign symmetric
    redundancy}, Lemma \ref{lem: tanh sign symmetric redundancy} and Example
    \ref{ex: softplus}).
\end{remark}

Obviously, a configuration is $(0,0)$-polynomially efficient if and only if it
is efficient in the sense of Definition~\ref{def: efficient}. But in general the
assumption of polynomial efficiency is stronger. 

As we will show next both notions generally coincide in the case where the
activation function is either \(\sigmoid\) or \(\tanh\)! Further we will show
that they also coincide with the explicit set representation \eqref{eq:
defintion efficient 0}.

\begin{theorem}[Characterization of efficient networks for \(\sigmoid\) and \(\tanh\)]
    \label{thm: characterization of efficient networks}
    Assume that in the considered ANN \(\network\) the activation function
    \(\activation\) is either  \(\sigmoid\) or \(\tanh\), further assume
    that \(\Domain\) contains an open set. Then for every \(n\in\nat_0\) and
    \(m=(m_{\emptyset}, m_0, \dots, m_n)\in\nat_0^{n+2}\) one
    has
    \[
        \efficient(\Domain)=\polyEfficient^m(\Domain)=\efficient_0,
    \]
   where, as in \eqref{eq: defintion efficient 0},
    \[
        \efficient_0
        := \Biggl\{
            \param =(\weight, \bias)\in \real^{E \times (V\setminus V_0)}:
            {
            \begin{aligned}[c]
                \scriptstyle \weight_{j\placeholder}
                &\scriptstyle\neq 0
                &&\scriptstyle \forall j \in V_1,
                \\[-0.5em]
                \scriptstyle\weight_{\placeholder j}
                &\scriptstyle\neq 0
                &&\scriptstyle\forall j\in V_1, 
                \\[-0.5em]
                \scriptstyle(\weight_{\placeholder i}, \bias_i)
                &\scriptstyle\neq \pm(\weight_{\placeholder j}, \bias_j)
                &&\scriptstyle\forall i,j\in V_1\text{ with }i\not=j
            \end{aligned}
            }
        \Biggr\}.
    \]
\end{theorem}
\begin{proof}[Proofsketch]
    Since the activation function is either \(\sigmoid\)
    or \(\tanh\) and therefore real-analytic, the equations in the definition of
    the efficient set (Definition~\ref{def: efficient}) and the definition of
    polynomial independence (Definition~\ref{def: polynomial independence}) are
    real-analytic. Since analytic functions which are zero on an open
    set are zero anywhere, the open set contained in \(\Domain\) therefore
    ensures that we can assume without loss of generality \(\Domain =
    \real^{V_\tin}\)

    Suppressing \(\Domain\) in the notation of \(\efficient\) and
    \(\polyEfficient^m\) the proof is then established by showing that
    \(\efficient\subseteq \efficient_0\subseteq \polyEfficient^m\subseteq
    \efficient\). Note that \(\polyEfficient^m\subseteq \efficient\) is trivial
    as polynomials can always chosen to be constant.
\begin{itemize}
	\item
	To prove ``\(\efficient \subseteq \efficient_0\)'' we will show that any
	parameter \(\param\notin \efficient_0\) is not in \(\efficient\). More
	explicitly, we will construct a redundancy and show that one of the
	properties \ref{it: disused neuron} or \ref{it:  linear combination of
	neurons} does not hold.
    
    \begin{remark}
    As part of this proof in Subsection~\ref{subsec: explicit open set}, we will
    prove that \(\efficient\) is always contained in
    \begin{equation}
       \label{eq: bound efficient} 
        \bar{\efficient}
        :=\Biggl\{
            \param =(\weight, \bias)\in \real^{E \times (V\setminus V_0)}:
            {
            \begin{aligned}[c]
                \scriptstyle \weight_{j\placeholder}
                &\scriptstyle\neq 0
                &&\scriptstyle \forall j \in V_1,
                \\[-0.5em]
                \scriptstyle\weight_{\placeholder j}
                &\scriptstyle\neq 0
                &&\scriptstyle\forall j\in V_1, 
                \\[-0.5em]
                \scriptstyle(\weight_{\placeholder i}, \bias_i)
                &\scriptstyle\neq (\weight_{\placeholder j}, \bias_j)
                &&\scriptstyle\forall i\neq j\in V_1
            \end{aligned}
            }
        \Biggr\}
    \end{equation}
    regardless of the activation function \(\activation\) (Lemma~\ref{lem:
    general redundancies}). With a counterexample (Example~\ref{ex: softplus})
    we further show that the \emph{sign-symmetric redundancy} \((\weight_{\placeholder
    i}, \bias_i) = -(\weight_{\placeholder j}, \bias_j)\) is specific to the
    activation functions \(\activation\in \{\sigmoid, \tanh\}\). This shows
    that the explicit definition of \(\efficient_0\) does not generalize well.
    \end{remark}
    
    \item
    It will be harder to prove that ``\(\efficient_0\subseteq \polyEfficient^m\)''
    and we will first consider the case with one dimensional input (i.e.
    \(\# V_\tin =1\)) in Subsection~\ref{subsec: polynomial independence, 1-dim}.
    This proof relies on the complex poles of the meromorphic activation
    function \(\activation\in\{\sigmoid, \tanh\}\). We will then show that
    the one-dimensional result also implies the general result in
    Subsection~\ref{subsec: polynomial independence, general case}.
    
    \begin{remark}
        For the transfer from the one-dimensional result to the general result we
        we do not make use of the assumption \(\activation\in\{\sigmoid,
        \tanh\}\). We only use that the result holds for the \(1\)-dimensional case.
        This suggests that this part of the proof should be transferrable to other
        activation functions except for the fact that \(\efficient_0\) may be different
        for other activation functions in general and we use the specific structure
        of \(\efficient_0\).
    \qedhere
    \end{remark}
\end{itemize}
\end{proof}

\subsection{Proof of \texorpdfstring{\(\efficient\subseteq\efficient_0\)}{E ⊆ E0}}
\label{subsec: explicit open set}

Take \(\param\notin \efficient_0\), then it is sufficient to prove this
parameter to be not efficient, i.e. \(\param\notin\efficient\).
For this we are going to consider the possible constraints of \(\efficient_0\)
the parameter \(\param\) can violate and match them with the types of redundancies
we classified in Remark~\ref{rem: taxonomy of redundant parameters}. Recall
\[
    \efficient_0
    \overset{\text{def}}= \Biggl\{
        \param =(\weight, \bias)\in \real^{E \times (V\setminus V_0)}:
        {
        \begin{aligned}[c]
            \scriptstyle \weight_{j\placeholder}
            &\scriptstyle\neq 0
            &&\scriptstyle \forall j \in V_1,
            \\[-0.5em]
            \scriptstyle\weight_{\placeholder j}
            &\scriptstyle\neq 0
            &&\scriptstyle\forall j\in V_1, 
            \\[-0.5em]
            \scriptstyle(\weight_{\placeholder i}, \bias_i)
            &\scriptstyle\neq \pm(\weight_{\placeholder j}, \bias_j)
            &&\scriptstyle\forall i,j\in V_1\text{ with }i\not=j
        \end{aligned}
        }
    \Biggr\}.
\]
For \(\theta\notin \efficient_0\) one of the inequalities must be violated.
We consider all possibilities:

\begin{enumerate}
    \item \emph{Deactivation redundancy:}
    \label{it: deactivation redundancy} 
    If there is a neuron \(j\in V_1\) such that \(\weight_{j \placeholder} = 0\),
    then the parameter \(\param\) has a deactivation redundancy (Remark~\ref{rem: taxonomy of redundant parameters})
    as the output of the neuron is ignored and the parameter violates
    requirement \ref{it: disused neuron} of Definition~\ref{def: efficient}. Thus \(\param\notin \efficient\).

    \item \emph{Bias redundancy:}
    \label{it: bias redundancy}
    There is a neuron \(k\in V_1\) such that \(\weight_{\placeholder k} = 0\).
    Since this implies that the output of neuron \(k\) is constant
    and equal to \(\activation(\bias_k)\) irrespective of the input \(x\) it
    falls into the category of bias redundancies (Remark~\ref{rem: taxonomy of
    redundant parameters}). Specifically, the realization function can be
    replicated by removing the neuron \(k\) and adjusting the bias. This
    allows for a non-trivial linear combination
    \begin{equation}
        \label{eq:lin_indep1}    
    	\lambda_\emptyset + \sum_{j\in V_1} \lambda_j
		\activation\Bigl(\bias_j + \sum_{i\in V_0} x_i\weight_{ij}\Bigr)
		= 0 
    \end{equation}
    with
    \[
     \lambda_j=\begin{cases}
        - \psi(\beta_k), &\text{ if } j=\emptyset,\\
        1, &\text{ if } j=k,\\
        0, &\text{ if } j\in V_1\backslash\{k\}.
    \end{cases}
    \]
    This is a violation of \ref{it: linear combination of neurons} from
    Definition~\ref{def: efficient} and we thus have \(\param\notin \efficient\).

    \item \emph{Duplication redundancy:}
    \label{it: duplication redundancy}
    There are two neurons \(k,\ell\in V_1\) such that their parameters are equal
    \((\weight_{\placeholder k}, \bias_k) = (\weight_{\placeholder \ell},
    \bias_\ell)\). We call this a duplication redundancy since both neurons have
    identical parameters and therefore identical output. Here
    \[
        \lambda_j=\begin{cases}
            1, &\text{ if } j=k,\\
            -1,&\text{ if } j=\ell,\\
            0,&\text{ if } j\in \{0\}\cup(V_1\backslash\{k,\ell\})
        \end{cases}
    \]
    defines a nontrivial solution for \eqref{eq:lin_indep1}, violating \ref{it: linear combination of neurons}
    of Definition~\ref{def: efficient}. Thus \(\param\not\in \efficient\).
    Note that deactivation redundancies fall into the category of `generalized
    deactivation redundancies' in Remark~\ref{rem: taxonomy of redundant
    parameters}. 
\end{enumerate}

Observe that so far, we have not made use of \(\activation\in\{\sigmoid,
\tanh\}\). This leads to the following upper bound on the set of
efficient parameters regardless of the activation function.

\begin{lemma}[General redundancies]
    \label{lem: general redundancies}
    Let \(\network=(G, \activation)\) with \(G=(V,E)\) be a shallow
    neural network. Then the set of efficient parameters \(\efficient\)
    satisfies
    \[
        \efficient
        \subseteq\Biggl\{
            \param =(\weight, \bias)\in \real^{E \times (V\setminus V_0)}:
            {
            \begin{aligned}[c]
                \scriptstyle \weight_{j\placeholder}
                &\scriptstyle\neq 0
                &&\scriptstyle \forall j \in V_1,
                \\[-0.5em]
                \scriptstyle\weight_{\placeholder j}
                &\scriptstyle\neq 0
                &&\scriptstyle\forall j\in V_1, 
                \\[-0.5em]
                \scriptstyle(\weight_{\placeholder i}, \bias_i)
                &\scriptstyle\neq (\weight_{\placeholder j}, \bias_j)
                &&\scriptstyle\forall i\neq j\in V_1
            \end{aligned}
            }
        \Biggr\}
        =: \bar \efficient.
    \]
\end{lemma}
\begin{proof}
    The general arguments \ref{it: deactivation redundancy}, \ref{it: bias redundancy} and \ref{it: duplication redundancy}
    imply \(\bar{\efficient}^\complement \subseteq \efficient^\complement\).
\end{proof}

Continuing with our proof of \(\efficient\subseteq \efficient_0\)
there is one possible constraint violation left for \(\param\notin \efficient_0\):

\begin{enumerate}[resume]
    \item \emph{Sign-symmetric redundancy:}
    \label{it: sign-symmetric redundancy}
    There are two neurons \(i,j\in V_1\) such that \((\weight_{\placeholder i},
    \bias_i) = -(\weight_{\placeholder j}, \bias_j)\). The reason this results in
    a redundancy are symmetries of the activation function \(\activation\).
    Details in Lemma~\ref{lem: sigmoid sign symmetric redundancy} and \ref{lem: tanh sign symmetric redundancy}.
\end{enumerate}

\begin{lemma}[\(\sigmoid\)]
    \label{lem: sigmoid sign symmetric redundancy}
    Let \(\activation=\sigmoid\) and let $\param$ be a parameter
    such that there exist \(k,\ell\in V_1\) with \((\weight_{\placeholder k},
    \bias_k) =- (\weight_{\placeholder \ell}, \bias_\ell)\). Then \(\param\)
    is not efficient.
\end{lemma}

\begin{proof}
    Observe that we have for all \(x\in\real\) that
    \begin{equation}
        \label{eq: sigmoid sign symmetry}
        \sigmoid(-x)
        = \frac{e^{-x}}{1+ e^{-x}}
        = \frac{1+e^{-x}}{1+e^{-x}} - \frac{1}{1+e^{-x}}
        = 1 - \sigmoid(x).
    \end{equation}
   This allows for a non-trivial linear combination 
    \[
    		\lambda_0 + \sum_{j\in V_1} \lambda_j
		\activation\Bigl(\bias_j + \sum_{i\in V_0} x_i\weight_{ij}\Bigr)
		= 0 
    \]
    via
    \[
        \lambda_j= \begin{cases}
            1 , &\text{ if } j\in\{k,\ell\},\\
            0,&\text{ if } j\in V_1\backslash\{k,\ell\},\\
            -1,&\text{ if } j=\emptyset.
        \end{cases}
    \]
    This nontrivial solution violates \ref{it: linear combination of neurons}
    of Definition~\ref{def: efficient} and thus implies \(\param\not\in \efficient\).
\end{proof}

\begin{lemma}[\(\tanh\)]
    \label{lem: tanh sign symmetric redundancy}
    Let \(\activation=\tanh\) and let $\param$ be a parameter
    such that there exist \(k,\ell\in V_1\) with \((\weight_{\placeholder k},
    \bias_k) =- (\weight_{\placeholder \ell}, \bias_\ell)\). Then \(\param\)
    is not efficient.
\end{lemma}
\begin{proof}
    Observe that we have for every \(x\in\real\)
        \begin{equation}
        \label{eq: tanh sign symmetry}    
        \tanh(-x) = \frac{e^{-x} - e^{x}}{e^{-x} + e^x} = - \tanh(x).
    \end{equation}
    Similarly, to the proof of Lemma~\ref{lem: sigmoid sign symmetric redundancy}
    we use this symmetry to construct a non-trivial linear combination via
     \[
        \lambda_j=\begin{cases}
            1 , &\text{ if } j\in\{k,\ell\},\\
            0,&\text{ if } j\in \{\emptyset\}\cup V_1\backslash\{k,\ell\}
        \end{cases}
    \]
    violating \ref{it: linear combination of neurons} of Definition~\ref{def: efficient}
    and finishing the proof.
\end{proof}

The redundancy that is treated in the Lemmas \ref{lem: sigmoid sign symmetric redundancy}
and \ref{lem: tanh sign symmetric redundancy} is caused by certain
symmetries in the particular activation function. In general, the particular
structure of the activation function can cause complex additional redundancies.

\begin{example}[Softplus]
    \label{ex: softplus}
    Consider the case of the softplus activation
    function \(\activation(x) = \ln(1+\exp(x))\). If we have \((\weight_{\placeholder i}, \bias_i) =
    -(\weight_{\placeholder j}, \bias_j)\), then
    \begin{align*}
        \activation\bigl(\bias_i + \langle x,  \weight_{\placeholder i}\rangle \bigr)
        &= \ln\bigl(1+\exp\bigl[-\bigl(\bias_j + \langle x, \weight_{\placeholder j}\rangle\bigr)\bigr]\bigr)
        \\
        &= -\bigl(\bias_j + \langle x,\weight_{\placeholder j}\rangle \bigr)
        + \ln\bigl(1+\exp\bigl[\bias_j + \langle x, \weight_{\placeholder j}\rangle\bigr]\bigr)
        \\
        &= \bigl(\bias_i + \langle x, \weight_{\placeholder i}\rangle \bigr)
        + \activation\bigl(\bias_j + \langle x,\weight_{\placeholder j}\rangle\bigr).   
    \end{align*}
    So the neurons are equal up to a linear term, which can be cancelled out by another
    sign pair \((\weight_{\placeholder k},\bias_k) = -(\weight_{\placeholder l},
    \bias_l)\) and appropriate \(\weight_{k\placeholder}\) and
    \(\weight_{l\placeholder}\). For this it is important to keep in mind that
    we only need to take care of the first order term, as the zero order term
    can be absorbed by the bias \(\bias_m\) for \(m\in V_2\).

    Pruned networks in the case of the softplus function may therefore
    include one sign symmetry, but not more. The set of efficient parameters is
    therefore slightly larger. On the other hand polynomial efficiency
    results in the set \(\efficient_0\) again if \(P^{(\emptyset)}\) in Definition~\ref{def:
    polynomial independence} may be of degree \(1\) (i.e. \(m_\emptyset \ge 1\))
    as the linear term can be absorbed by a polynomial.
\end{example}

\subsection{Proof of \texorpdfstring{\(\efficient_0\subseteq \polyEfficient^m\)}{E ⊆ EmP} in the case of one dimensional input} \label{subsec: polynomial independence, 1-dim}

Let \(\param\in\efficient_0\). We need to show \(\param\) to be
\(m\)-polynomially independent (Definition~\ref{def: polynomial independence}).
In this first step we assume one dimensional input, i.e. \(V_\tin = \{\inSgt\}\).

For ease of notation, we define \(\alpha_j := \weight_{\inSgt j}\) for all \(j\in V_1\)
and write with slight misuse of notation \(x := x_{\inSgt}\). We thus have to prove that if the representation
\begin{equation}
    \label{eq:lindep2}
    0 = P^{(\emptyset)}(x)
    + \sum_{j\in V_1}
    \sum_{k=0}^n P^{(k)}_j(x) \activation^{(k)}\bigl(\bias_j + \alpha_j x\bigr) 
\end{equation}
is true for all \(x\in\real\),\footnote{
    Recall that we could translate \(x\in \Domain\) to \(x\in \real^{V_\tin}\) without
    loss of generality since the activation functions \(\tanh\) and
    \(\sigmoid\) are analytic and Theorem~\ref{thm: characterization of
    efficient networks} assumes an open set is contained in \(\Domain\).
} then 
all polynomials \(P^{(k)}_j\)  in the equation are zero. We will use complex analysis to show this.
Recall that the activation functions we consider are given by
\begin{align*}
    \sigmoid(x) &= \frac1{1+e^{-x}}\\
    \tanh(x) &= \frac{e^x -e^{-x}}{e^x + e^{-x}} = \frac{1- e^{-2x}}{1+ e^{-2x}}.
\end{align*}
Both activation functions are given as quotients of entire functions and are thus
meromorphic functions on \(\{x\in \complex: e^{-x}\not=-1\}\) and \(\{x\in
\complex: e^{-2x}\not=-1\}\), respectively. The singularities form
an infinite chain on the imaginary axis of the form
\[
    z_m = cm i \qquad 
    (m\in \integer_\mathrm{odd}),
\]
where \(\integer_\mathrm{odd}\) denotes the odd integers and \(c=\pi\) in the
case of \(\sigmoid\) and \(c=\frac12\pi\) in the case of \(\tanh\).  

For every neuron $j\in V_1$ and $k=0,\dots,n$ the singularities of the meromorphic function
\[
    x\mapsto \psi^{(k)}(\alpha_j x+\beta_j)
\]
are of the form 
\[
    z_m^{(j)}
    =\frac{cm i-\beta_j}{\alpha_j}\qquad(m\in\integer_\mathrm{odd}).
\]
In particular, the singularities $S_j:=\{z_m^{(j)}:m\in\mathbb Z_\mathrm{odd}\}$ do not depend on $k$ and we call them the singularities of  neuron $j$. Consequently, the right hand side of (\ref{eq:lindep2}) always defines a meromorphic function on \[\complex\backslash\bigcup_{j\in V_1} S_j.\]
In order to have equality in (\ref{eq:lindep2}), in particular, all singularities have to be liftable.

Now suppose we are given a nontrivial solution to (\ref{eq:lindep2}). 
We will show that this would cause a contradiction as consequence of the following principles that we will state in detail and prove below:

\begin{enumerate}[label={\arabic*)}]
    \item \label{it: active neurons}
    We call a neuron \(j\in V_1\) \emph{active}, if one of the polynomials
    \(P_j^{(k)}\) with $k\in\{0,\dots,n\}$ is nonzero. If the set of
    active neurons is non-empty, there is an active neuron \(j^*\in V_1\) such
    that infinitely many of its singularities in \(S_{j^*}\) are not ``served''
    by another active neuron meaning that the singularity does not lie in one of
    the singularity sets of the other active
    neurons.
    
    \item\label{it: non-liftable singularity}
    If \(j\in V_1\), \(z\in S_j\) and \(k\in\{0,\dots,n\}\) with
    \(P_j^{(k)}(z)\not=0\), then the merophormorphic function
    \[
        \sum_{k=0}^n P_j^{(k)}(x) \psi^{(k)}(\alpha_j x+\beta_j)
    \]
    has a non-liftable singularity in \(z\).
\end{enumerate}
Indeed, the two facts then almost immediately imply a contradiction if a
non-trivial solution to (\ref{eq:lindep2}) would exist: we fix \(j^*\) as in
\ref{it: active neurons} and observe that for infinitely many \(z\in S_{j^*}\),
the singularity \(z\) is not served by other neurons. This entails that for all
these \(z\)

\[
    x\mapsto \sum_{k=0}^n P_{j^*}^{(k)}(x) \psi^{(k)}(\alpha_{j^*} x+\beta_{j^*})
\]
has  a liftable singularity in \(z\). Now \ref{it: non-liftable singularity} implies that for all these \(z\) and \(k=0,\dots,n\), \(P_{j^*}^{(k)}(z)=0\). Since there is no polynomial that satisfies this for an infinite number of points \(z\), we produced a contradiction showing that $j^*$ is not active. 


It remains to prove the two statements. 
\begin{lemma}Let \(\theta\in\efficient_0\) and suppose there exists a nontrivial solution to (\ref{eq:lindep2}). There is an active neuron \(j^*\in V_1\) (meaning that there exists $k\in\{0,\dots,n\}$ with $P_k^{(j^*)}\not\equiv0$) such that 
\[ \# \Bigl(S_{j^*}\Big\backslash\bigcup_{j\text{ active}:\ j\not=j^*}S_j\Bigr)=\infty.
\]
\end{lemma}

\begin{proof}First note that if none of the neurons $j\in V_1$ would be active then one has that $0\equiv P^{(\emptyset)}$ which would imply that the solution is trivial.

Now fix an active neuron $j^*\in V_1$ that maximizes $|\alpha_j|$ over all active neurons~$j$ and denote by \(V_1^*\) the set of all active neurons with 
\[
 \frac{\beta_j}{\alpha_j}=\frac{\beta_{j^*}}{\alpha_{j^*}}.
\]
If there were another \(j\in V_1^*\backslash \{v^*\}\) with \(|\alpha_j|=|\alpha_{j^*}|\), then
we get together with 
\(  \frac{\beta_j}{\alpha_j}=\frac{\beta_{j^*}}{\alpha_{j^*}}
\)
 that \((\alpha_j,\beta_j)\) equals either \((\alpha_{j^*},\beta_{j^*})\) or \((-\alpha_{j^*},-\beta_{j^*})\). But this is not possible since \(\param\in\efficient_0\). Hence, for all \(j\in V_1^*\backslash \{v^*\}\) we have that \(|\alpha_j|<|\alpha_{j^*}|\).

Now let \(j\in V_1^*\backslash \{v^*\}\). Then 
 $z_m^{(j^*)}$ with \(m\in\integer_\mathrm{odd}\) is in $S_j$ iff there exists \(m'\in\integer_\mathrm{odd}\) with
\(m'/\alpha_j=m/\alpha_{j^*}\) or, equivalently,
\begin{align}\label{eq123}
\frac{ \alpha_{j^*}}{\alpha_{j}}m'=m.
\end{align} 
This entails that whenever \(\alpha_{j^*}/\alpha_{j}\) is not a rational we have that $S_j\cap S_{j^*}=\emptyset$. Now suppose that \(\alpha_{j^*}/\alpha_{j}\) is a rational
and let $p\in\integer$ and $q\in\nat$ such that the  representation  \(\alpha_{j^*}/\alpha_{j}=p/q\) is minimal (meaning that $|p|$ and $q$ are minimal). Since \(|\alpha_{j^*}|>|\alpha_{j}|\) we have that \(|p|\ge 2\). The integers $p$ and $q$ have no common divisor and equation  (\ref{eq123}) can only be true if $m'$ is a multiple of \(q\). Consequently, for all but at most one odd prime number \(m\) (namely \(|p|\)) we have that
\(z_m^{(j^*)}\not \in S_j\). Since there are only finitely many neurons in $V_1^*$ we conclude that all but finitely many odd primes \(m\) correspond to singularities \(z_m^{(j^*)}\) of neuron \(j^*\) that are not served by other active neurons.  
\end{proof}

\begin{lemma}
    \label{lem: retaining singularities}
    Let \(\alpha \in\real\backslash\{ 0\}\), \(\bias\in \real\) and \(z^*\in\complex\) be a singularity
    of a meromorphic function \(\activation\).  Let \(m\in\nat\) and for every  \(k\in \{0,\dots,m\}\) let
    \(P^{(k)}:\complex\to\complex\) be a polynomial such that either \(P^{(k)}
    \equiv 0\) or \(P^{(k)}\bigl(\frac{z^* - \bias}\alpha\bigr)\neq 0\).
    Then the polynomial
    \[
        \Phi(x) = \sum_{k=0}^m P^{(k)}(x) \activation^{(k)}(\alpha x + \bias),
    \]
     has a non-liftable singularity in  \(\frac{z^* - \alpha}\bias\) or
     \(P^{(k)} \equiv 0\) for all \(k=0,\dots,m\).
\end{lemma}
\begin{proof}
    By defining
    \[
        \tilde{\Phi}(y) := \Phi\Bigl(\frac{y-\bias}\alpha\Bigr)
        = \sum_{k=0}^m P^{(k)}\Bigl(\frac{y-\bias}\alpha\Bigr) \activation^{(k)}(y),
    \]
    and absorbing \(\alpha,\bias\) into the polynomials we can assume without loss of generality
    that \(\alpha=1\) and \(\bias=0\). Since these polynomials are zero if and only if
    the modified polynomials are zero.

    As a meromorphic function, all singularities are poles, i.e., there exists
    a smallest order \(r\) such that
    \[
        (z-z^*)^{r} \activation(z) 
    \]
    can be continuously extended in \(z^*\). Moreover its Laurent series is then
    of the form
    \[
        \activation(z) = \sum_{l=-r}^\infty a_l (z-z^*)^l 
    \]
    with \(a_{-m}\neq 0\). This representation allows us to argue that the order of the
    pole increases with every derivative and therefore we cannot cancel out the singularities
    with a simple linear combination. So either we retain the singularity, or we have to
    set everything to zero.
    More specifically, we have
    \begin{align*}
        \activation^{(k)}(z)
        &= \sum_{l=-r}^\infty a_ll\cdots(l-k+1) (z-z^*)^{l-k}
        \\
        &= \sum_{l=-(r+k)}^\infty a_{l+k}(l+k)\cdots(l+1) (z-z^*)^l
    \end{align*}
    Let us assume that \(P^{(m)}\neq 0\), then we have \(P^{(m)}(z^*)\neq 0\) and thus
    \begin{align*}
        &\Bigl|(z-z^*)^{r+m-1}\Phi(z)\Bigr|
        \\
        &\ge 
            \Biggl|
                \underbrace{|(z-z^*)^{r+m-1}P^{(m)}(z) \activation^{(m)}(z)|}_{\to \infty}
                - \underbrace{\Bigl|(z-z^*)^{r+m-1}\sum_{k=0}^{m-1}P^{(k)}(z) \activation^{(k)}(z)\Bigr|}_{\to c \in \real}
            \Biggr|
        \\
        &\to \infty \qquad (z\to z^*).
    \end{align*}
    Therefore \(\Phi\) has a singularity at \(z^*\) (more specifically a pole of
    order at least \(r+m\)).
    If \(P^{(m)} \equiv 0\), we repeat the same argument with \(m-1\) until we have
    either a pole at \(z^*\) or \(P^{(m)} \equiv \dots \equiv P^{(0)} \equiv 0\).
\end{proof}

\subsection{Proof of \texorpdfstring{\(\efficient_0\subseteq \polyEfficient^m\)}{E ⊆ E}, general case}
\label{subsec: polynomial independence, general case}

In this section we reduce the proof of \(\efficient_0 \subseteq \polyEfficient^m\)
to the one dimensional result we have already proven in Subsection~\ref{subsec:
polynomial independence, 1-dim}. For an element \(\param \in \efficient_0\)
recall that this requires that the equation
\[
    0 = P^{(\emptyset)}(x)
    + \sum_{j\in V_1}
    \sum_{k=0}^n P^{(k)}_j(x) \activation^{(k)}\bigl(\bias_j + \langle x, \weight_{\placeholder j}\rangle \bigr),
    \quad \forall x\in \real^{V_\tin}
\]
has the unique solution of all polynomials being zero (Definition~\ref{def:
polynomial independence}). Since this equation holds for all inputs \(x\in
\real^{V_\tin}\), it holds in particular for \(1\)-dimensional slices
\[
    x_v(\lambda):= \lambda v, \qquad \lambda\in \real
\]
for directions \(v\in \real^{V_\tin}\). The following proof hinges on the fact
that the mappings
\[
    \lambda \mapsto  P_j^{(k)}(x_v(\lambda))
\]
are polynomials in \(t\), which we can prove to be zero with the one dimensional
result. If sufficiently many appropriate directions \(v\) are chosen
and all directional polynomials are zero, then Theorem~\ref{thm: optimal
polynomial slicing} allows us to deduce that the original polynomial is zero.
But to apply the one dimensional result, we need to ensure that we do not
introduce degeneracies with the directions we choose. For example
\[
    \lambda\mapsto \langle x_v(\lambda), \weight_{\placeholder j}\rangle = \lambda \langle v, \weight_{\placeholder j}\rangle
\]
may be constantly zero if the direction \(v\) is orthogonal to
\(\weight_{\placeholder,j}\). This introduces a bias redundancy. For the number
of directions
\(
    N= \binom{\max(m) +\, \#V_\tin -1}{\max(m)}
\)
we therefore want to select \(v^{(1)}, \dots, v^{(N)}\in \real^{V_\tin}\) such that
the following conditions hold at the same time:
\begin{enumerate}
    \item the directions \(v^{(l)}\) characterize polynomials in the sense of
    Theorem~\ref{thm: optimal polynomial slicing}. This is required
    for us to deduce that the original polynomials are zero.
    \item \(\alpha_k^{(l)} := \langle v^{(l)}, \weight_{\placeholder k}\rangle  \neq 0\) for all \(k\in V_1\).

    \item \((\alpha_i^{(l)}, \bias_i) \neq \pm (\alpha_j^{(l)}, \bias_j)\) for all \(i, j\in V_1\) with \(i\neq j\).
\end{enumerate}
The last two requirements ensure non-redundancy for the parameters of
the surrogate neural networks.

\paragraph*{Why is this possible?} Select iid entries \(v^{(l)}_i\sim\normal(0,1)\)
with \(l\in \{1,\dots,N\}\) and \(i\in V_\mathrm{in})\). Then we
satisfy the conditions of Theorem~\ref{thm: optimal polynomial slicing} and
almost surely have the first condition. Since \(\weight_{\placeholder k} \neq 0\)
for all \(k\in V_1\) by assumption, we also have almost surely
\[
    \alpha_k^{(l)} = \langle v^{(l)}, \weight_{\placeholder k} \rangle \neq 0.
\]
That is we have the second condition almost surely. For the last condition,
we use the assumption \((\weight_{\placeholder i}, \bias_i)\neq
\pm(\weight_{\placeholder j}, \bias_j)\). Let us only consider
the case where ``\(\pm\)'' is ``\(+\)''. The other
case is analogous. Then by assumption, we have
\[
        (\weight_{\placeholder i}, \bias_i) - (\weight_{\placeholder j}, \bias_j) \neq 0
\]
and thus either \(\weight_{\placeholder i}-\weight_{\placeholder j} \neq 0\) (case 1)
or \(\bias_i - \bias_j \neq 0\) (case 2).  This implies
\[
    (\alpha_i^{(l)}, \bias_i) - (\alpha_j^{(l)}, \bias_j) 
    = \bigl(
        \langle v^{(l)}, \underbrace{
            \weight_{\placeholder i} - \weight_{\placeholder j}
        }_{\overset{\text{case 1}}{\neq} 0}
        \rangle,
        \underbrace{\bias_i - \bias_j}_{\overset{\text{case 2}}{\neq} 0}
    \bigr)\neq 0.
\]
Note that due to the iid selection of the entries of \(v^{(l)}\) the inequality of the
tuple with zero only holds almost surely in case 1.

In summary, with the selection of random \(v^{(l)}\) we can satisfy all three conditions
almost surely. In particular, there \emph{exist} directions \(v^{(1)}, \dots, v^{(N)}\)
which satisfy all three conditions simultaneously.

\paragraph*{\(1\)-dimensional slices of the \(\# V_\tin\)-dimensional input}

For the network \(\network = (\mathbb V, \activation)\)  we consider the \(1\)-dimensional network
\(\tilde{\network} := (\tilde{\mathbb{V}}, \activation)\) whose input layer is reduced to a
single node
\[
    \tilde{\mathbb{V}} := (\{\inSgt\}, V_1, V_\tout).
\]
From parameters \(\param=(\weight, \bias)\) of \(\network\) and direction \(v^{(l)}\) we
construct parameters \(\param^{(l)}=(\weight^{(l)}, \bias)\) of \(\tilde{\network}\) by retaining
the bias \(\bias\) and all connections from the hidden layer \(V_1\) to the
output \(V_\tout\) that is \(\weight^{(l)}_{\placeholder k} :=
\weight_{\placeholder k}\) for all \(k\in V_\tout\). For the input layer
connections, we set
\[
    \weight^{(l)}_{\inSgt j} := \alpha_j^{(l)} = \langle v^{(l)}, \weight_{\placeholder j}\rangle
    \qquad \forall j\in V_1.
\]
Then \(\param^{(l)} \in \efficient_0\) since we have
\begin{align*}
    \weight^{(l)}_{\inSgt i} &\neq 0 \quad \forall i\in V_1
    &&\text{due to } \alpha_j^{(l)} \neq 0,\\
    \weight^{(l)}_{i\placeholder } &\neq 0 \quad \forall i\in V_1
    &&\text{due to } \weight_{i \placeholder}^{(l)} = \weight_{i \placeholder} \neq 0,\\
    (\weight^{(l)}_{\inSgt i}, \bias_i) &\neq \pm (\weight^{(l)}_{\inSgt j}, \bias_j) \quad \forall i\neq j\in V_1
    &&\text{due to } (\alpha_i^{(l)}, \bias_i) \neq \pm (\alpha_j^{(l)}, \bias_j).
\end{align*}

\begin{remark}[Response function slices]
    The response functions \(\response_{\param^{(l)}}\) of the new parameter
    \(\param^{(l)}\) has the following relation with the response
    \(\response_\param\)
    \[
        \response_{\param^{(l)}}(\lambda)
        = \response_\param(\lambda v^{(l)})
        = \response_\param(x_{v^{(l)}}(\lambda))
        \qquad \forall \lambda \in \real.
    \]
\end{remark}

\paragraph*{Using the slices for the proof of polynomial independence}
We are now finally ready to prove polynomial independence for multi-dimensional
input. To do so, assume we have
\[
    0 = P^{(\emptyset)}(x)
    + \sum_{j\in V_1}
    \sum_{k=0}^m P^{(k)}_j(x) \activation^{(k)}\Bigl(\bias_j + \langle x, \weight_{\placeholder j}\rangle \Bigr) 
    \quad \forall x\in \real^{V_\tin}.
\]
In particular we can select \(x = \lambda v^{(l)}\) to obtain
\[
    0 = P^{(\emptyset)}(\lambda v^{(l)})
    + \sum_{j\in V_1}
    \sum_{k=0}^m P^{(k)}_j(\lambda v^{(l)}) \activation^{(k)}\Bigl(\bias_j + \weight^{(l)}_{\inSgt j} \lambda \Bigr) 
    \quad \forall \lambda\in \real.
\]
By the polynomial independence of the \(1\)-dimensional input network slices
\(\network^{(l)}\), we then have that the \(1\)-dimensional polynomial
slices
\[
    \lambda \mapsto P_j^{(k)}(\lambda v^{(l)})
\]
are all identically zero. As we selected the directions \(v^{(1)},\dots, v^{(N)}\)
to characterize polynomials in the sense of Theorem~\ref{thm: optimal
polynomial slicing}, we thus have \(P_j^{(k)}\equiv 0\) for all \(j\in V_1\) and
all \(k=\bias,0,\dots, n\). That is, polynomial independence for the case of
multivariate input.

\subsection{Polynomial slicing}

The main tool to translate the \(1\)-dimensional input result to the general
case are slices of polynomials that characterize the full polynomial.
This is formalized in the following theorem, which is proven in the remainder
of this section.

\begin{theorem}[Optimal polynomial slicing]
    \label{thm: optimal polynomial slicing}
    Let \(\dims, n\in \nat\) and \(N=\binom{n+\dims -1}{n}\).
    
    \begin{enumerate}[label=\text{\normalfont(\Roman*)}]
        \item \label{item: almost all selections work}
        \textbf{Almost all selections of directions \(v_1,\dots,v_N\) characterize the \(\dims\)-variate polynomials of degree \(n\).} If the
        matrix \((v_1,\dots, v_N)\in \real^{\dims\times N}\) is selected
        randomly with a density with respect to the Lebesgue measure on
        \(\real^{\dims\times N}\) (in particular there exist
        such \(v_i\)), then the following property holds almost surely:
        
        For any \(\dims\)-variate polynomial \(p\in
        \real[x_1,\dots,x_\dims]\) of order \(n\) we have \(p\equiv 0\)
        if and only if all slices in the directions \(v_i\) are zero, i.e.
        \[
            p_{v_i}(\lambda) := p(\lambda v_i) = 0 \quad \forall \lambda \in \real,\quad\forall i=1,\dots,N.
        \]

        \item \label{item: optimality of number of directions}
        \textbf{The number \(N\) of directions is optimal.} That is,
        for any smaller selection of directions \(v_1,\dots, v_M\in\real^\dims\) with \(M<N\)
        there exists a \textbf{non-zero} \(\dims\)-variate polynomial
        \(p\in \real[x_1,\dots,x_\dims]\) of order \(N\) such that all the
        slices \(p_{v_i}\)
        are identically zero.

    \end{enumerate}
\end{theorem}

A crucial object in the proof of this theorem is the vector of all monomials of
degree \(n\). In the multivariate case, the definition of such a vector requires
an ordering of the tuple of powers.
\begin{definition}[Vector of monomials]
    \label{def: vector of monomials}
    We define
    \[
        \mon_{n}(x) = \Biggl(
            \prod_{i=1}^\dims x_i^{r_i}
        \Biggr)_{r\in R} \quad\text{for}\quad x\in \real^\dims
    \]
    where \(R\) is a subset of \(\dims\)-tuples of non-negative integers
    that form monomials of exactly degree \(n\)
    \[
        R = \Bigl\{
            r \in \nat_0^\dims : \sum_{i=1}^\dims r_i = n
        \Bigr\}.
    \]
    With the following injection into the ordered set of
    non-negative integers \(\nat_0\)
    \[
        \phi : \begin{cases}
            R \to \nat_0
            \\
            r \mapsto \sum_{i=1}^{d}r_{i}(n+1)^{i-1},
        \end{cases}
    \]
    we equip \(R\) with the pullback of this order.  That is we define \(r <
    \tilde{r}\) if and only if \(\phi(r) < \phi(\tilde{r})\).
    In other words, we assume \(R\) has reverse lexicographic ordering.

    This ensures \(\mon_n(x)\) to be a vector and not just an unordered set.
    \end{definition}

\begin{prop}[Independent monomials]
    \label{prop: independent monomials}
    Let \(\dims, n\in \nat\) and \(N=\binom{n+\dims -1}{n}\).
    Then there exist \(v_1,\dots, v_N\) such that \(\mon_n(v_i)\) are linearly
    independent.

    Almost all selections of \(v_1,\dots, v_N\) have this property. That is,
    if the directions \((v_1,\dots, v_N)\in \real^{\dims\times N}\) are
    random variables with a density with respect to the Lebesgue measure
    on \(\real^{\dims\times N}\), then \(\mon_n(v_i)\) are almost surely
    linearly independent.
\end{prop}

\begin{proof}
    Using \(0< a_1<\dots < a_N\) with \(a_i\in \real\),
    we define
    \[
        v_k
        = \Bigl(a_k, a_k^{n+1}, \dots, a_k^{(n+1)^{\dims-1}}\Bigr)
    \]
    Then we have by definition of \(v_k\), \(\mon_n\) and \(\phi\)
    \[
        \mon_n(v_k)
        = \Biggl(\prod_{i=1}^\dims (v_k^{(i)})^{r_i}\Biggr)_{r\in R}
        = \Biggl(\prod_{i=1}^\dims a_k^{r_i(n+1)^{i-1}}\Biggr)_{r\in R}
        = (a_k^{\phi(r)})_{r\in R}.
    \]
    Therefore we have
    \begin{equation}
        \label{eq: generalized vandermonde}
        (\mon_n(v_1),\dots, \mon_n(v_N))
        = \begin{pmatrix}
            a_1^{\lambda_1} & \cdots & a_N^{\lambda_1}
            \\
            \vdots & & \vdots
            \\
            a_1^{\lambda_N} & \cdots & a_N^{\lambda_N}
        \end{pmatrix},
    \end{equation}
    with \((\lambda_1, \dots, \lambda_N) = (\phi(r))_{r\in R} \subseteq \nat_0\), where
    the size of the set \(R\) is given by Lemma~\ref{lem: number of monomials} and we
    have \(\lambda_1 < \dots < \lambda_N\) by the ordering defined for \(R\)
    in Definition~\ref{def: vector of monomials}. But the matrix
    \eqref{eq: generalized vandermonde} is a generalized Vandermonde
    matrix as in Lemma~\ref{lem: det generalized vandermonde} and its
    determinant is thus not equal to zero by Lemma~\ref{lem: det generalized vandermonde}. The
    monomial vectors are therefore linearly independent.

    Observe that \(\det(\mon_n(v_1),\dots, \mon_n(v_N))\) is a multivariate
    polynomial in the entries of \(v_i\). In particular it is a (real) analytic
    function. By \citet{mityaginZeroSetReal2020} the zero set of a real analytic
    which is not identically zero is a Lebesgue zero set. Since we found an
    example above where this determinant is non-zero, we ruled out that the
    function is identically zero. Thus almost all selections of \(v_1,\dots,
    v_N\) result in a non-zero determinant.
\end{proof}

\begin{lemma}[Number of monomials]
    \label{lem: number of monomials}
    \(|R| = N = \binom{n+\dims -1}{n}\)
\end{lemma}
\begin{proof}
    The number \(N\) is also sometimes referred to as ``\(\dims\) multichoose \(n\)''
    and denoted by
    \[
        \left(\!\!\binom{\dims}{n}\!\!\right)
        = \binom{n+\dims -1}{n}
    \]
    as it is equal to the number of ways to create a multiset of size \(n\) from
    \(\dims\) elements. In our case, we are picking an \(n\)-sized multiset of
    \(x_i\) to finally multiply all elements together to obtain a monomial.
    Details of the proof can be found in textbooks such as
    \citet[25-26]{stanleyEnumerativeCombinatoricsVolume2011} or \citet[Sec.
    3.2]{riordanIntroductionCombinatorialAnalysis2002}
\end{proof}

\begin{lemma}[Generalized Vandermonde]
    \label{lem: det generalized vandermonde}
    Let \(0< a_1<\dots< a_N\) for \(a_i \in \real\) and \(\lambda_1<\dots<\lambda_N\)
    with \(\lambda_i\in \real\). Then we have that
    the following generalized Vandermonde matrix has non-zero determinant, i.e.
    \[
        \det\begin{pmatrix}
            a_1^{\lambda_1} & \cdots & a_N^{\lambda_1}
            \\
            \vdots & & \vdots
            \\
            a_1^{\lambda_N} & \cdots & a_N^{\lambda_N}
        \end{pmatrix} \neq 0.
    \]
\end{lemma}
\begin{proof}
    The proof is adapted from a stack exchange answer \citep{szwarcHowProveThat2022}.
    We conduct an induction over \(N\) and note that for the induction start
    \(N=1\) the conclusion obviously holds.

    For the induction step \(N-1 \to N\), assume that there exist
    \(c_1,\dots, c_N\in \real\) such that the rows weighted
    by \(c_i\) of the generalized Vandermonde sum to zero, i.e. we have for all 
    columns \(k\)
    \[
        c_1 a_k^{\lambda_1} + \dots + c_N a_k^{\lambda_N} = 0.
    \]
    In order to prove linear independence of these rows and thus that
    the determinant is zero, we only need to show that these equations
    imply \(c_i =0\) for all \(i=1,\dots, N\).

    Dividing the equations above by \(a_k^{\lambda_1}\), we observe
    that the \(a_k\) are zeros of the function
    \[
        f(x) = c_1 + c_2 x^{\lambda_2 - \lambda_1} + \dots + c_N x^{\lambda_N-\lambda_1}
    \]
    Since \(\lambda_k - \lambda_1>0\) by assumption, \(f\) is a continuously
    differentiable function. Between any two points where \(f\) is zero
    there is therefore a point where its derivative
    \[
        f'(x)
        = c_2(\lambda_2 - \lambda_1)x^{\lambda_2 - \lambda_1-1}
        + \dots
        + c_N(\lambda_N - \lambda_1)x^{\lambda_N - \lambda_1 - 1}
    \]
    is zero by the mean value theorem. In the gaps of \(a_1 <  \dots < a_n\) are thus
    \(N-1\) points \(0< u_1< \dots, u_{N-1}\) such that \(f'\) is zero at all \(u_i\).
    Since by induction hypothesis we have
    \[
        \det\begin{pmatrix}
            u_1^{\tilde{\lambda}_1} & \cdots & u_{N-1}^{\tilde{\lambda}_1}
            \\
            \vdots & & \vdots
            \\
            u_1^{\tilde{\lambda}_{N-1}} & \cdots & u_N^{\tilde{\lambda}_{N-1}}
        \end{pmatrix} \neq 0
    \]
    for \(\tilde{\lambda}_i = \lambda_{i+1} - \lambda_1 -1\), we have that the
    rows of this matrix are linearly independent. And since
    we have that all \(u_k\) are zeros of \(f'\), we have for
    the weighted colum sums
    \[
       0 = c_2(\lambda_2-\lambda_1) u_k^{\tilde{\lambda}_1}+ \dots + c_N(\lambda_N-\lambda_1) u_k^{\tilde{\lambda}_{N-1}}
    \]
    for all \(k\). By linear independence of the rows this implies
    that the coefficients \(c_i(\lambda_i-\lambda_1)\) for \(i\ge 2\) have  to
    be be zero. Since \(\lambda_i -\lambda_1 >0\), this
    implies \(c_2 = \dots = c_N = 0\). We thus have
    \(f \equiv c_1\) and since the \(a_k\) are zeros of \(f\)
    this also implies \(c_1=0\). Thus all \(c_i\) are zero which is what we needed to prove. 
\end{proof}

\subsubsection*{Proof of polynomial slicing (Theorem~\ref{thm: optimal polynomial slicing})}

\begin{enumerate}[wide]
    \item[\ref{item: almost all selections work}. ] 
    For the proof of the first statement of the theorem we intend to use the
    directions \(v_1,\dots, v_N\) of Proposition~\ref{prop: independent
    monomials} which result in linearly independent monomials. Note that these
    are only monomials of \emph{exactly} degree \(n\), while the polynomials of
    degree \(n\) admit all monomials \emph{up to} degree \(n\).

    We address this difference with an induction over the degree and by sorting
    the monomials of the polynomials into buckets with the same degree. The
    induction step is enabled by the following lemma.

    \begin{lemma}
        If the monomials \(\mon_m(v_1), \dots, \mon_m(v_N)\) span the space,
        then the lower degree monomials \(\mon_k(v_1), \dots, \mon_k(v_N)\) also
        span the space for all degrees \(k\leq m\).
    \end{lemma}
    \begin{proof}
        Without loss of generality assume \(k=m-1\).  Let \(K\) be the length of
        \(\mon_k(x)\) and \(M\) be the length of \(\mon_m(x)\) for \(x\in
        \real^\dims\). Choose any \(y\in \real^K\).
        We now have to prove that there is a linear combination of \(\mon_k(v_i)\)
        equal to \(y\).

        Observe that the vector
        \begin{equation}
            \label{eq: monomial subset}
            x_1 \mon_k(x)
            = x_1 \mon_{m-1}(x) \qquad x \in \real^\dims
        \end{equation}
        contains a subset of the entries of \(\mon_m(x)\). We obtain the vector
        \(\tilde{y} \in \real^M\) from \(y\in \real^K\) by setting the positions
        of all other entries to zero. Since \(\mon_m(v_1), \dots, \mon_m(v_N)\)
        span the space, there exists a linear combination
        \[
            \tilde{y} = \sum_{i=1}^N c_i \mon_m(v_i).
        \]
        By the observation \eqref{eq: monomial subset} this implies
        \[
            y = \sum_{i=1}^N \underbrace{c_i v_i^{(1)}}_{=:\tilde{c_i}}\mon_k(v_i).
        \]
        Thus the \(\mon_k(v_i)\) span the space.
    \end{proof}

    With this lemma we can now finish the proof of the first statement of the theorem.
    As already mentioned we take the directions \(v_1,\dots, v_N\) of
    Proposition~\ref{prop: independent monomials} and note that with this lemma
    we have that \(\mon_m(v_1),\dots, \mon_m(v_N)\) span the space for all
    \(m\leq n\). We proceed by induction over \(m\) up to \(n\). That is,
    we assume that the polynomial \(p\) is of degree \(m\) and assume that
    \(\mon_m(v_1),\dots, \mon_m(v_N)\) span the space but are not necessarily
    linearly independent (this is only the case for \(m=n\)).

    The base case \(m=0\) is trivially true as one direction is enough to figure out
    if a constant polynomial is zero.

    For the induction step \(m-1 \to m\) let \(p\) be a \(\dims\)-variate
    polynomial of degree \(m\). We decomposition the polynomial into
    \[
        p(x) = \sum_{k=0}^n p^{(k)}(x),
    \]
    where the polynomials \(p^{(k)}\) consist of all monomials of exactly degree \(k\).
    For all \(k< m\) we then have for all \(i\)
    \begin{equation}
        \label{eq: lower order monomials disappear}    
        \lim_{\lambda\to \infty} \frac{p^{(k)}(\lambda v_i)}{\lambda^m} = 0.
    \end{equation}
    With the assumption that the slices of \(p\) are zero, i.e. \(p_{v_i}(\lambda) = 0\)
    we thus obtain
    \begin{align}
        \nonumber
        0 &= \lim_{\lambda\to \infty} \frac{p_{v_i}(\lambda)}{\lambda^m}
        \overset{\text{def.}}= \lim_{\lambda\to \infty} \frac{p(\lambda v_i)}{\lambda^m}
        = \lim_{\lambda\to \infty} \sum_{k=0}^n \frac{p^{(k)}(\lambda v_i)}{\lambda^m}
        \overset{\eqref{eq: lower order monomials disappear}}= \lim_{\lambda\to \infty} \frac{p^{(m)}(\lambda v_i)}{\lambda^m}
        \\
        \label{eq: monomials of degree n zero}
        &= p^{(m)}(v_i).
    \end{align}
    For the last equation we used that \(p^{(m)}\) consists only of monomials of
    exactly degree \(m\), for which we have
    \begin{equation}
        \label{eq: monomial scaling} 
        \mon_m(\lambda x) = \lambda^m \mon_m(x).
    \end{equation}
    Since \(p^{(m)}\) consists only of monomials of exactly degree \(m\),
    there exists a vector \(q\in \real^M\) with \(M\) the length of
    \(\mon_m(x)\) for \(x\in \real^\dims\) such that
    \begin{equation}
        \label{eq: p as linear combination of monomials}
        p(x) = q^T \mon_m(x).
    \end{equation}
    With \eqref{eq: monomials of degree n zero} we thus have
    \[
        q^T \underbrace{(\mon_m(v_1), \dots, \mon_m(v_N))}_{\in \real^{M\times N}} = 0.
    \]
    As the monomials \(\mon_m(v_1), \dots, \mon_m(v_N)\) span the space
    \(\real^M\), the matrix has rank \(M\) and thus \(q=0\). By \eqref{eq: p as
    linear combination of monomials} we thus have \(p^{(m)}\equiv 0\). Therefore
    \(p\) is of degree \(m-1\) and we finish the proof of the
    first claim of the theorem using the induction assumption.

    \item[\ref{item: optimality of number of directions}. ]
    Let \(N = \binom{n+\dims -1}{n}\) and let \(v_1,\dots, v_M\) be fewer
    directions \(M<N\). To prove the statement, we construct a non-zero
    \(\dims\)-variate polynomial of degree \(n\) (more specifically it only
    consists of monomials of exactly degree \(n\)), which is zero in all
    directions \(v_i\).
    To do so, consider the matrix
    \[
        A:= (\mon_n(v_1), \dots, \mon_n(v_M)) \in \real^{N\times M}.
    \]
    Since \(M<N\) it is at most of rank \(M\) there exists \(0\neq q\in \real^N\)
    such that \(q^T A = 0\). We then define the polynomial \(p(x) = q^T \mon_n(x)\).
    Then by construction this polynomial is zero at all \(v_i\). Moreover, by the
    scaling property of the monomials \eqref{eq: monomial scaling} we have
    \[
        p_{v_i}(\lambda) = q^T \mon_n(\lambda v_i) = \lambda^n \underbrace{q^T \mon_n(v_i)}_{= p(v_i)} = 0. 
        \quad \forall \lambda \in \real.
    \]
    Thus we have \(p_{v_i} \equiv 0\) for all \(i=1,\dots,M\) but \(p\neq 0\)
    since \(q \neq 0\).
    
    The intuition is in essence, that the scaling property of the monomials
    \eqref{eq: monomial scaling} implies that we only collect a single information
    point for each direction \(v_i\). To ensure the polynomial is zero, we thus
    have to collect enough points \(v_i\) to ensure the polynomial has to be zero.
    As the space of polynomials of exactly degree \(n\) has dimension \(N\), this
    is the number of points required.
\end{enumerate}

%% file: sections/neighborhood_redundant_params.tex

\section{The neighborhood of redundant parameters}
\label{sec: neighborhood of redundant parameters}

In this section, we analyze the redundant domain.  We will show that every redundant parameter lies on a line of (redundant) parameters for which the realization function is identical. In the setting with \emph{no regularization}, i.e., \(R\equiv 0\), this entails that redundant parameters always have a degenerate Hessian (in the sense that its determinant is zero) and it can never be a strict local minimum.\smallskip

In a second step,  we will show that for redundancies that are not deactivation redundancies typically either all or no points on the latter line are critical points of the optimization landscape. 

%

\begin{theorem}[Neighborhood of redundant critical points]
	\label{thm: critical points of redundant type} 	Let \(\network\) be an ANN and assume \(\param\) is redundant, i.e. \(\param
	\in \paramSpace\setminus \efficient(\Domain)\). Then there exists a straight line  \(\ell\subset \Theta\) containing \(\param\) such that for all \(\vartheta\in \ell\) 
	\[
		\response_{\vartheta} = \response_{\param}, \text{ \ on }\Domain.
	\]    
   \end{theorem}
   
   \begin{proof}
    If \(\param\) has a deactivation redundancy (Definition~\ref{def: efficient}
    \ref{it: disused neuron}) and \(w_{k\bullet}= 0\) for a \(k\in V_1\), then changing the parameters \(w_{i,j}\) and \(\beta_j\) (\(i\in V_0, j\in V_1\)) does have no impact on the response and clearly the respective set contains a line.
    
  Now suppose that there is \(\lambda\not\equiv 0\) such that
      \begin{equation}
        \label{eq: b redundancy assumption} 
  \lambda_\emptyset + \sum_{j\in V_1}\lambda_j
        \activation\Bigl(\bias_j + \sum_{i\in V_0}x_i \weight_{ij} \Bigr)=0
        \quad \forall x\in \Domain.
    \end{equation}
    We define \(\param(t) = (\weight(t),\bias(t))\) for \(t\in
    \real\) as follows: We retain the weights connecting the input to the first
    layer and its biases, i.e.
    \[
        \weight_{ij}(t) := \weight_{ij}
        \text{ \ \ and \ \ }
        \bias_j(t):=\bias_j
        \qquad
        \forall i\in V_0, j\in V_1,
    \]
    and in the second layer we add multiples of \(\lambda\) in an appropriate way:
    \begin{align*}
        \weight_{jk}(t) &:= \weight_{jk} + t\lambda_j\text{ \ \ and \ \ }\bias_k(t):= \bias_k + t\lambda_\emptyset
        &&\forall j\in V_1, k\in V_2.
    \end{align*}
	Since \(\lambda\not=0\), the function \((\theta(t))_{t\in \real}\)
	parametrizes a line \(\ell\) that contains \(\param\) and basic linear
	algebra implies with \eqref{eq: b redundancy assumption} that the response
	does not depend on the choice of \(t\): In terms of
    \[
	\psi _j(x):= \activation\Bigl(\bias_j + \sum_{i\in V_0}x_i \weight_{ij} \Bigr)\qquad \forall j\in V_1, x\in\Domain,
	\]
	one has for  every \(k\in V_2\) and \(x\in\Domain\) that
     \begin{align}
        \label{eq: outer parameters influence on response}
        (\response_{\param(t)}(x))_k
        &= \bias_k(t) + \sum_{j\in V_1} \weight_{jk}(t) \activation_j(x)
        \\
        \nonumber
        &= \bias_k + \sum_{j\in V_1} \weight_{jk} \activation_j(x)
        + t\underbrace{
            \Bigl(
                \lambda_0
                + \sum_{j\in V_1} \lambda_j\activation_j(x)
            \Bigr)
        }_{\overset{\eqref{eq: b redundancy assumption}}=0}
        \\
        \nonumber
        &=(\response_{\param}(x))_k.
    \end{align}
    \end{proof}

\begin{theorem}\label{thm:672345}
  Let \(\network\) be an ANN, \(X\) and \(Y\) be
\(\real^{V_\tin}\)- and \(\real^{V_\tout}\)-valued random variables,
\(\ell: \real^{V_\tout}\times \real^{V_\tout}\to [0,\infty)\) a \(C^1\)-function
such that the optimization landscape
\[
  \cost(\param)= \E[\ell(\Psi_\param(X), Y)]
\]
is \(C^1\) on \(\paramSpace\), and differentiation and integration can be interchanged.
Let the parameter \(\param=(w,\beta)\in\Theta\) exhibit a bias or duplication
redundancy (cf.\@ Remark \ref{rem: taxonomy of redundant parameters}) and let \(\param(t)\) be the parametrization of the line as
introduced after (\ref{eq: b redundancy assumption}).
Then either
\begin{itemize} \item for all \(t\in\real\), \(\param(t)\) is a critical parameter or
\item there it at most one  \(t\in\real\), for which \(\theta(t)\) is critical. 
\end{itemize}
If \(\# V_\tout =1\) and there are no deactivation redundancies, then
\(\param\) being critical implies that  \(\param(t)\) is critical for all  \(t\in
\real\).
\end{theorem}
\begin{proof} 
  In the  following, \(i\in V_0, j\in V_1,k,l\in V_2, t\in\real\) and \(x\in\Domain\) are arbitrary. Consider 
  \[
    \psi _j(x):= \activation\Bigl(\bias_j + \sum_{i\in V_0}x_i \weight_{ij} \Bigr)\text{ \ \ and \ \ }\psi_j'(x):=\psi'\Bigl(\bias_j + \sum_{i\in V_0}x_i \weight_{ij} \Bigr)
  \]
  Then we get for the inner differentials of the realization function
  \[
    \partial_{w_{i,j}} \bigl(\Psi_{\theta(t)}(x)\bigr)_l=\psi_j'(x)x_i w_{j,l}(t) \text{ \ \ and \ \ }  \partial_{\beta_{j}} \bigl(\Psi_{\theta(t)}(x)\bigr)_l=\psi_j'(x) w_{j,l}(t)
  \]
  and for the outer differentials
  \[
    \partial_{w_{j,k}} \bigl(\Psi_{\theta(t)}(x)\bigr)_l=\delta_{k,l} \psi_j(x)\text{ \ \ and \ \ }  \partial_{\beta_{k}} \bigl(\Psi_{\theta(t)}(x)\bigr)_l=\delta_{k,l},
  \]
  where \(\delta\) denotes the Kronecker-Delta. By assumption, we have that
  \begin{equation}
    \label{eq: cost representation}
    \nabla  \cost(\param(t))
    = \E\Bigl[
        \sum_{l\in V_\tout}
        \partial_{\hat{y}_l}\loss(\response_{\param(t)}(X), Y) \nabla ( \response_{\param(t)}(x))_l
    \Bigr].
  \end{equation}
  With the above identities we thus get for the inner and outer differentials
  \begin{align*}
    \partial_{w_{i,j}} \cost(\param(t))
    & = \sum_{l\in V_\tout} w_{j,l}(t)  \, \E\Bigl[
      \partial_{\hat{y}_l}\loss(\response_{\param}(X), Y)\, \psi_j'(X)\,X_i \Bigr]
    \\ 
    \partial_{\beta_{j}} \cost(\param(t))
    & = \sum_{l\in V_\tout} w_{j,l}(t)  \, \E\Bigl[
      \partial_{\hat{y}_l}\loss(\response_{\param}(X), Y)\, \psi_j'(X) \Bigr],
    \\
    \label{eq: outer weight derivative}
    \partial_{w_{j,k}} \cost(\param(t))
    & = \E\Bigl[
      \partial_{\hat{y}_k}\loss(\response_{\param}(X), Y)\, \psi_j(X) \Bigr]
    = \partial_{w_{j,k}} \cost(\param),
    \\
    \partial_{\beta_{k}} \cost(\param(t))
    &= \E\Bigl[
      \partial_{\hat{y}_k}\loss(\response_{\param}(X), Y) \Bigr]
    = \partial_{\beta_{k}} \cost(\param).
  \end{align*}
  By the latter two identities,
  the derivatives with respect to the outer parameters do not depend on \(t\).
  The inner derivatives can be expressed in terms of
  \[
    a_{i,j,l}
    = \E\Bigl[          
      \partial_{\hat{y}_l}\loss(\response_{\param}(X), Y)\, \psi_j'(X)\,X_i 
    \Bigr]
    \text{ \ and \ }
    b_{j,l}
    =  \E\Bigl[          
      \partial_{\hat{y}_l}\loss(\response_{\param}(X), Y)\, \psi_j'(X)\,X_i
    \Bigr],
  \]
  specifically
  \begin{align*}
    \partial_{w_{i,j}} \cost(\param(t))
    &= \sum_{l\in V_\tout} w_{j,l}(t) \,a_{i,j,l}
    = \sum_{l\in V_\tout} (w_{j,l}+t\lambda _j) a_{i,j,l},
    \\
    &=\sum_{l\in V_\tout} w_{j,l}  a_{i,j,l} + t \lambda_j \sum_{l\in V_\tout}  a_{i,j,l} \\
    \partial_{\beta_{j}} \cost(\param(t))
    &= \sum_{l\in V_\tout} w_{j,l}(t) \, b_{j,l}
    =\sum_{l\in V_\tout} (w_{j,l}+t\lambda _j) b_{j,l}
    \\
    &=\sum_{l\in V_\tout} w_{j,l}  b_{j,l} + t \lambda_j \sum_{l\in V_\tout}  b_{j,l}.
  \end{align*}       
  Note that all differentials \(\partial_{w_{i,j}} \cost(\param(t))\) and
  \(\partial_{\beta_{j}} \cost(\param(t))\) are affine functions in \(t\).
  Hence, each differential is either zero for all \(t\in\real\) or at most one
  point \(t\). Consequently, on the line \((\param(t))_{t\in\real}\) either all
  points are critical or there is at most one point that is critical.
  In the case of \(\# V_\tout =1\), and no deactivation redundancies, i.e.
  \(\weight_{j\placeholder} \neq 0\), a critical point in \(t=0\) implies \(a_{i,j,l} = b_{j,l}=0\)
  for all \(i,j\) and thereby all \(\param(t)\) are critical.
\end{proof}

%% file: sections/existence_critical_points.tex

\section{Existence of efficient critical points}
\label{sec: existence of efficient critical points}

In this section, we analyze the existence of local minima in the standard setting
(Def.~\ref{def: standard model}). More explicitly, we prove that for every open
set \(U\subseteq \paramSpace\) containing a polynomially efficient parameter \(\param\)
one has with strictly positive probability that the random
unregularized squared error loss contains a local minimum in~\(U\).   The
result illustrates that local minima may exist in the unregularized
setting. In the case of non-trivial regularization the cost typically tends to
infinity when the parameter \(\param\) tends to infinity. In this case the
existence of (local) minima is  trivial.

\begin{restatable}[Efficient minima exist with positive probability]{theorem}{efficientExist}
	\label{thm: existence of efficient minima}
	Assume that we are in the unregularized (i.e., \(R\equiv0\)) standard setting (Definition~\ref{def: standard model}) and that the random target function \(\rf=(f_{\mathbf M}(\param))_{\param\in\Theta}\) additionally satisfies that for all continuous functions \(\phi:\real^{V_\tin}\to \real\) and \(\delta \in(0,\infty)\) one has
	\[
	\Pr(\|\rf-\phi\|_{\Pr_X}<\delta)>0.
	\]

	Then every non-empty, open set \(U\) of \((0,0,1)\)-polynomially efficient parameters
	contains a local minimum of
	the MSE cost with positive probability, i.e.
	\[
		\Pr\Bigl(
			\exists \param \in U:
			\param \text{ is a local minimum of } \Cost
		\Bigr) > 0.
	\]
\end{restatable}


\begin{proof}[Proof of Theorem~\ref{thm: existence of efficient minima}]

%

	Recall that for every \(\problem \in \problemSpace\) the MSE cost function is of the form
	\begin{align*}
		\cost_\problem(\param)
		&= \E_\problem[\|\response_\param(X) - Y\|^2]
		\\
		\overset{(*)}&= \E_\problem[\|\response_\param(X) - \target_\problem(X)\|^2]
		+ \underbrace{\E_\problem[\|\target_\problem(X)-Y\|^2 ]}_{\text{`noise' const. in }\param}.
	\end{align*}
	For \((*)\) we note that \(\E_\problem[Y-\target_\problem(X)\mid X] =0\) by
	definition of the target function \(\target_\problem(x) = \E_\problem[Y\mid X=x]\),
	and the mixed term therefore disappears.
	Since we assumed \(\# V_\tout=1\), the norm is simply a square. Consequently, we get by interchanging differentiation and integration (this can be justified in complete analogy to Lemma~\ref{lem: differentiability}) that
	\begin{align}
		\label{eq: cost gradient}
		\nabla \cost_\problem(\param)
		&= 2\int(\response_\param(x) - f_\mathbf{m}(x))\nabla_\param\response_\param(x) \,\Pr_X(dx) \quad \text{and}
		\\
		\label{eq: cost hessian}
		\nabla^2 \cost_\problem(\param)
		&= 
			2\E\bigl[\nabla_\param\response_\param(X)\nabla_\param\response_\param(X)^\transpose\bigr]
			+ 2 \int (\response_\param(x) - f_\mathbf{m}(x))\nabla^2\response_\param(x) \Pr_X(dx).
	\end{align}
	Note that if the realization \(f_\mathbf{m}\) is very close to the response \(\Psi_{\param}\) for an  efficient parameter \(\param\in U\), then we informally have that
	\[
		\nabla \cost_\problem(\param)\approx 0
		\quad\text{and}\quad
		\nabla^2\cost_\problem(\param)\approx 2\E\bigl[
			\nabla_\param\response_\param(X)\nabla_\param\response_{\param}(X)^\transpose
		\bigr]
		=:2 G_{\param}.
	\]
	Our proof strategy is therefore to 
	\begin{itemize}
		\item show that \(G_\param\) is strictly positive definite,

		\item carry out a spectral analysis for  \(\nabla^2\cost_\problem(\param)\)

		\item show that in the case that the target function \( f_\mathbf{m}\) is
		close to \(\Psi_\param\), there exists a local minimum in the neighborhood of
		an efficient parameter \(\param_0\).
	\end{itemize}
%
%
	\paragraph*{Strict positive definiteness of \(G_\param\).}
	 Let \(v\in \real^{\dim(\param)}\). We need to show that 
	 \[v^\transpose G_\param v\ge0 \text{ \ \  and \ \ }   [v^\transpose G_\param v=0 \ \Rightarrow \ v=0].\] 	 
	One has 
	\[
		 v^\transpose G_\param v
		= v^\transpose \E\bigl[
			\nabla_\param\response_\param(X)\nabla_\param\response_\param(X)^\transpose
		\bigr] v
		= \Bigl\| \langle v, \nabla_\param\response_\param(\cdot)\rangle \Bigr\|_{\Pr_X}^2\ge 0.
	\]
	Now suppose that \(v^\transpose G_\param v=0\). Then 
	\[
		\langle v, \nabla_\param\response_\param(\cdot)\rangle = 0, \text{ \ \ \(\Pr_X\)-almost surely.}
	\]
	 The latter function is analytic (since
	\(\response_\param\) is analytic and \(\Pr_X\) is compactly supported). Consequently,  it is zero on the entire support \(\Domain\)
	of \(\Pr_X\).

	Recall that \(\# V_\tout=1\) and let \(V_\tout = \{\outSgt\}\).  The
	derivatives of \(\response_\param\) are then given by
	\begin{align}
		&x\mapsto 1 
		\tag{\(\partial \bias_\outSgt\)}
		\\
		&x \mapsto \activation\bigl(\bias_j + \langle x, \weight_{\placeholder j}\rangle \bigr)
		&& j\in V_1
		\tag{\(\partial \weight_{j\outSgt}\)}
		\\
		&x \mapsto
		\activation'\bigl(\bias_j + \langle x, \weight_{\placeholder j} \rangle\bigr) \weight_{j\outSgt}
		&& j\in V_1
		\tag{\(\partial \bias_j\)}
		\\
		&x \mapsto \activation'\bigl(\bias_j + \langle x, \weight_{\placeholder j} \rangle\bigr) \weight_{j\outSgt}x_i
		&&  j\in V_1,\; i\in V_0 
		\tag{\(\partial \weight_{ij}\)}
	\end{align}
	We  thus get the representation 
	\[
		 \langle v, \nabla_\param \response_\param(x)\rangle
		= v_{\bias_\outSgt} + \sum_{j\in V_1} \Phi_j(x)
		\qquad \forall x \in \Domain
	\]
	with
	\[
		\Phi_j(x)
		:= \underbrace{v_{\weight_{j\outSgt}}}_{=:P_0^{(j)}}
		\activation\bigl(\bias_j + \langle x, \weight_{\placeholder j}\rangle\bigr)
		+ \underbrace{\Bigl(v_{\bias_j} w_{j*} + \sum_{i\in V_0} v_{\weight_{ij}}  w_{j*}x_i\Bigr)}_{=:P_1^{(j)}(x)}
		\activation'\bigl(\bias_j + \langle x, \weight_{\placeholder j}\rangle\bigr),
	\]
	where we write \(v_{\param_i}:= v_i\) (to easily refer to the particular
	parameters). Recall that since \(\param\) is \((0,0,1)\)-polynomially
	efficient the function \(\langle v, \nabla_\param
	\response_\param(\cdot)\rangle\) can only be zero on \(\Domain\), if all
	polynomials (`coefficients') are zero. By assumption, every \(w_{j*}\not=0\)
	so that we get that indeed all entries of the vector \(v\) are zero. Thereby we
	proved that \(G_\param\) is strictly positive definite and that the minimal
	eigenvalue \(\underline{\lambda} _\param\) of \(\param\) is strictly
	positive.	
	
	\paragraph*{Spectral analysis  of \(\nabla^2 \cost_\problem(\param)\).}
	Observe that in terms of 
	\[
		\overline \lambda_\param
		:= \sup_{\|v\|=1}\bigl\| v^T \nabla_\param^2\response_\param(\cdot)v \bigr\|_{\Pr_X}
		\le \Bigl(\int \| \nabla^2\Psi_\param (x)\|_{\text{op}}\,\Pr_X(dx)\Bigr)^{1/2}
	\] 
	one has for  \(v\in\paramSpace\) that by the Cauchy-Schwarz inequality
	\begin{align}
		\nonumber
		v^\transpose \nabla^2\cost_\problem(\param) \,v
		\overset{\eqref{eq: cost hessian}}&= 2\Bigl(
			v^\transpose G_\param v
			+ \int (\response_\param(x) - f_\mathbf{m}(x)) v^\transpose \nabla_\param^2\response_\param(x) v \;\Pr_X(dx) 
		\Bigr)
		\\
		\label{eq: lower bound hessian cost}
		&\ge 2\Bigl(
			\underline{\lambda} _\param
			- \|\response_\param - f_\mathbf{m}\|_{\Pr_X} \overline{\lambda}_\param
			\Bigr)\|v\|^2.
	\end{align}

	\paragraph*{Finding a local minimum.}
	We will prove that there exists a local minimum in \(B_\delta(\param_0)\)
	for \(\delta>0\), if there exists a lower bound on the spectrum of the
	Hessian \(\rho>0\) such that
	\begin{equation}
		\label{eq: sufficient for local minimum}
		\|\nabla \cost_\problem(\param_0)\| 
		< \tfrac{\delta}2 \rho
		\qquad \text{and}\qquad
		\bigl[\nabla^2\cost_\problem(\param) \succeq \rho
		\quad \forall \param \in B_\delta(\param_0)\bigr].
	\end{equation}
	Since \(\cost_\problem\) is thereby \(\rho\)-strongly convex on
	\(B_\delta(\param_0)\) \citep[Thm.\@
	2.1.11]{nesterovLecturesConvexOptimization2018} we have for all \(\param \in
	B_\delta(\param_0)\) \citep[Def.\@ 2.1.3]{nesterovLecturesConvexOptimization2018}
	\begin{align*}
		\cost_\problem(\param)
		&\ge \cost_\problem(\param_0) + \langle \nabla\cost_\problem(\param_0), \param- \param_0\rangle
		+ \tfrac\rho2 \|\param - \param_0\|^2
		\\
		&\ge \cost_\problem(\param_0)
		- \|\nabla \cost_\problem(\param_0)\| \|\param - \param_0\| + \tfrac\rho2\|\param-\param_0\|^2.
	\end{align*}
	For all parameters \(\param\) on the boundary \(\partial B_\delta(\param_0)\) of the ball
	we therefore have
	\[
		\cost_\problem(\param)
		\ge \cost_\problem(\param_0)
		- \|\nabla \cost_\problem(\param_0)\| \delta + \tfrac\rho2 \delta^2
		\overset{\eqref{eq: sufficient for local minimum}}> \cost_\problem(\param_0).
	\]
	The minimum, which the cost \(\cost_\param\) assumes on the (compact) closed ball
	\(\overline{B_\delta(\param_0)}\), can therefore not be on the boundary. Consequently,
	there must be a local minimum in \(B_\delta(\param_0)\).
	
	\paragraph*{Finishing the proof}
	By reducing the size of the open set \(U\) if necessary, we can
	assume without loss of generality that its closure \(\closure{U}\) is
	compact and also contained in the set of polynomially efficient parameters
	\(\polyEfficient^{(0,0,1)}\).
	Now suppose that \(\theta_0\) is an arbitrary efficient element of \(U\).
	The continuous maps \((\underline {\lambda}_\param)_{\param\in\paramSpace}\) and
	\((\overline {\lambda}_\param)_{\param\in\paramSpace}\) both attain their
	minimum and maximum on the compact set \(\overline U\), where
	all \(\underline\lambda_\param >0\) and all \(\overline\lambda_\param<\infty\) so that
	\[
		\underline {\lambda}
		= \min _{\param\in \overline U}\underline {\lambda}_\param>0 \text{ \ \ and \ \ }\overline {\lambda}=\max _{\param\in \overline U}\overline {\lambda}_\param<\infty.
	\]
	To satisfy \eqref{eq: sufficient for local minimum} with \(\rho:=\underline{\lambda}\),
	let \(\epsilon:= \underline{\lambda}/(2\overline{\lambda})\) and select
	\(\delta\)
	sufficiently small such that
	\begin{itemize}
		\item \(B_\delta(\param_0) \subseteq U\subseteq \polyEfficient^{(0,0,1)}\)
		(this ensures a minimum in \(U\))
		\item  \(\|\response_\param - \response_{\param_0}\|_{\Pr_X}\le
		\epsilon/2\) for all \(\param \in B_\delta(\param_0)\) (using continuity of \(\response_\param\)).
	\end{itemize}
	Then choose \(r\in (0, \epsilon/2)\) such that 
	\(\|\nabla_\param\response_{\param_0}\|_{\Pr_X} < \tfrac{\delta}{2r} \rho\).
	Consequently the inequality \(\|\response_{\param_0}-\target_\problem\|\le r\) implies
	\begin{enumerate}
		\item by \eqref{eq: cost gradient} and Cauchy's inequality
		\[
			\|\nabla\cost_\problem(\param_0)\|
			\le \|\response_{\param_0} - \target_\problem\|_{\Pr_X} \|\nabla\response_{\param_0}\|_{\Pr_X}
			< \tfrac{\delta}{2} \rho,
		\]
		\item
		and for all \(\param\in B_\delta(\param_0)\)
		\[
			\|\response_\param - \target_\problem\|_{\Pr_X}
			\le \|\response_\param - \response_{\param_0}\|_{\Pr_X} + \|\response_{\param_0} - \target_\problem\|_{\Pr_X}
			\le \epsilon .
		\]
		Using \(\epsilon= \underline{\lambda}/(2\overline{\lambda})\) and
		\eqref{eq: lower bound hessian cost} we can lower bound spectrum by \(\rho\),
		i.e.\ for all \(v\) such that \(\|v\|=1\)
		\[
			v^\transpose \nabla^2\cost_\problem(\param) \,v
			\ge 2(\underline{\lambda} - \epsilon
			\overline\lambda)=\underline\lambda\overset{\text{def.}}=\rho.
		\]
	\end{enumerate}
	This means we satisfy \eqref{eq: sufficient for local minimum}
	if \(\|\response_{\param_0}-\target_\problem\|\le r\).
	By assumption on the random statistical model
	\(\mathbf M\) the latter property holds with strict positive probability.
\end{proof}

%% file: sections/existence_redundancies.tex
\section{Existence of redundant critical points}
\label{sec: existence of redundant critical points}

Since the set of redundant parameters is generally a thin set with respect to
the Lebesgue measure (e.g.\ \(\efficient_0\) in Theorem~\ref{thm: characterization of efficient networks}), one may
reasonably hope that this set does not contain any critical points of
the MSE with probability one. In that case the MSE would be a Morse function
over the entire set of parameters with probability one. Unfortunately, this
hypothesis is wrong in general as the following theorem shows. We further break
down the set of redundant parameters, using the taxonomy introduced in
Remark~\ref{rem: taxonomy of redundant parameters}, to make more precise
statements about the existence of redundancies which are required to be of a
certain type.

\begin{theorem}[Redundancies cannot be ruled out in general]
    \label{thm: redundanct critical points exist}
    Assume the standard setting (Definition~\ref{def: standard model})
    without regularization, i.e. \(\regularizer\equiv 0\).

    Assume \(\polyEfficient^{(0,0,1)}\) contains an open set\footnote{
        \label{footnote: e.g. sigmoid tanh}
        e.g. \(\activation\in \{\sigmoid, \tanh\}\) and an open set in the support of \(\Pr_X\) by Theorem~\ref{thm: characterization of efficient networks}
    } and that there is at least one hidden neuron (\(\#V_1\ge 1\)),
	then, with \textbf{positive probability}, critical points of the MSE
	\(\Cost\) do \textbf{exist} in the sets of
	\begin{enumerate}[label={\normalfont(\Alph*)},series=redundantSets]
		\item\label{it: rule out not possible - all} redundant parameters,
		\item\label{it: rule out not possible - dup}
		redundant parameters that only admit duplication redundancies (assuming \(\# V_1\ge 2\))
		\item\label{it: rule out not possible - bias and deac} 
        redundant parameters that only admit bias \emph{and} deactivation redundancies,
	\end{enumerate}
\end{theorem}

\begin{proof}[Proof (outline)]
    Clearly, the existence of redundant critical points \ref{it: rule out not possible - all}
    follows from the existence of critical points with more specific redundancies, i.e. \ref{it: rule out
    not possible - dup} or \ref{it: rule out not possible - bias and deac}. So we
    only need to prove \ref{it: rule out not possible - dup} and \ref{it: rule out not possible - bias and deac}.
    To do so, we make use of the fact that we have proven efficient
    critical points exist with positive probability (Theorem~\ref{thm: existence
    of efficient minima}). Using \(\#V_1\ge 1\) we can therefore find a critical
    point of a smaller network with \(\#V_1-1\)
    hidden neurons with positive probability. We then carefully \textbf{extend} this
    network and its parameters by a redundancy in a fashion that retains the
    criticality of the parameters. But for a duplication redundancy we obviously
    need at least two hidden neurons. Details follow in Section~\ref{sec: extending}.
\end{proof}

Conversely, pure bias redundancies can be ruled out.

\begin{prop}[Pure bias redundancies can be ruled out]
    \label{prop: pure bias redundancies can be ruled out}
    Assume the standard unregularized setting (Definition \ref{def: standard model}).
    If \(\activation'(x)\neq 0\) for all \(x\in \real\), the support \(\Domain\)
    of \(\Pr_X\) contains an open set and an efficient parameter is automatically \((1,0,1)\)-polynomially efficient,\footref{footnote: e.g. sigmoid tanh}
    then, with \textbf{probability one}, critical points of the MSE \(\Cost\)
    do \textbf{not exist} in the set of
    \begin{enumerate}[resume*=redundantSets]
        \item\label{it: no pure bias redundancies} redundant parameters that
        only admit bias redundancies.
    \end{enumerate}
    
\end{prop}
\begin{proof}[Proof (outline)]
    The proof of \ref{it: no pure bias redundancies} relies on \textbf{pruning} the bias
    redundancies to obtain an efficient parameter for a smaller network. This
    efficient parameter must then also be a critical point of the cost and satisfy
    an additional condition. We then show that there are almost surely no efficient
    critical points which satisfy this additional condition and thereby
    rule out critical bias redundancies. Details follow in Section \ref{sec: critical bias red do not exist} 
    after we outline a general pruning process in Section \ref{sec: pruning}.
\end{proof}

\subsection{Extending (Proof of Theorem \ref{thm: redundanct critical points exist})}
\label{sec: extending}

Using the following lemma to extend an efficient critical point
of a smaller network, \ref{it: rule out not possible - dup} and \ref{it: rule
out not possible - bias and deac} clearly follow from the existence of such
an efficient critical point in the smaller network positive probability.
This follows from Theorem~\ref{thm: existence of efficient
minima}, for which we require the existence of an open set in the set
\(\polyEfficient^{(0,0,1)}\).

\begin{lemma}[Extension]
    Assume the setting of Theorem \ref{thm: redundanct critical points exist}
    and without loss of generality \(V_1 = \{1,\dots, \# V_1\}\). Define
    the reduced ANN to be \(\tilde \network = (\tilde{\mathbb V}, \activation)\)
    with neurons \(\tilde{\mathbb V}:= \bigl(V_0, V_1\setminus\{1\},
    V_2\bigr)\).
    Assume that the parameter \(\tilde \param\) of the network \(\tilde \network\)
    is a critical point of \(\cost_\problem\).
    Then there exists a parameter \(\param\) of the network
    \(\network\) such that it is a critical point of \(\cost_\problem\) and
    either
    \begin{enumerate}
        \item\label{it: single dup} \(\param\) only has a single duplication redundancy and no
        other redundancies (if we further assume \(\#V_1\ge 2\)), or
        \item\label{it: only deac and bias} \(\param\) only has a deactivation and bias redundancy at the
        same neuron and no other redundancies.
    \end{enumerate}
\end{lemma}
\begin{remark}
    We will not make use of the fact that the loss \(\loss\) is the squared
    error. We only require sufficient regularity such that derivatives may be
    moved into the expectation (e.g. Lemma~\ref{lem: differentiability}).
\end{remark}

\paragraph*{Proof of \ref{it: single dup}.}

We are going to construct a parameter \(\param\) with
a single duplication redundancy from the parameter \(\tilde \param\) of the
reduced network. Assume that the parameters we do not mention are kept as is.
Our plan is to duplicate the neuron \(2\) so we define for all \(i\in V_0\)
\begin{equation}
    \tag{duplication}
    \weight_{i1} := \tilde\weight_{i2}
    \qquad\text{and}\qquad 
    \bias_1 := \tilde \bias_2.
\end{equation}
To ensure neither neuron is deactivated pick \(\lambda \in
\real\setminus\{0,1\}\) and
define
\begin{equation}
    \tag{`convex' combination}    
    \weight_{1l} := \lambda \tilde \weight_{2l}
    \qquad\text{and}\qquad
    \weight_{2l} := (1-\lambda)\tilde \weight_{2l}.
\end{equation}
Clearly, \(\param\) is in the set of parameters which only admit duplication
redundancies.

It is straightforward to see, that the response must remain the same, i.e.
\(\response_\param = \response_{\tilde{\param}}\), as we have just split one
identical neuron into a `convex' combination of two identical ones. That is
\begin{align*}
    (\response_{\param}(x))_l
    &= \bias_l + \sum_{j=1}^{\# V_1} \activation\bigl(\bias_j + \langle x, \weight_{\placeholder j}\rangle\bigr)\weight_{jl}
    \\
    &= 
    \bias_l
    + \activation\bigl(\bias_{2} + \langle x, \weight_{\placeholder 2}\rangle\bigr)
    \underbrace{\bigl((1-\lambda)\weight_{1 l} + \lambda \weight_{2 l}\bigr)}_{
        = \tilde{\weight}_{2 l} 
    }
    + \smashoperator{\sum_{j=3}}^{\# V_1}\activation\bigl(\bias_j + \langle x, \weight_{\placeholder j}\rangle\bigr) \weight_{jl}
    \\
    &= \tilde{\bias}_l + \sum_{j\in V_1\setminus\{1\}} \activation\bigl(\tilde{\bias}_j + \langle x, \tilde{\weight}_{\placeholder j}\rangle\bigr)\tilde{\weight}_{jl}
    \\
    &=(\response_{\tilde{\param}}(x))_l.
\end{align*}
With this fact under our belt, we can now consider the derivatives
of the cost. Recall that
we denote by \(\partial_{\hat y_l}\loss\) the partial derivative of the loss
\(\loss(\hat{y}, y)\) with respect to the \(l\)-th component of the prediction
\(\hat y\). For \(l\in V_2\), \(j\in V_1\) and \(i\in V_1\) we then have
\begin{align*}
    \partial_{\bias_l} \cost_\problem(\param)
    &=
    \E_\problem\bigl[
        \partial_{\hat{y}_l}\loss(\response_{\param}(X), Y) \underbrace{\partial_{\bias_l}(\response_{\param}(X))_l}_{=1}
    \bigr]
    \overset{\response_\param=\response_{\tilde\param}}=
    \partial_{\tilde\bias_l} \cost_\problem(\tilde\param)
    = 0,
    \\
    \partial_{\weight_{jl}} \cost_\problem(\param)
    &=
    \E_\problem\bigl[\partial_{\hat{y}_l}\loss(\response_{\param}(X), Y)
        \underbrace{
            \partial_{\weight_{jl}}(\response_{\param}(X))_l
        }_{
            =\activation(\tilde{\bias}_j + \langle \tilde{\weight}_{\placeholder j}, X\rangle)
        }
    \bigr]
    \overset{\response_\param=\response_{\tilde\param}}=
    \partial_{\tilde\weight_{jl}} \Cost(\tilde\param) = 0,
    \\
    \partial_{\bias_j} \cost_\problem(\param)
    &=
    \E_\problem\Bigl[
        \sum_{l\in V_2}\partial_{\hat{y}_l}\loss(\response_{\param}(X), Y)
        \underbrace{\weight_{jl}}_{=\mathrlap{\begin{cases}
            (1-\lambda \delta_{j2})\tilde\weight_{jl}
            & j\neq 1 
            \\
            \lambda \tilde\weight_{2l}
            & j = 1 
        \end{cases}
        }\qquad}
        \activation'(\bias_j
        + \langle X, \weight_{\placeholder j}\rangle)
    \Bigr]
    \\
    \overset{\response_\param=\response_{\tilde\param}}&=
    \begin{cases}
        (1-\lambda \delta_{j2})\partial_{\tilde\bias_j} \cost_\problem(\tilde\param) & j\neq 1
        \\
        \lambda \partial_{\tilde\bias_2} \cost_\problem(\tilde\param)  & j= 1
    \end{cases}
    \\
    &= 0,
    \\
    \partial_{\weight_{ij}} \cost_\problem(\param)
    &= \E\Bigl[
        \sum_{l\in V_2}\partial_{\hat{y}_l}\loss(\response_{\param}(X), Y)
        \activation'(\bias_j
        + \langle X, \weight_{\placeholder j}\rangle)
        \weight_{jl}X_i
    \Bigr]
    \\
    \overset{\response_\param=\response_{\tilde\param}}&=
    \begin{cases}
        (1-\lambda \delta_{j2})\partial_{\tilde\weight_{ij}} \cost_\problem(\param) & j\neq 1
        \\
        \lambda \partial_{\tilde\weight_{i2}} \cost_\problem(\param)  & j= 1
    \end{cases}
    \\
    &= 0.
\end{align*}
the parameter \(\param\) is thereby clearly a critical point with no other redundancies
except for a single duplication.

\paragraph*{Proof of \ref{it: only deac and bias}.}
To construct a parameter \(\param\) with a deactivation and bias redundancy
from the reduced network, define
\[
    \weight_{\placeholder 1} = 0,
    \qquad
    \weight_{1 \placeholder} = 0,
\]
select \(\bias_1 \in \real\) arbitrarily and retain all parameters of \(\tilde
\param\) for the other neurons.

Since the additional neuron is deactivated, the response remains the same, i.e.
\(\response_\param = \response_{\tilde\param}\).
And since \(\tilde\param\) is a critical point, it is straightforward
to check that the derivatives with respect to the old parameters remain
the same and are thereby zero. For the derivatives with respect to the new
parameters let us consider the outer derivatives first
\begin{align*}
    \partial_{\weight_{1 l}}\cost_\problem(\param)
    &=
    \E_\problem\bigl[
        \partial_{\hat{y}_l}\loss(\response_\param(X), Y)
        \activation\bigl(
            \bias_1 + \langle X, \weight_{\placeholder 1}\rangle
        \bigr)
    \bigr]
    \\
    \overset{\weight_{\placeholder 1}=0}&=
    \underbrace{\E_\problem\bigl[
        \partial_{\hat{y}_l}\loss(\response_\param(X), Y)
    \bigr]}_{=\partial_{\tilde\bias_l}\cost_\problem(\tilde\param) = 0}
    \activation(\bias_{1}).
\end{align*}
In the last equation we used that the response remains the same.
The derivatives with respect to the inner derivatives on the other hand are all
zero due to the deactivation with the outer parameter
\(\weight_{1\placeholder}=0\).

\subsection{Pruning}
\label{sec: pruning}

The following result shows that for any redundant parameter there exists an
efficient parameter of a smaller network with equal response function on
the support \(\Domain\) of the input \(X\). We also show that criticality is
retained under the standard assumption of \(\# V_\tout=1\) (cf.
Definition~\ref{def: standard model}).

\begin{prop}
    \label{prop: pruning}
    Let \(\network = (\mathbb V, \activation)\) with \(\mathbb V=(V_0, V_1,
    V_2)\) be a shallow ANN as in Definition~\ref{def: shallow neural network}.
    Assume that the parameter \(\param\) of this network is redundant. Then
    there exists a pruned network \(\tilde{\network} = ((V_0, \tilde{V}_1, V_2),
    \activation)\) with \(\tilde V_1 \subseteq V_1\) and an \textbf{efficient}
    parameter \(\tilde \param\) of this pruned network such that the response
    remains the same, i.e.
    \[
        \response_\param(x) = \response_{\tilde \param}(x) \qquad \forall x\in \Domain.
    \]
    Furthermore, if \(\# V_\tout=1\) and \(\tilde \param\) was a critical point
    of some cost function
    \[
        \cost_\problem(\param) = \E_\problem[\loss(\response_{\param}(X), Y)]
    \]
    for some loss function \(\loss\) and expectation \(\E_\problem\) induced by
    some distribution \(\Pr_\problem\), then (assuming
    suitable regularity on \(\loss\) and \(\activation\) such that derivatives
    may be moved into the expectation \(\E_\problem\), cf. Lemma~\ref{lem:
    differentiability}) the pruned parameter \(\tilde \param\) is also a
    critical point of \(\cost_\problem\).
\end{prop}

\begin{proof}
    A parameter is redundant if either of the two criterions \ref{it: disused neuron} 
    or \ref{it: linear combination of neurons} in the Definition of Efficiency
    \ref{def: efficient} are violated. Recall, that we called the violation
    of criterion \ref{it: disused neuron} a deactivation redundancy
    (Remark~\ref{rem: taxonomy of redundant parameters}).
    
    \paragraph*{Deactivation pruning}
    It is straightforward to see that the response of an ANN does not change
    if all the deactivated hidden neurons, i.e. all \(j\in V_1\) where
    \(\weight_{j\placeholder} =0\), are removed from \(V_1\). Similarly,
    it is straightforward to show that critical parameters remain critical
    since the previous gradient contains all the partial derivatives with
    respect to the remaining parameters in the pruned network.

    \paragraph*{Pruning the other redundancies}
    Until the parameter is efficient we will iteratively remove a single neuron,
    while ensuring that the response stays the same and critical points remain
    critical. Since there only a finite number of neurons, this procedure will
    eventually terminate -- if there are no hidden neurons left, then there
    is only a bias on the output which is clearly efficient (Definition
    \ref{def: efficient}). We therefore only describe the procedure of a single
    step.

    Note that if a pruning step reintroduces deactivation redundancies, we
    interject a deactivation pruning step. We can therefore always assume there
    are no deactivation redundancies at the beginning of a pruning step.

    If the parameter \(\param\) is redundant without deactivation redundancies
    \ref{it: disused neuron}, then there must be a hidden neuron \(k\in V_1\) that
    can be linearly combined from the others (cf. Remark~\ref{rem: taxonomy of
    redundant parameters}), i.e.
    \[
        \activation\Bigl(\bias_k + \sum_{i\in V_0}  x_i\weight_{ik} \Bigr)
        = \lambda_\emptyset + \sum_{j\in V_1\setminus \{k\}}\lambda_j
        \activation\Bigl(\bias_j + \sum_{i\in V_0} x_i\weight_{ij}\Bigr)
        \quad \forall x\in \Domain.
    \]
    We define a parameter \(\tilde \param\) for the pruned network
    \(\tilde \network = ((V_0, V_1\setminus\{k\}, V_2), \activation)\) using the
    parameter \(\param\) from the old network. Specifically, we retain all inner
    parameters and define the outer parameters to be
    \[
        \tilde \weight_{jl} := \weight_{jl} +  \weight_{kl}\lambda_j
        \qquad \tilde \bias_l := \bias_l + \weight_{kl}\lambda_\emptyset 
        \qquad \forall l\in V_2,\; j\in V_1\setminus\{k\}.
    \]
    Using this definition it is straightforward to check that the response
    remains the same, i.e. \(\response_\param = \response_{\tilde \param}\).
    
    In the case \(\# V_\tout=1\), i.e. \(V_\tout = \{\outSgt\}\),
    we need to show that this pruning step retains criticality. Recall that
    the derivatives are given by
    \begin{equation}
        \label{eq: previously critical}    
        \partial_{\param_i}\cost_\problem(\param)
        = \E_\problem\bigl[\partial_{\hat y}(\response_\param(X), Y) \partial_{\param_i}\response_\param(X)\bigr].
    \end{equation}
    Since the derivatives of the response
    \(\partial_{\tilde \bias_\outSgt}\response_{\tilde\param}\) and
    \(\partial_{\tilde \weight_{j\outSgt}}\response_{\tilde\param}\) with respect to the outer
    parameters only contain inner parameters (which we have not changed) and the
    response remains the same \(\response_{\tilde \param} = \response_\param\),
    we immediately get the criticality of the partial derivatives with
    respect to the outer parameters. What is left to consider are the derivatives
    with respect to the inner derivatives. Since \(\param\) had no deactivation
    redundancies, we have \(\weight_{j \outSgt} \neq 0\) for all \(j\in V_1\).
    In particular we have for all \(j\in V_1\setminus \{k\}\)
    \begin{align*}
        \partial_{\tilde \bias_j}\response_{\tilde \param}(x)
        &= \activation'(\bias_j + \langle x, \weight_{\placeholder j}\rangle)
        \tilde\weight_{j \outSgt}
        = \partial_{\bias_j}\response_{\param}(x) \frac{\tilde \weight_{j \outSgt}}{\weight_{j\outSgt}}
        \\
        \partial_{\tilde \weight_{ij}}\response_{\tilde \param}(x)
        &= \activation'(\bias_j + \langle x, \weight_{\placeholder j}\rangle)
        x_i\tilde\weight_{j \outSgt}
        = \partial_{\weight_{ij}}\response_{\param}(x) \frac{\tilde \weight_{j \outSgt}}{\weight_{j\outSgt}}
    \end{align*}
    The derivatives of the response therefore only change up to a constant that
    can be moved out of the expectation in \eqref{eq: previously critical}.
    Together with \(\response_{\tilde \param} = \response_\param\) this yields criticality.
\end{proof}

\subsection{Bias redundancies (Proof of Proposition \ref{prop: pure bias redundancies can be ruled out})}
\label{sec: critical bias red do not exist}

Recall that a bias redundancy as defined in Remark \ref{rem: taxonomy of redundant parameters}
implies that \(\activation_k(x) := \activation(\bias_k + \langle x,
\weight_{\placeholder
k}\rangle)\) is constant on \(\Domain\).

\begin{lemma}[Bias redundancy characterization]
    \label{lem: bias redundancy characterization}
    Let the activation function \(\activation\) be injective and assume
    \begin{equation}
        \label{eq: orthogonal}    
        \{x-y: x,y\in \Domain\}^\perp = \{0\}.
    \end{equation}
    Then a bias redundancy at neuron \(k\) implies \(\weight_{\placeholder k} = 0\).
\end{lemma}
Observe that \eqref{eq: orthogonal} is satisfied as soon as \(\Domain\) contains
an open set.
\begin{proof}
    Let there be a bias redundancy at neuron \(k\). If there were
    \(x,y\in \Domain\) such that \(\langle x - y, \weight_{\placeholder k}\rangle \neq 0\),
    then \(\activation_k(x) \neq \activation_k(y)\) due to
    injectivity. Consequently \(\langle x-y, \weight_{\placeholder k}\rangle = 0\)
    for all \(x,y\in \Domain\) and \eqref{eq: orthogonal}
    thereby implies \(\weight_{\placeholder k}  = 0\).
\end{proof}

Recall that we assumed in Proposition \ref{prop: pure bias redundancies can be
ruled out} \(\# V_\tout=1\),  \(\activation'(x) \neq 0\) for all \(x\in \real\),
which also implies \(\activation\) is injective as it is
strictly monotonous, and an open set in \(\Domain\) such that Lemma~\ref{lem:
bias redundancy characterization} is satisfied.

In Section~\ref{sec: pruning} we discussed a general pruning procedure that
proceeds in steps, removing one neuron at a time. Consequently, this procedure
is path dependent. If a different neuron were removed first one might
end up with a different pruned network and parameter. In the case where we
only have bias redundancies we can do better.
Assume the requirements of Lemma~\ref{lem: bias redundancy characterization}
are satisfied, let \(I\subseteq V_1\) be the maximal set
such that \(\weight_{\placeholder j} = 0\) for all \(j\in I\).
We then define the pruned network \(\tilde{\network} := ((V_0, V_1\setminus I, V_2), \activation)\)
in a single step: The parameter \(\tilde \param\) 
retains all the weights and biases from \(\param\)
restricted to the pruned ANN-graph except for
\[
    \tilde \bias_l := \bias_l + \sum_{j\in I}\weight_{jl}\activation(\bias_j)
    \qquad l \in V_2.
\]
It is straightforward to show that the response then remains the same.
In the following lemma we will relate the criticality of \(\tilde \param\)
to that of \(\param\).

\begin{lemma}[Characterization of critical bias redundancies]
        \label{lem: characterization of critical bias redundancies}
        Assume the setting of Proposition \ref{prop: pure bias redundancies can be ruled out}.
        If a critical point \(\param\) of the cost \(\cost_\problem\) only has
        bias redundancies, then the parameter \(\tilde \param\) 
        of the pruned network is also a critical point of \(\cost_\problem\) and
        the following equation is satisfied
        \begin{equation}
            \label{eq: condition for extended is crit. point}
            \E_\problem\bigl[
                \partial_{\hat y}\loss(\response_{\tilde\param}(X), Y)X_i
            \bigr]
            = 0 \qquad \forall i\in V_0.
        \end{equation}
        This is furthermore sufficient, i.e. if \(\tilde \param\) is critical
        and \eqref{eq: condition for extended is crit. point} is satisfied
        then the original parameter \(\param\) is critical.
\end{lemma}
\begin{remark}
    We do not make use of the squared error loss function and only
    require sufficient regularity that derivatives may be moved into the
    expectation.
\end{remark}
\begin{proof}
    ``\(\Rightarrow\)'':
    That \(\nabla\cost_\problem(\param)=0\) implies \(\nabla\cost_\problem(\tilde{\param})=0\)
    is a straightforward exercise since the response remains the same and we retain
    almost all parameters except for the outer bias which does not occur in
    any of the partial derivatives of the response. Since we assume
    \(V_1=\{\outSgt\}\) in this section we furthermore have
    \begin{equation}
        \label{eq: derive bias critical condition}
        0 = \partial_{\weight_{ij}}\cost_\problem(\param)
        = \weight_{j\outSgt} \E_\problem\bigl[
            \partial_{\hat y}\loss(\response_\param(X), Y)X_i
        \bigr]\activation'(\bias_j).
    \end{equation}
    And since we assume \(\activation'(x) \neq 0\) for all \(x\in \real\)
    in this section and \(\weight_{j\outSgt} \neq 0\) since we ruled out
    deactivation redundancies, we obtain \eqref{eq: condition for extended is crit. point}.
    
    ``\(\Leftarrow\)'':
    Using \eqref{eq: condition for extended is crit. point}
    and \(\nabla\cost_\problem(\tilde{\param}) = 0\) we now have to prove
    \(\nabla\cost_\problem(\param)=0\). The directional derivatives of the
    parameters that remained on the pruned ANN-graph are zero because they
    coincide with those of \(\tilde{\param}\). What is left are therefore
    the directional derivatives of the parameters attached to the nodes \(j\in
    I\). Using \eqref{eq: derive bias critical condition}
    in reverse with \eqref{eq: condition for extended is crit. point}
    we obtain that \(\partial_{\weight_{ij}}\cost_\problem(\param) =0\)
    What is left are therefore the biases \(\bias_j\) with \(j\in I\).
    Those are given by
    \[
        \partial_{\bias_j} \cost_\problem(\param)
        = \E\bigl[
            \partial_{\hat y}\loss(\response_\param(X), Y)
            \activation'\bigl(\bias_j\bigr)
        \bigr]\weight_{j\outSgt} 
        = \underbrace{\partial_{\tilde{\bias}_\outSgt}\cost_\problem(\tilde \param)}_{=0}
        \activation'\bigl(\bias_j\bigr)\weight_{j\outSgt}.
        \qedhere
    \]
\end{proof}

\subsubsection{The pruned condition a.s.\ never happens (Proof of \ref{it: no pure bias redundancies})}

With the characterization of critical bias redundancies (Lemma~\ref{lem: characterization
of critical bias redundancies}), proving the non-existence of bias redundancies
is equivalent to proving that a parameter vector
of an efficient network can never be a critical point which also
satisfies \eqref{eq: condition for extended is crit. point}.
To prove \ref{it: no pure bias redundancies} we therefore simply have to show
that \eqref{eq: condition for extended is crit. point} almost
surely never coincides with an efficient critical point.

Since the efficient parameters are automatically
\((1,0,1)\)-polynomially efficient by assumption, we
need to show that the set \(\polyEfficient^{(1,0,1)}\)
almost surely does not contain critical points that
satisfy \eqref{eq: condition for extended is crit. point}.
Recall that we assume the squared error \(\loss(\hat{y}, y) = (\hat y - y)^2\)
in \ref{it: no pure bias redundancies} and therefore
\eqref{eq: condition for extended is crit. point}
reduces to
\[
    0 = \E_\problem\bigl[(\response_\param(X) - Y)X_i\bigr]
    \overset{\text{tower}}= \E_\problem\bigl[(\response_\param(X) - \target_\problem(X))X_i\bigr]
    \qquad \forall i \in V_0.
\]
For the random cost \(\Cost=\cost_\Problem\) this means
\[
    0 = \int (\response_\param(x) - \rf(x))x_i \Pr_X(dx)
    = \langle \response, \pi_i\rangle_{\Pr_X} - \langle \rf, \pi_i\rangle_{\Pr_X}
\]
with random target \(\rf=\target_\Problem\) and projection \(\pi_i:x\mapsto x_i\).
This combination can be captured by	the level set \(\rg^{-1}(0)\) of
\[
    \rg:\begin{cases}
        \polyEfficient^{(1,0,1)} \to \real^{\dim(\param)}\times \real^{V_\tin}
        \\
        \param \mapsto
        \Bigl(
            \nabla\Cost(\param),
            \bigl(
                \langle \response, \pi_i\rangle_{\Pr_X} - \langle \rf, \pi_i\rangle_{\Pr_X}
            \bigr)_{i\in V_\tin}
        \Bigr).
    \end{cases}
\]
Since \(\polyEfficient^{(1,0,1)}\subseteq \real^{\dim(\param)}\), \(\rg\) is a
mapping into a larger dimension. Its level sets are therefore empty with
probability one by Lemma~\ref{lem: generalization 11.2.10},
assuming we can prove it to be non-degenerate for every \(\param\in
\polyEfficient^{(1,0,1)}\). This will therefore be the finial step of
the proof. Since the variance of \(\rg\) is independent of the mean,
we may consider \(\hat{\Cost}(\param)=\langle \response_\param, \rf\rangle_{\Pr_X}\) as introduced in Proposition~\ref{prop:
decomposition} instead and similarly prune \(\langle \response_\param, \pi_i\rangle_{\Pr_X}\), i.e. we may consider
\[
    \hat{\rg}(\param)
    = \Bigl(\nabla_\param\langle \response_\param, \rf\rangle_{\Pr_X}, (\langle \rf, \pi_i\rangle_{\Pr_X})_{i\in V_\tin}\Bigr)
    = \Bigl(\langle \nabla_\param\response_\param, \rf\rangle_{\Pr_X}, (\langle \pi_i, \rf\rangle_{\Pr_X})_{i\in V_\tin}\Bigr)
\]
Similar to our argument in the proof of Proposition~\ref{prop: non-degenrate g}
it is therefore sufficient to prove the linear independence of the following
functions
\begin{align}
    &x\mapsto 1 
    \tag{\(\partial \bias_{\outSgt}\)}
    \\
    &x \mapsto \activation\bigl(\bias_j + \langle x, \weight_{\placeholder j}\rangle\bigr)
    && j\in V_1 
    \tag{\(\partial \weight_{j\placeholder}\)}
    \\
    &x \mapsto \activation'\bigl(\bias_j + \langle x, \weight_{\placeholder j}\rangle\bigr)\weight_{j\placeholder} x_i
    && j\in V_1, i\in V_0
    \tag{\(\partial \weight_{ij}\)}
    \\
    &x \mapsto \activation'\bigl(\bias_j + \langle x, \weight_{\placeholder j}\rangle\bigr)\weight_{j\placeholder} 
    && j\in V_1 
    \tag{\(\partial \bias_j\)}
    \\
    &x\mapsto x_i
    && i\in V_0
    \tag{\(\pi_i\)},
\end{align}
where the last equation is the extra condition that follows from \eqref{eq: condition for extended is crit. point}. But their linear
independence follows from the \((1,0,1)\)-polynomial independence
(Definition~\ref{def: polynomial independence}).

Note, that we did not require second order derivatives, but
first order polynomials in the affine term due to the extra condition
\eqref{eq: condition for extended is crit. point}.
This explains why we required \((1,0,1)\)-polynomial independence
instead of \((0,0,1,2)\)-polynomial independence as in Proposition~\ref{prop:
non-degenrate g}.

%% file: ms.bbl
\begin{thebibliography}{26}
\providecommand{\natexlab}[1]{#1}
\providecommand{\url}[1]{\texttt{#1}}
\expandafter\ifx\csname urlstyle\endcsname\relax
  \providecommand{\doi}[1]{doi: #1}\else
  \providecommand{\doi}{doi: \begingroup \urlstyle{rm}\Url}\fi

\bibitem[Adler and Taylor(2007)]{adlerRandomFieldsGeometry2007}
R.~J. Adler and J.~E. Taylor.
\newblock \emph{Random {{Fields}} and {{Geometry}}}.
\newblock Springer {{Monographs}} in {{Mathematics}}. Springer New York, New
  York, NY, 2007.
\newblock ISBN 978-0-387-48112-8.
\newblock \doi{10.1007/978-0-387-48116-6}.

\bibitem[Bogachev(1998)]{bogachevGaussianMeasures1998}
V.~I. Bogachev.
\newblock \emph{Gaussian {{Measures}}}, volume~62 of \emph{Mathematical
  {{Surveys}} and {{Monographs}}}.
\newblock American Mathematical Society, Providence, RI, 1998.
\newblock ISBN 0-8218-1054-5.

\bibitem[Carmeli et~al.(2010)Carmeli, De~Vito, Toigo, and
  Umanit{\`a}]{carmeliVectorValuedReproducing2010}
C.~Carmeli, E.~De~Vito, A.~Toigo, and V.~Umanit{\`a}.
\newblock Vector valued reproducing kernel hilbert spaces and universality.
\newblock \emph{Analysis and Applications}, 08\penalty0 (01):\penalty0 19--61,
  Jan. 2010.
\newblock ISSN 0219-5305.
\newblock \doi{10.1142/S0219530510001503}.

\bibitem[Choromanska et~al.(2015)Choromanska, Henaff, Mathieu, Arous, and
  LeCun]{choromanskaLossSurfacesMultilayer2015}
A.~Choromanska, {\relax Mi}.~Henaff, M.~Mathieu, G.~B. Arous, and Y.~LeCun.
\newblock The {{Loss Surfaces}} of {{Multilayer Networks}}.
\newblock In \emph{Proceedings of the {{Eighteenth International Conference}}
  on {{Artificial Intelligence}} and {{Statistics}}}, pages 192--204. PMLR,
  Feb. 2015.

\bibitem[Dereich and Kassing(2024)]{dereichConvergenceStochasticGradient2024}
S.~Dereich and S.~Kassing.
\newblock Convergence of {{Stochastic Gradient Descent Schemes}} for
  {{{\L}ojasiewicz-Landscapes}}.
\newblock \emph{Journal of Machine Learning}, 3\penalty0 (3):\penalty0
  245--281, June 2024.
\newblock ISSN 2790-203X, 2790-2048.
\newblock \doi{10.4208/jml.240109}.

\bibitem[Freeman and Bruna(2017)]{freemanTopologyGeometryHalfRectified2017a}
C.~D. Freeman and J.~Bruna.
\newblock Topology and {{Geometry}} of {{Half-Rectified Network Optimization}}.
\newblock In \emph{International {{Conference}} on {{Learning
  Representations}}}, Feb. 2017.

\bibitem[Ge et~al.(2015)Ge, Huang, Jin, and Yuan]{geEscapingSaddlePoints2015}
R.~Ge, F.~Huang, C.~Jin, and Y.~Yuan.
\newblock Escaping {{From Saddle Points}} --- {{Online Stochastic Gradient}}
  for {{Tensor Decomposition}}.
\newblock In \emph{Conference on {{Learning Theory}}}, pages 797--842. PMLR,
  June 2015.

\bibitem[Hinton(2012)]{hintonNeuralNetworksMachine2012}
G.~Hinton.
\newblock Neural {{Networks}} for {{Machine Learning}}, 2012.

\bibitem[Kawaguchi(2016)]{kawaguchiDeepLearningPoor2016}
K.~Kawaguchi.
\newblock Deep {{Learning}} without {{Poor Local Minima}}.
\newblock In \emph{Advances in {{Neural Information Processing Systems}}},
  volume~29. Curran Associates, Inc., 2016.

\bibitem[Kingma and Ba(2015)]{kingmaAdamMethodStochastic2015}
D.~P. Kingma and J.~Ba.
\newblock Adam: {{A Method}} for {{Stochastic Optimization}}.
\newblock In \emph{Proceedings of the 3rd {{International Conference}} on
  {{Learning Representations}}}, San Diego, 2015.

\bibitem[Micchelli et~al.(2006)Micchelli, Xu, and
  Zhang]{micchelliUniversalKernels2006}
C.~A. Micchelli, Y.~Xu, and H.~Zhang.
\newblock Universal {{Kernels}}.
\newblock \emph{Journal of Machine Learning Research}, 7\penalty0 (12), 2006.

\bibitem[Mityagin(2020)]{mityaginZeroSetReal2020}
B.~S. Mityagin.
\newblock The {{Zero Set}} of a {{Real Analytic Function}}.
\newblock \emph{Mathematical Notes}, 107\penalty0 (3):\penalty0 529--530, Mar.
  2020.
\newblock ISSN 1573-8876.
\newblock \doi{10.1134/S0001434620030189}.

\bibitem[Moulines and Bach(2011)]{moulinesNonAsymptoticAnalysisStochastic2011}
E.~Moulines and F.~Bach.
\newblock Non-{{Asymptotic Analysis}} of {{Stochastic Approximation
  Algorithms}} for {{Machine Learning}}.
\newblock In \emph{Advances in {{Neural Information Processing Systems}}},
  volume~24. Curran Associates, Inc., 2011.

\bibitem[Nesterov(2018)]{nesterovLecturesConvexOptimization2018}
Y.~E. Nesterov.
\newblock \emph{Lectures on {{Convex Optimization}}}.
\newblock Springer Optimization and {{Its}} Applications; Volume 137. Springer,
  Cham, second edition edition, 2018.
\newblock ISBN 978-3-319-91578-4.
\newblock \doi{10.1007/978-3-319-91578-4}.

\bibitem[Nguyen(2019)]{nguyenConnectedSublevelSets2019}
Q.~Nguyen.
\newblock On {{Connected Sublevel Sets}} in {{Deep Learning}}.
\newblock In \emph{Proceedings of the 36th {{International Conference}} on
  {{Machine Learning}}}, pages 4790--4799. PMLR, May 2019.

\bibitem[Polyak(1964)]{polyakMethodsSpeedingConvergence1964}
B.~T. Polyak.
\newblock Some methods of speeding up the convergence of iteration methods.
\newblock \emph{USSR Computational Mathematics and Mathematical Physics},
  4\penalty0 (5):\penalty0 1--17, Jan. 1964.
\newblock ISSN 0041-5553.
\newblock \doi{10.1016/0041-5553(64)90137-5}.

\bibitem[Polyak(1990)]{polyakNewMethodStochastic1990}
B.~T. Polyak.
\newblock {A new method of stochastic approximation type}.
\newblock \emph{Avtomatika i telemekhanika}, 51\penalty0 (7):\penalty0 98--107,
  1990.

\bibitem[Riordan(2002)]{riordanIntroductionCombinatorialAnalysis2002}
J.~Riordan.
\newblock \emph{Introduction to {{Combinatorial Analysis}}}.
\newblock Dover Publications Inc., Mineola, N.Y, dover edition edition, Dec.
  2002.
\newblock ISBN 978-0-486-42536-8.

\bibitem[Robbins and Monro(1951)]{robbinsStochasticApproximationMethod1951a}
H.~Robbins and S.~Monro.
\newblock A {{Stochastic Approximation Method}}.
\newblock \emph{The Annals of Mathematical Statistics}, 22\penalty0
  (3):\penalty0 400--407, Sept. 1951.
\newblock ISSN 0003-4851, 2168-8990.
\newblock \doi{10.1214/aoms/1177729586}.

\bibitem[Rumelhart et~al.(1986)Rumelhart, Hinton, and
  Williams]{rumelhartLearningRepresentationsBackpropagating1986}
D.~E. Rumelhart, G.~E. Hinton, and R.~J. Williams.
\newblock Learning representations by back-propagating errors.
\newblock \emph{Nature}, 323\penalty0 (6088):\penalty0 533--536, Oct. 1986.
\newblock ISSN 0028-0836, 1476-4687.
\newblock \doi{10.1038/323533a0}.

\bibitem[Ruppert(1988)]{ruppertEfficientEstimationsSlowly1988}
D.~Ruppert.
\newblock Efficient {{Estimations}} from a {{Slowly Convergent Robbins-Monro
  Process}}.
\newblock Technical report, {Cornell University Operations Research and
  Industrial Engineering}, Feb. 1988.

\bibitem[Sacks(1958)]{sacksAsymptoticDistributionStochastic1958}
J.~Sacks.
\newblock Asymptotic {{Distribution}} of {{Stochastic Approximation
  Procedures}}.
\newblock \emph{The Annals of Mathematical Statistics}, 29\penalty0
  (2):\penalty0 373--405, June 1958.
\newblock ISSN 0003-4851, 2168-8990.
\newblock \doi{10.1214/aoms/1177706619}.

\bibitem[Stanley(2011)]{stanleyEnumerativeCombinatoricsVolume2011}
R.~P. Stanley.
\newblock \emph{Enumerative {{Combinatorics}}: {{Volume}} 1}.
\newblock Cambridge University Press, Cambridge, NY, 2nd edition edition, Dec.
  2011.
\newblock ISBN 978-1-107-01542-5.

\bibitem[Szwarc(2022)]{szwarcHowProveThat2022}
R.~Szwarc.
\newblock How to prove that generalized {{Vandermonde}} matrix is invertible?
\newblock Mathematics Stack Exchange, Oct. 2022.

\bibitem[Talagrand(1987)]{talagrandRegularityGaussianProcesses1987}
M.~Talagrand.
\newblock Regularity of gaussian processes.
\newblock \emph{Acta Mathematica}, 159\penalty0 (none):\penalty0 99--149, Jan.
  1987.
\newblock ISSN 0001-5962, 1871-2509.
\newblock \doi{10.1007/BF02392556}.

\bibitem[Venturi et~al.(2019)Venturi, Bandeira, and
  Bruna]{venturiSpuriousValleysOnehiddenlayer2019}
L.~Venturi, A.~S. Bandeira, and J.~Bruna.
\newblock Spurious {{Valleys}} in {{One-hidden-layer Neural Network
  Optimization Landscapes}}.
\newblock \emph{Journal of Machine Learning Research}, 20\penalty0
  (133):\penalty0 1--34, 2019.
\newblock ISSN 1533-7928.

\end{thebibliography}
